\pdfoutput=1

\documentclass[11pt]{article}

\usepackage[]{ACL2023}

\usepackage{times}
\usepackage{latexsym}

\usepackage[T1]{fontenc}

\usepackage[utf8]{inputenc}

\usepackage{microtype}

\usepackage{inconsolata}

\usepackage{graphicx}
\usepackage{microtype}
\usepackage{array}    
\usepackage{siunitx}  
\usepackage{multirow}
\usepackage{graphicx}
\usepackage{subcaption}
\usepackage{listings}
\usepackage{xcolor}
\usepackage{booktabs}
\usepackage{longtable}
\usepackage{array}
\usepackage{xcolor}
\usepackage{tcolorbox}
\usepackage{geometry}
\usepackage{amsmath, amssymb}
\usepackage{newunicodechar}
\usepackage[ruled, lined, linesnumbered, commentsnumbered, longend]{algorithm2e}

\title{Verification and Refinement of Natural Language Explanations\\ through LLM-Symbolic Theorem Proving}

\author{Xin Quan$^1$, Marco Valentino$^2$, Louise A. Dennis$^1$, Andr\'e Freitas$^{1,2,3}$ \\ 
$^{1}$ Department of Computer Science, University of Manchester, UK \\ 
$^{2}$Idiap Research Institute, Switzerland \\
$^{3}$ National Biomarker Centre, CRUK-MI, University of Manchester, UK\\
$^1$\tt{\{name.surname\}@manchester.ac.uk}\\
$^2$\tt{\{name.surname\}@idiap.ch}}

\begin{document}
\maketitle
\begin{abstract}
Natural language explanations represent a proxy for evaluating explanation-based and multi-step Natural Language Inference (NLI) models. However, assessing the validity of explanations for NLI is challenging as it typically involves the crowd-sourcing of apposite datasets, a process that is time-consuming and prone to logical errors. To address existing limitations, this paper investigates the verification and refinement of natural language explanations through the integration of Large Language Models (LLMs) and Theorem Provers (TPs). Specifically, we present a neuro-symbolic framework, named Explanation-Refiner, that integrates TPs with LLMs to generate and formalise explanatory sentences and suggest potential inference strategies for NLI. In turn, the TP is employed to provide formal guarantees on the logical validity of the explanations and to generate feedback for subsequent improvements. We demonstrate how Explanation-Refiner can be jointly used to evaluate explanatory reasoning, autoformalisation, and error correction mechanisms of state-of-the-art LLMs as well as to automatically enhance the quality of explanations of variable complexity in different domains.\footnote{Code and data are available at: \href{https://github.com/neuro-symbolic-ai/explanation_refinement}{https://github.com/neuro-symbolic-ai/explanation\_refinement}}
\end{abstract}

\section{Introduction}
A recent line of research in Natural Language Inference (NLI) focuses on developing models capable of generating natural language explanations in support of their predictions \citep{ thayaparan-etal-2021-explainable,chen-etal-2021-kace, valentino2022hybrid, bostrom-etal-2022-natural, weir2023nellie}. Since natural language explanations can be used as a proxy to evaluate the underlying reasoning process of NLI models \citep{kumar-talukdar-2020-nile, Zhao_Vydiswaran_2021, chen-etal-2021-kace}, researchers have proposed different methods for assessing their intrinsic quality \citep{camburu-etal-2020-make, wiegreffe2021teach, valentino-etal-2021-natural, atanasova-etal-2023-faithfulness, quan-etal-2024-enhancing, dalal2024inference}, including the adoption of language generation metrics for a direct comparison between models' generated explanations and human-annotated explanations.

However, this process is subject to different types of limitations. First, the use of language generation metrics requires the crowd-sourcing of explanation corpora to augment existing NLI datasets \cite{wiegreffe2021teach}, a process that is time-consuming and susceptible to errors \citep{valentino-etal-2021-natural, liu-etal-2022-toward, zhao-etal-2023-abductive}. Second, language generation metrics have been shown to fail capturing fine-grained properties that are fundamental for NLI such as logical reasoning, faithfulness, and robustness \citep{camburu-etal-2020-make, chan-etal-2022-comparative, atanasova-etal-2023-faithfulness, quan-etal-2024-enhancing}. Third, human explanations in NLI datasets tend to be incomplete and contain logical errors that could heavily bias the evaluation \cite{elazar-etal-2021-back, valentino-etal-2021-natural}.

In this paper, we investigate the integration of state-of-the-art LLM-based explanation generation models for NLI with external logical solvers to jointly evaluate explanatory reasoning \cite{pan-etal-2023-logic, olausson-etal-2023-linc, jiang2024leanreasoner} and enhance the quality of crowd-sourced explanations. In particular, we present a neuro-symbolic framework, named Explanation-Refiner, that integrates a Theorem Prover (TP) with Large Language Models (LLMs) to investigate the following research questions: \emph{RQ1: ``Can the integration of LLMs and TPs provide a mechanism for automatic verification and refinement of natural language explanations?''; RQ2: ``Can the integration of LLMs and TPs improve the logical validity of human-annotated explanations?''; RQ3: ``To what extent are state-of-the-art LLMs capable of explanatory reasoning, autoformalisation, and error correction for NLI in different domains?''.} To answer these questions, Explanation-Refiner employs LLMs to generate and formalise explanatory sentences and to suggest potential inference strategies for building non-redundant, complete, and logically valid explanations for NLI. In turn, the TP is adopted to verify the validity of the explanations through the construction of deductive proofs and the generation of fine-grained feedback for LLMs. 

We instantiate Explanation-Refiner with state-of-the-art LLMs (i.e., GPT-4 \citep{DBLP:journals/corr/abs-2303-08774}, GPT-3.5 \citep{NEURIPS2020_1457c0d6}, LLama \citep{touvron2023llama}, and Mistral \citep{jiang2024mixtral}) and the Isabelle/HOL proof assistant \citep{nipkow2002isabelle} utilising Neo-Davidsonian event semantics \citep{Parsons1990EventsIT} coupled with First-Order Logic (FOL) to effectively and systematically translate natural language sentences into logical forms. 

Our empirical analysis, carried out on three NLI datasets of variable complexity (i.e., e-SNLI \citep{NEURIPS2018_4c7a167b}, QASC \citep{Khot2019QASC}, and WorldTree \citep{jansen-etal-2018-worldtree}), reveals that external feedback from TPs is effective in improving the quality of natural language explanations, leading to an increase in logical validity using GPT-4 from 36\% to 84\%, 12\% to 55\%, and 2\% to 37\% (on e-SNLI, QASC, and WorldTree respectively). At the same time, the results demonstrate that integrating external TPs with LLMs can reduce errors in autoformalisation, with an average reduction of syntax errors of 68.67\%, 62.31\%, and 55.17\%. Finally, we found notable differences in performance across LLMs and NLI datasets, with closed-sourced LLMs (i.e., GPT-4 and GPT-3.5) significantly outperforming open-source models (i.e., Mistral and LLama) on both explanatory reasoning and autoformalisation, along with a shared tendency of LLMs to struggle with increasing explanation complexity.

\begin{figure*}[t]
    \centering
    \includegraphics[width=\textwidth]{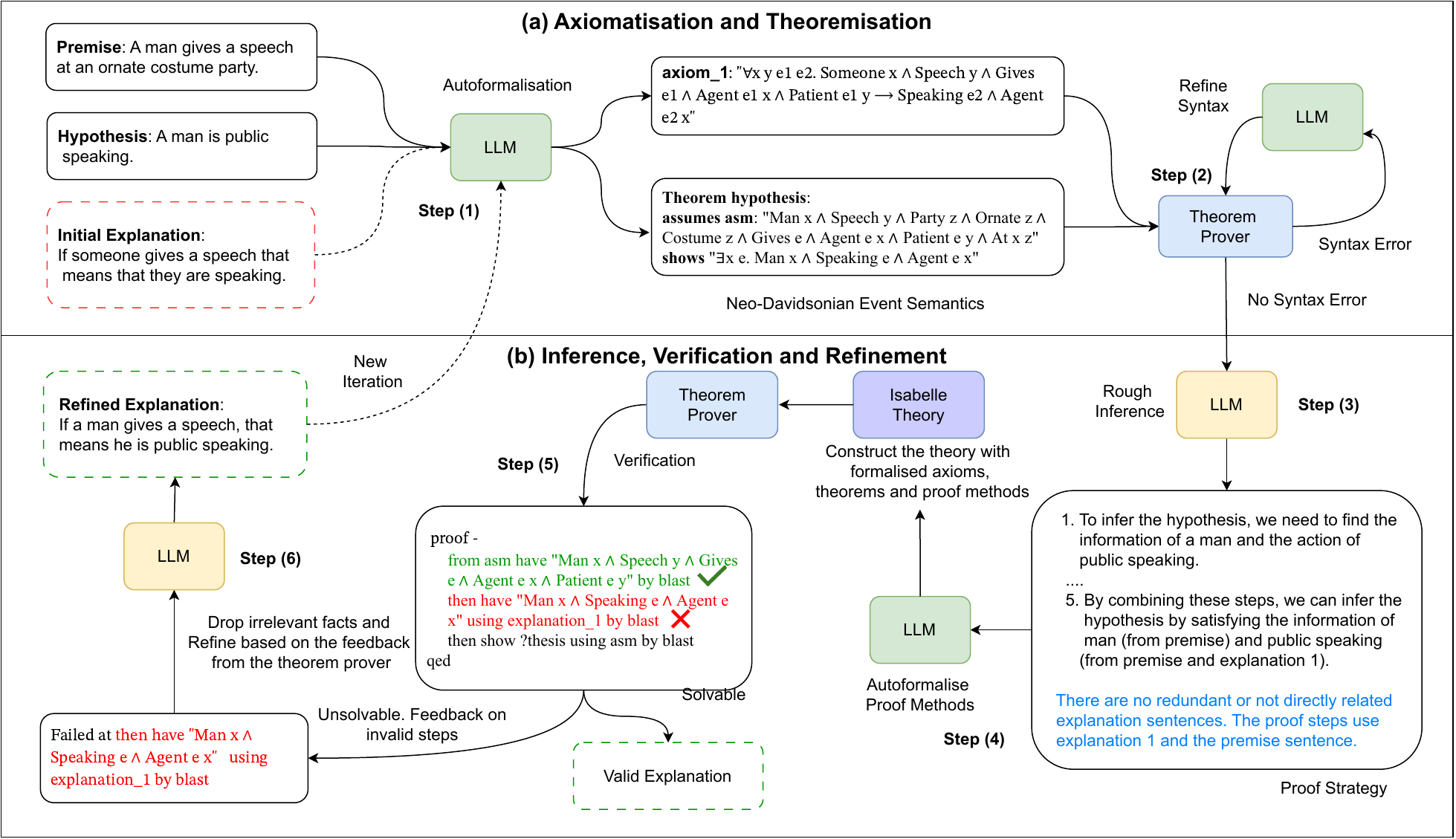}
    \caption{The overall pipeline of Explanation-Refiner: An NLI problem is converted into axioms and theorems for a theorem prover, along with some proof steps derived from a preliminary inference. In case the proof fails (logically invalid), the erroneous steps along with the constructed proof strategy are used as feedback to refine the explanation in a new iteration.}
\label{fig:framework}
\end{figure*}

To summarise, the main contributions of this paper are: 
\begin{enumerate}
    \item We introduce \textit{Explanation-Refiner}, a novel neuro-symbolic framework that integrates LLMs with an external theorem prover. This framework automatically verifies and refines explanatory sentences in NLI tasks using an objective external feedback.
    \item We integrate Neo-Davidsonian event semantics coupled with FOL to effectively translate natural language sentences into logical forms to minimise semantic information loss. Additionally, we introduce a novel method that leverages a theorem prover and a proof assistant for verifying NLI explanations and a syntactic refiner to minimise syntax errors in responses generated by LLMs.
    \item We conduct a comprehensive series of experiments with \textit{Explanation-Refiner} across five LLMs and three datasets, including 1 to 16 explanatory sentences, covering tasks from textual entailment to complex multiple-choice question answering in various domains.
    \item We perform extensive analyses to explore the explanation refinement process, characterising the LLMs' inference capabilities and revealing the strengths and limitations of different models in producing verifiable, explainable logical reasoning for NLI.
\end{enumerate}

\section{Explanation Verification and Refinement}
Explanation-based NLI is widely adopted to evaluate the reasoning process of multi-step inference models via the construction of natural language explanations. In this work, we refer to the following formalisation for Explanation-based NLI: given a premise sentence $p_i$, a hypothesis sentence $h_i$, and an explanation $E_i$ consisting of a set of facts $\{f_1,f_2,...,f_n\}$, the explanation $E_i$ is logically valid if and only if the entailment $p_i \cup E_i \models h_i$ holds. This entailment is considered verifiable if $\{p_i, E_i, h_i\}$ can be translated into a set of logical formulae $\Phi$ that compose a theory $\Theta$. The validity of the theory $\Theta$ is subsequently determined by a theorem prover, verifying whether $\Theta \vDash \psi$, where $\psi$ represents a logical consequence derived from the logical form of $h_i$.

In this paper, we aim to automatically verify the logical validity of an explanation $E_i$. To this end, if $\Theta \vDash \psi$ is rejected by the theorem prover, a further refinement stage should be initiated to refine the facts $\{f_1,f_2,...,f_n\}$ based on external feedback, resulting in an updated explanation $E_i'$. Thus, an explanation is accepted if all the facts are logically consistent, complementary and non-redundant to support the derivation. 

\section{Explanation-Refiner}
To verify the logical validity and refine any logical errors in explanatory sentences for NLI tasks, we present a neuro-symbolic framework that iteratively checks and refines the explanation $E_i$ based on external feedback. Figure \ref{fig:framework} shows an overview of our proposed framework. Given an NLI task, to evaluate the logical validity of the entailment, the LLM is prompted to perform an autoformalisation process that transforms natural language sentences into formal language represented in the form of an Isabelle/HOL theory. Each fact $f \in E_i$ is converted into an axiom $a_i$, where each $a_i$ is an element of the set $A = \{a_1, a_2, ..., a_n\}$. The premise $p_i$ and corresponding hypothesis $h_i$, is converted into a theorem for proving $p_i \wedge B \rightarrow h_i$, where $B \subseteq A$. A syntax refinement mechanism is subsequently applied to the previously transferred symbolic forms. The theorem prover is implemented as a checker to identify any syntax errors and provide these error details as feedback to an LLM, enabling the LLM to iteratively correct the syntax errors over a fixed number of iterations, denoted by $t$.

We can then perform automated reasoning via the theorem prover. To this end, in step 3 we use the LLM to generate a rough inference that states a preliminary proof strategy in natural language and elicit the facts $f \in E_i$ which are sufficient and necessary for entailing the hypothesis $h_i$. Based on this preliminary proof strategy, the LLM is prompted to construct and formalise the proof steps for proving the theorem. In step 5, the theorem prover will verify the constructed theory by attempting to prove the theorem. If it is solvable, we consider it a logically valid explanation. If the prover failed at one of the proof steps, we adopt the failed steps along with the applied axioms $B \subseteq A$ as an external feedback for the LLM. This feedback is used to refine the logical errors and consequently refine the facts $f \in E_i$. 
\begin{figure*}[t]
\centering
\begin{lstlisting}[]
theorem hypothesis:
  (* Premise: A smiling woman is playing the violin in front of a turquoise background. *)
  assumes asm: "Woman x $\wedge$ Violin y $\wedge$ Background z $\wedge$ Turquoise z $\wedge$ Smiling x $\wedge$ Playing e $\wedge$ Agent e x $\wedge$ Patient e y $\wedge$ InFrontOf x z"
  (* Hypothesis: A woman is playing an instrument. *)
  shows "$\exists$ x y e. Woman x $\wedge$ Instrument y $\wedge$ Playing e $\wedge$ Agent e x $\wedge$ Patient e y"
proof -
  from asm have "Woman x $\wedge$ Violin y $\wedge$ Playing e $\wedge$ Agent e x $\wedge$ Patient e y" by blast
  then have "Woman x $\wedge$ Instrument y $\wedge$ Playing e $\wedge$ Agent e x $\wedge$ Patient e y" using explanation_1 by blast
  then show ?thesis using asm by blast
qed
\end{lstlisting}
\caption{An example of representing the premise and hypothesis sentences in Isabelle/HOL theorem includes a proof constructed by the LLM for verifying the hypothesis.}
\label{isabelle_example}
\end{figure*}

\subsection{Autoformalisation}
In order to formally verify the logical validity of the explanations, we adopted Neo-Davidsonian event-based semantics and FOL. 
\paragraph{Neo-Davidsonian Event Semantics}
Preventing the loss of semantic information during the representation of natural language sentences in logical forms, such as FOL, poses significant challenges when using LLMs, particularly with long and complex sentences that are crucial for logical reasoning \citep{olausson-etal-2023-linc}. Neo-Davidsonian event semantics \citep{Parsons1990EventsIT} focused on event variables to represent the verb predicates and their corresponding object arguments as semantic roles. This approach establishes a predicate-argument structure that preserves the information content and faithfulness of complex sentences, closer to the surface form of the sentence \cite{quan-etal-2024-enhancing}. For example, the sentence ‘A wolf eating a sheep is an example of a predator hunting prey’ can be formalised as follows:
\begin{equation}
\begin{aligned}
    &\forall xye_1(\text{wolf}(x) \wedge \text{sheep}(y) \wedge \text{eating}(e_1) \\
    &\quad  \wedge \text{agent}(e_1,x) \wedge \text{patient}(e_1,y) \rightarrow\\
    &\quad  (\exists e_2 \; \text{predator}(x) \wedge \text{prey}(y) \wedge\\
    &\quad  \text{hunting}(e_2) \wedge \text{agent}(e_2,x) \wedge \\
    &\quad  \text{patient}(e_2,y) \wedge \text{example}(e_1,e_2)))
\end{aligned}
\label{wolf_example}
\end{equation}
In \ref{wolf_example}, the verbs are represented as the events ‘eating' and ‘hunting,' where the agent and patient arguments correspond to the entities performing and receiving the actions within these events, respectively. The logical form $\text{example}(e_1, e_2)$ explicitly captures the semantic meaning of this sentence: the event of a wolf eating a sheep as an exemplar of a predator hunting prey. Similarly, whenever there are no action verbs involved in a sentence, we use FOL to represent the static or descriptive aspects. For instance: 
\begin{align}
    &\forall x (\text{gravity}(x) \rightarrow \text{force}(x)) \\
    &\forall xy (\text{greater}(x,y) \rightarrow \text{larger}(x,y))
\end{align}

The above logical forms correspond to the sentences ‘gravity is a kind of force' and ‘greater means larger'.
\paragraph{Isabelle/HOL Theory Construction}
A theory script for the Isabelle/HOL theorem prover contains theorems that need to be proven from some axioms. Therefore, we adopt the sentences in an explanation to construct the set of axioms. For instance: 
\begin{lstlisting}[]
(* Explanation 1: A violin is an instrument. *)
axiomatization where
  explanation_1: "$\forall$x. Violin x $\longrightarrow$ Instrument x"
\end{lstlisting}
In addition, as illustrated in Figure \ref{isabelle_example}, both the premise and the hypothesis constitute parts of the theorem to be proven. In particular, the ‘assumes asm' clause includes unquantified, specific propositions or conjunctions of propositions which are recognised as known truths (i.e., premises). On the other hand, the ‘show' clause denotes the conclusion (i.e., hypothesis) for which we seek to build a proof through logical deductions based on the assumed propositions and axioms. 
\paragraph{Syntax Error Refiner}
Recent studies \citep{olausson-etal-2023-linc, gou2024critic} have revealed persistent syntax errors when prompting LLMs for code and symbolic form generation tasks. We categorised the syntax errors into two distinct subdomains based on feedback from Isabelle: type unification errors and other syntax errors. Type unification errors primarily arise from mismatches between declared and actual argument types in logical clauses. Other syntax errors typically involve missing brackets, undefined entity names, or invalid logical symbols. Our process involves using Isabelle to identify syntax errors in the transferred theory, extracting these error messages, and then prompting the LLM with these messages along with few-shot examples. This guides the model on how to correct each type of syntax error over a series of iterations, allowing for continuous verification and refinement. Details of the autoformalisation prompts are described in Appendix \ref{appendix_autoformalisation_prompts}. 
\subsection{Proof Construction}
A proof provides a detailed step-by-step strategy that elucidates the logical connections and unification among axioms to support the reasoning process aimed at achieving the solver's goal. Initially, we prompt the LLM to create a preliminary proof in natural language to assess how it infers the hypothesis and to identify which explanatory sentences are relevant, redundant, or unrelated. Based on this initial inference, we then guide the LLM to develop a formal proof (Figure \ref{isabelle_example}) that integrates with Isabelle/HOL to verify the explanatory sentences (axioms) that are required to derive the hypothesis. The general proof steps generated by an LLM are in the format 'show $X$ using $Y$ by $Z$', where the theorem prover is asked to prove $X$ given the assumptions $Y$, using the automated proof tactic $Z$. The proof tactic often applied is 'blast', which is one of broader Isabelle's FOL theorem proving tactics\citep{Paulson1999AGT}. Additional details of the proof construction process and the prompts used to guide the LLMs are described in Appendix \ref{appendix_proof_construction_prompts}.

\subsection{Verification and Refinement}
Finally, the constructed theory, which includes axioms, theorems, and proof steps, is submitted to the theorem prover for verification. If the theory is validated, it outputs a logically valid explanation. If the proof fails or timeouts, we extract the first error from the solver’s error message, identify the corresponding proof step, and locate the related explanatory sentences (axioms) from the theory. We begin by removing redundant and irrelevant facts that are not present in the preceding Isabelle/HOL proof steps or are declared as such in the text inference strategy. Then, we prompt the LLM to refine the explanatory sentences by providing it with the error message, the failed proof step, the associated proof strategy, and the relevant explanatory sentences for further iteration. This process is iterative and progressive; with each iteration, the framework addresses one or more logical errors, continually refining the explanatory sentences to ultimately yield a logically valid and verifiable explanation. Additional details on the prompts used for refinement are described in Appendix \ref{appendix_refine_explanation_prompts}.

\section{Empirical Evaluation}

\begin{figure*}[t]
\centering
\begin{subfigure}{.3\textwidth} 
  \centering
  \includegraphics[width=\linewidth]{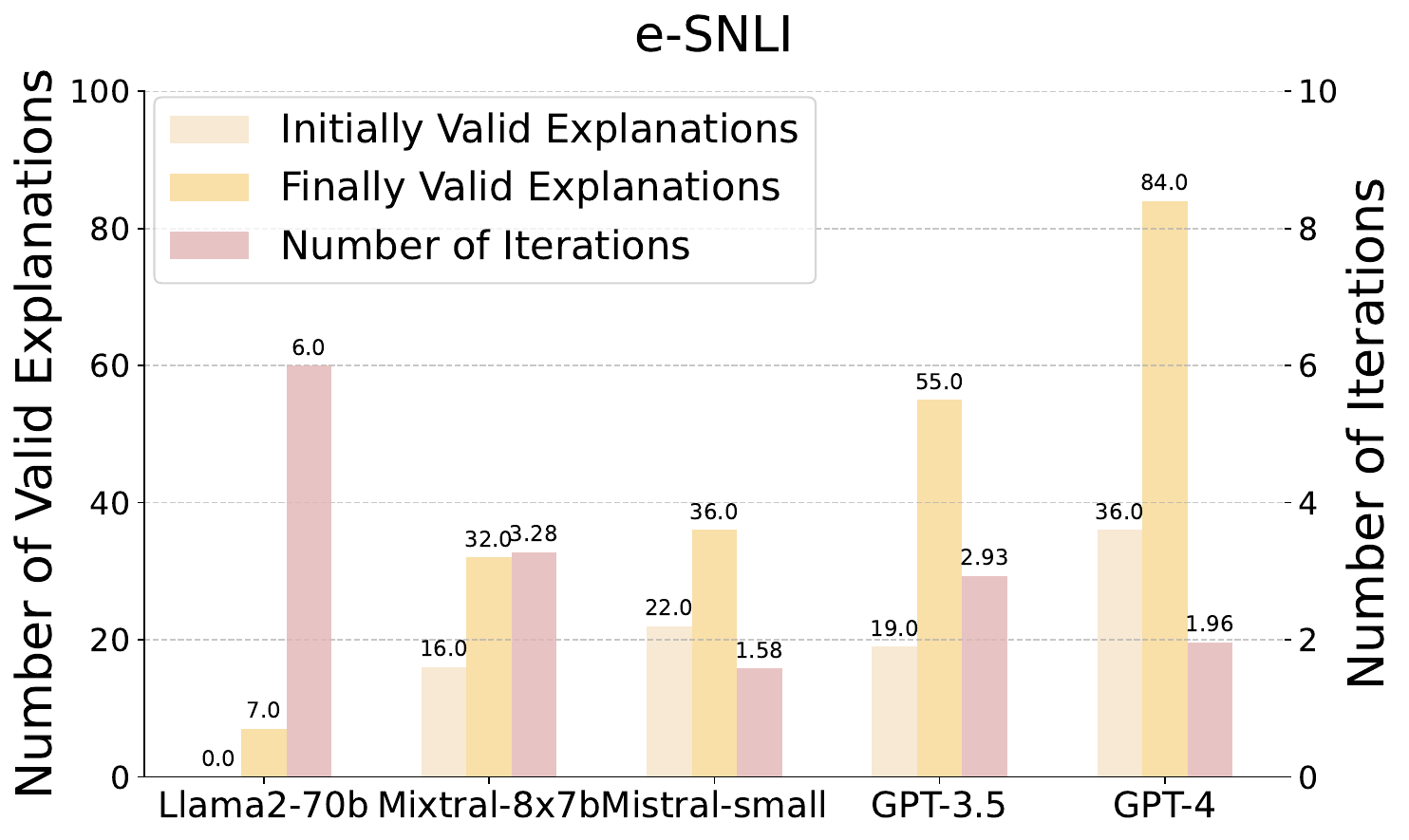}
  \caption{}
  \label{fig:sub1}
\end{subfigure}
\hfill
\begin{subfigure}{.3\textwidth} 
  \centering
  \includegraphics[width=\linewidth]{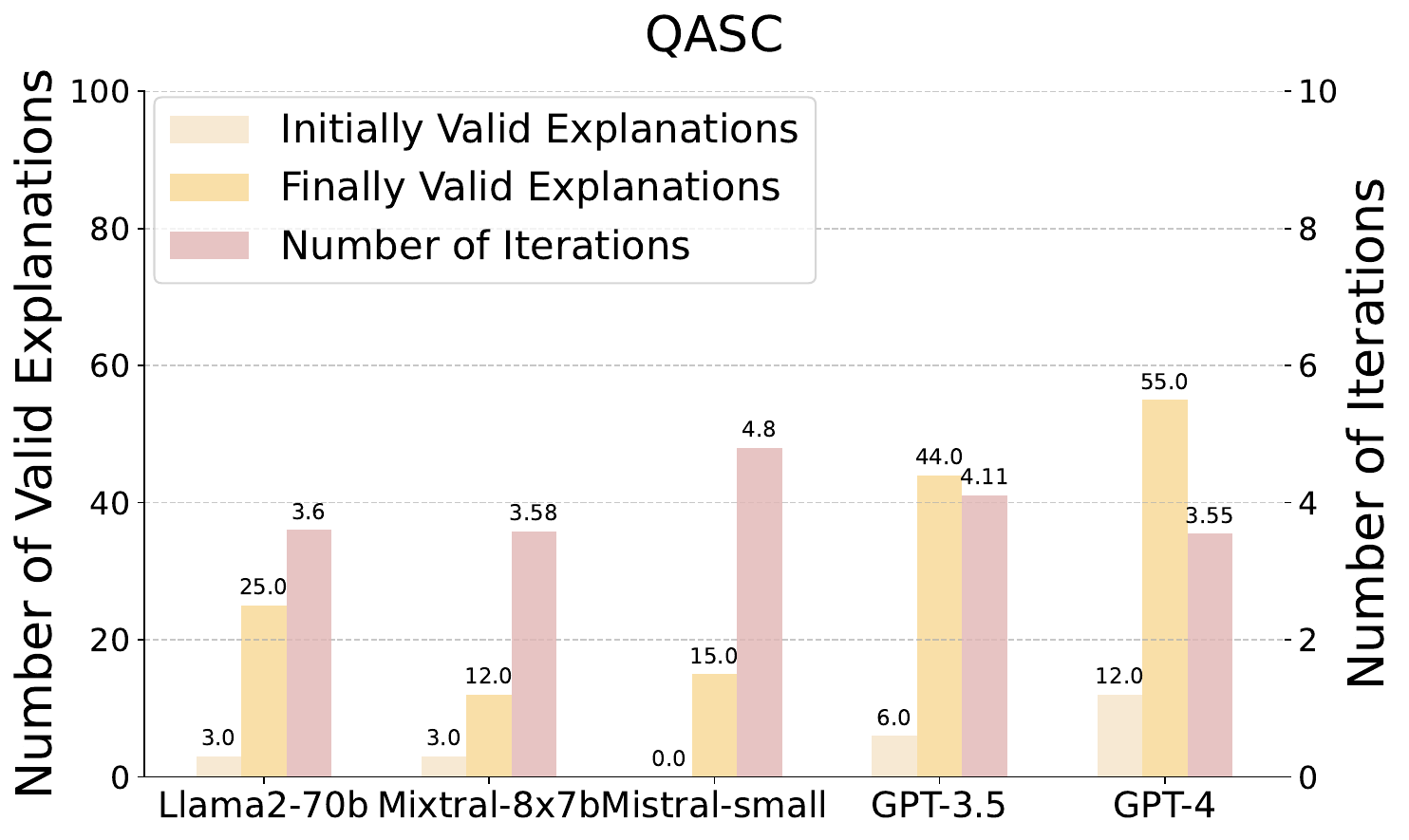}
  \caption{}
  \label{fig:sub2}
\end{subfigure}
\hfill
\begin{subfigure}{.3\textwidth} 
  \centering
  \includegraphics[width=\linewidth]{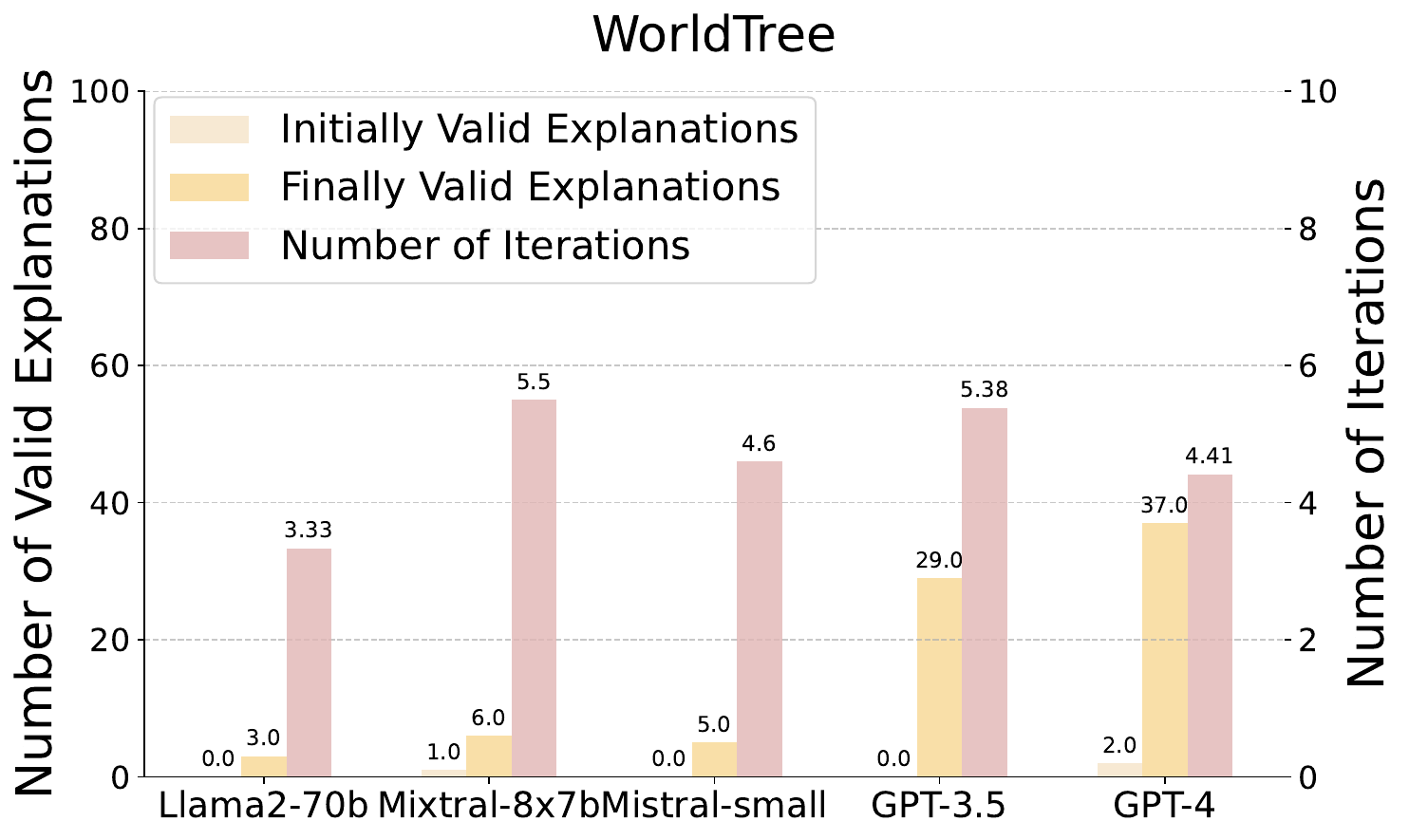}
  \caption{}
\end{subfigure}
\caption{The initial and final number of logically valid explanations, along with the average iteration times required to refine an explanation for each LLM}
\label{fig:overall_figure_bar_chart}
\end{figure*}

\begin{figure*}[t]
\centering
\begin{subfigure}{.3\textwidth} 
  \centering
  \includegraphics[width=\linewidth]{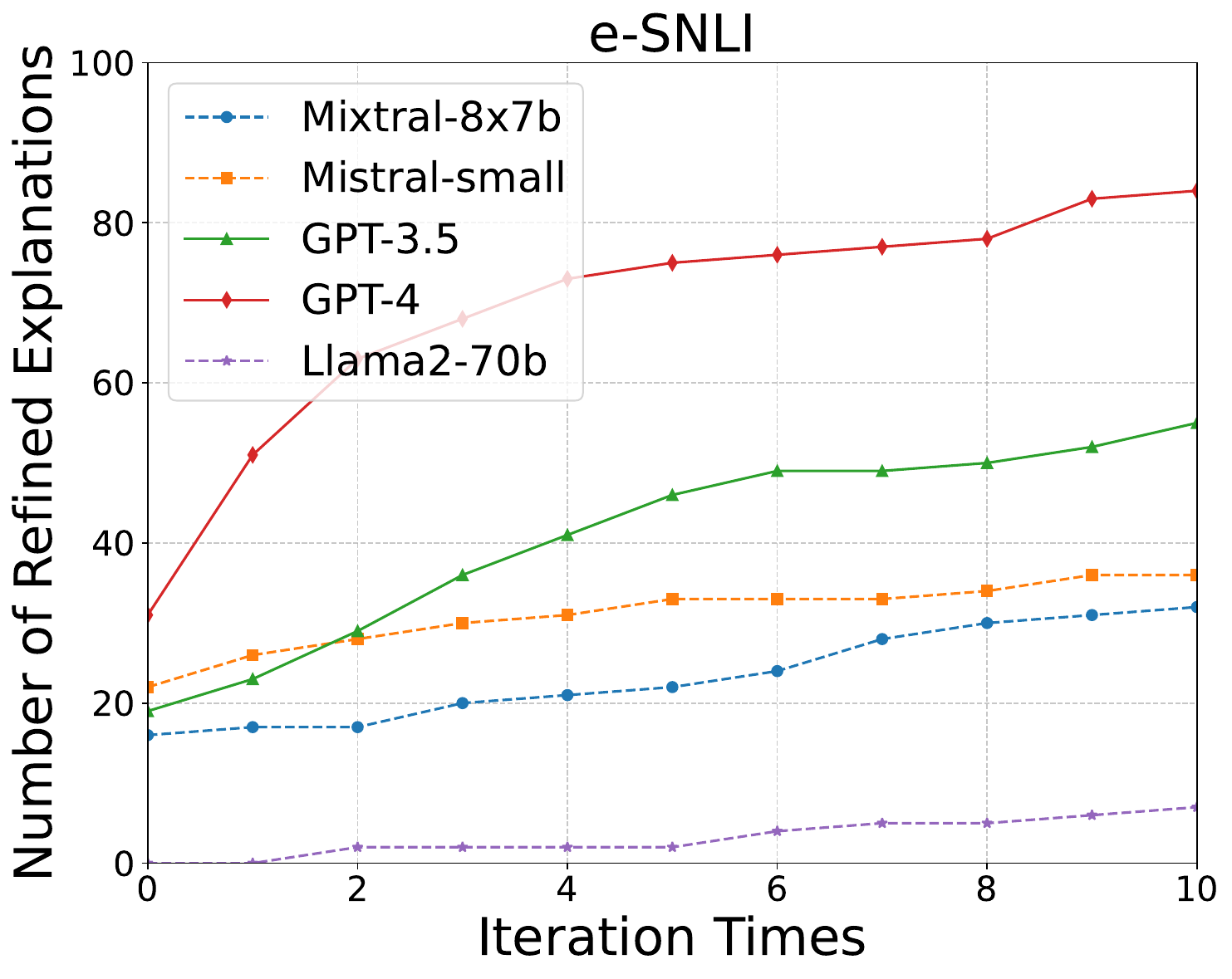}
  \caption{}
  \label{fig:sub1}
\end{subfigure}
\hfill
\begin{subfigure}{.3\textwidth} 
  \centering
  \includegraphics[width=\linewidth]{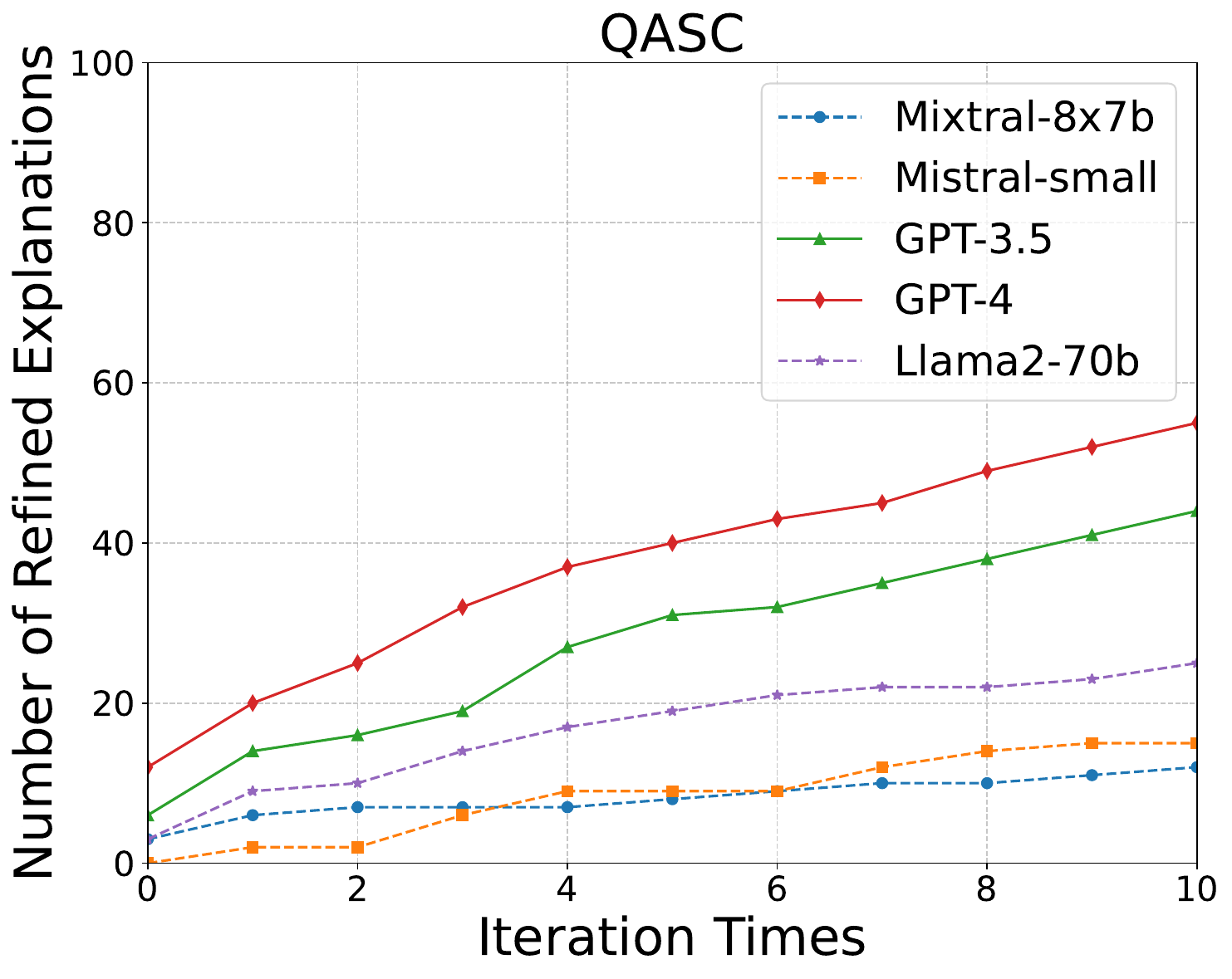}
  \caption{}
  \label{fig:sub2}
\end{subfigure}
\hfill
\begin{subfigure}{.3\textwidth} 
  \centering
  \includegraphics[width=\linewidth]{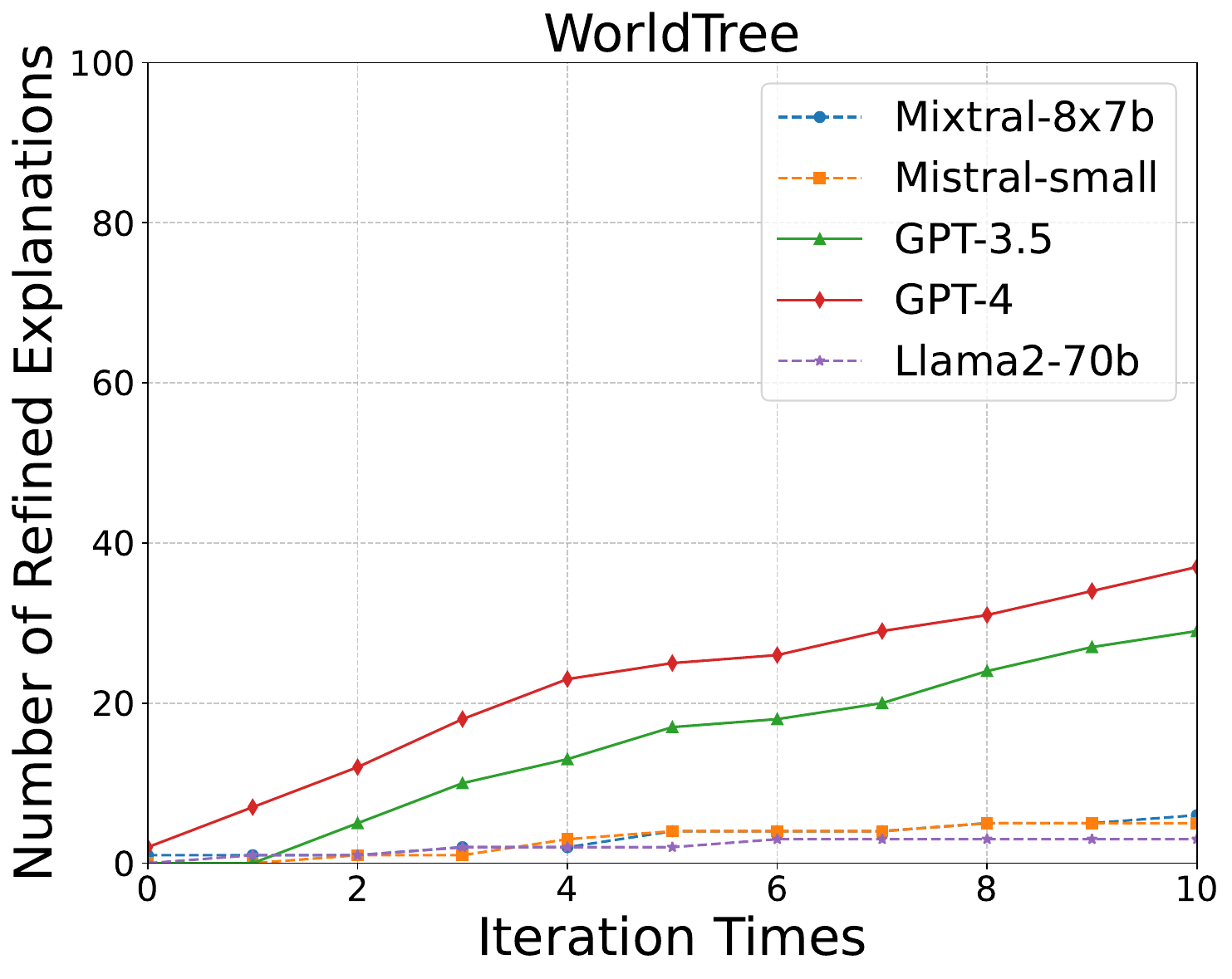}
  \caption{}
\end{subfigure}
\caption{Number of successfully refined explanations at each iteration step.}
\label{fig:overall_figure}
\end{figure*}

\begin{figure*}[h]
\centering
\begin{subfigure}{.3\textwidth} 
  \centering
  \includegraphics[width=\linewidth]{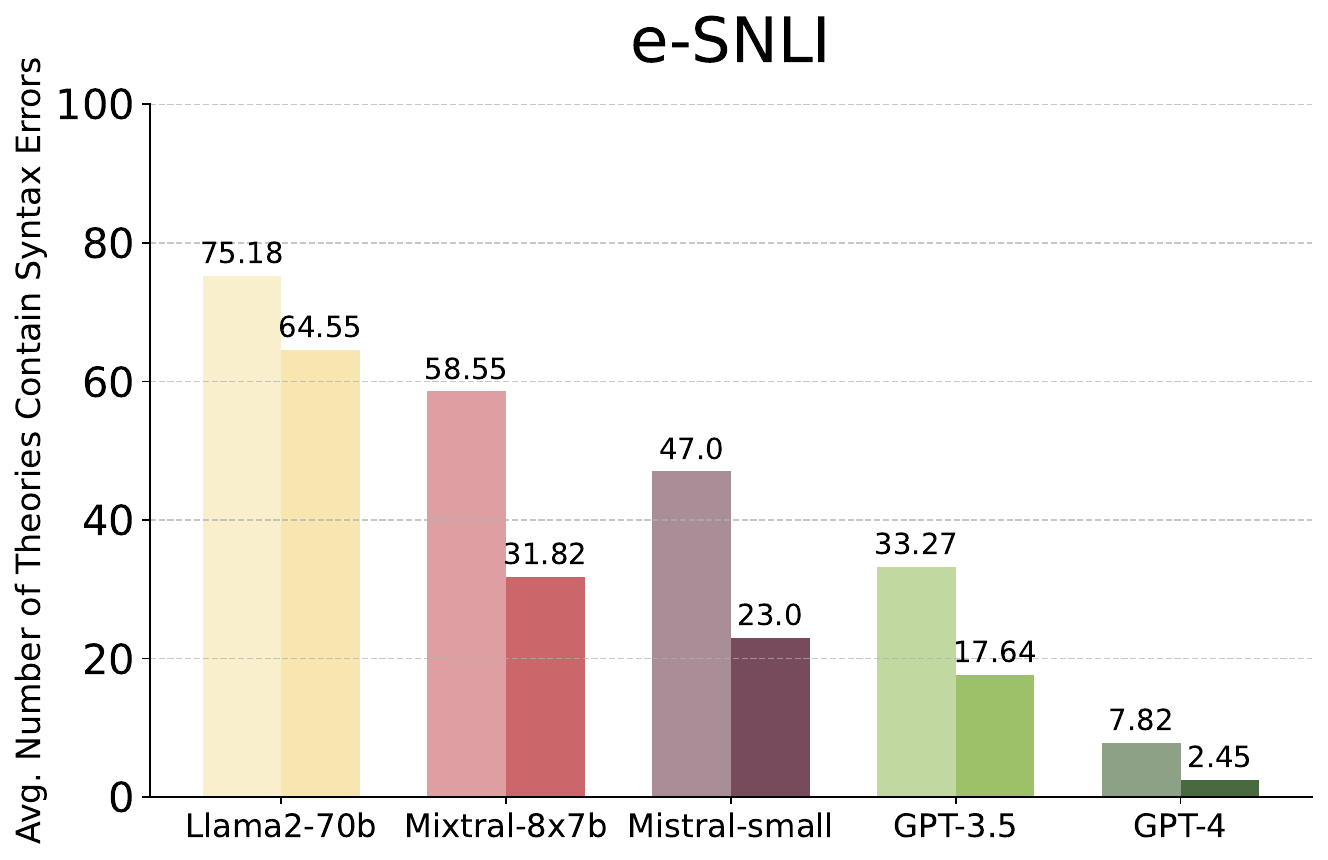}
  \caption{}
  \label{fig:sub1}
\end{subfigure}
\hfill
\begin{subfigure}{.3\textwidth} 
  \centering
  \includegraphics[width=\linewidth]{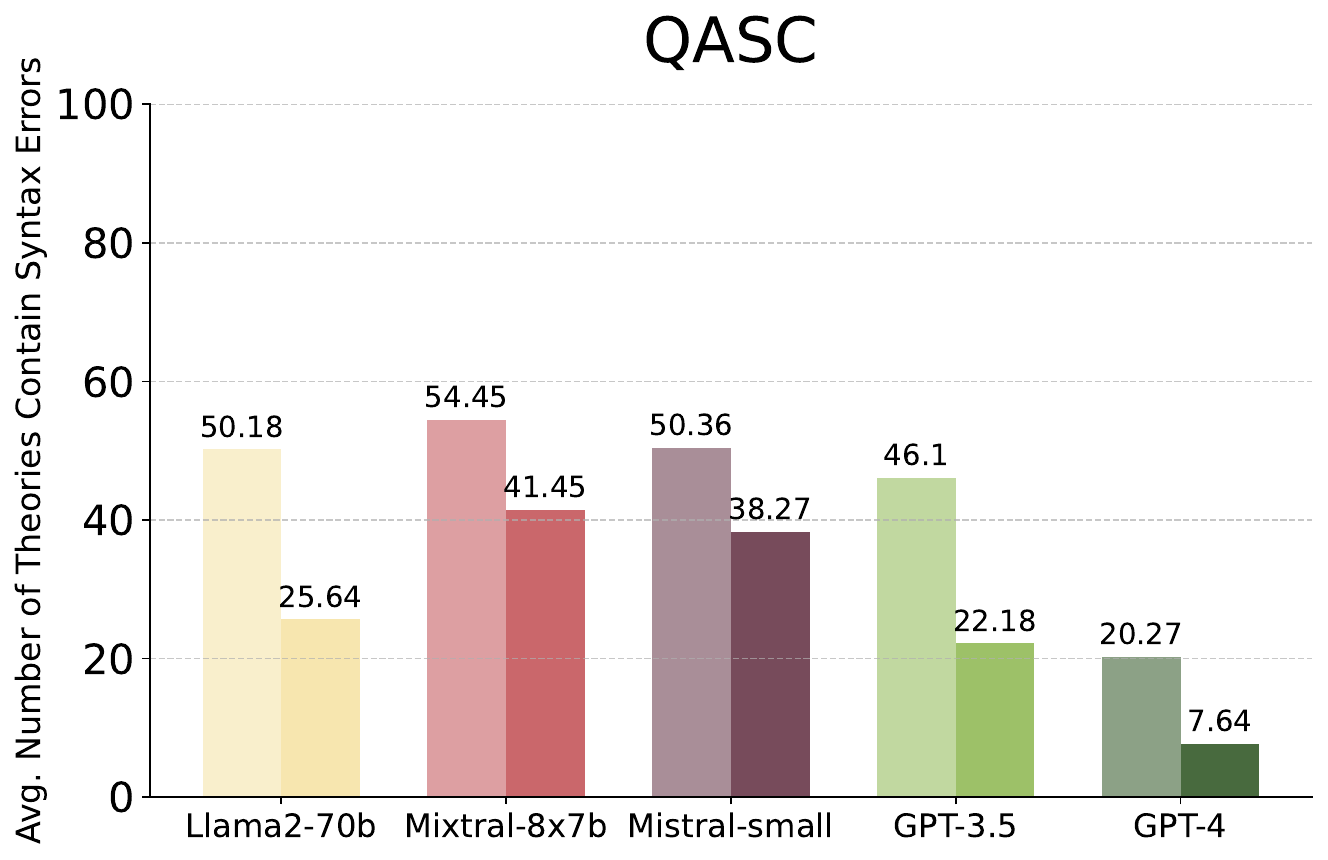}
  \caption{}
  \label{fig:sub2}
\end{subfigure}
\hfill
\begin{subfigure}{.3\textwidth} 
  \centering
  \includegraphics[width=\linewidth]{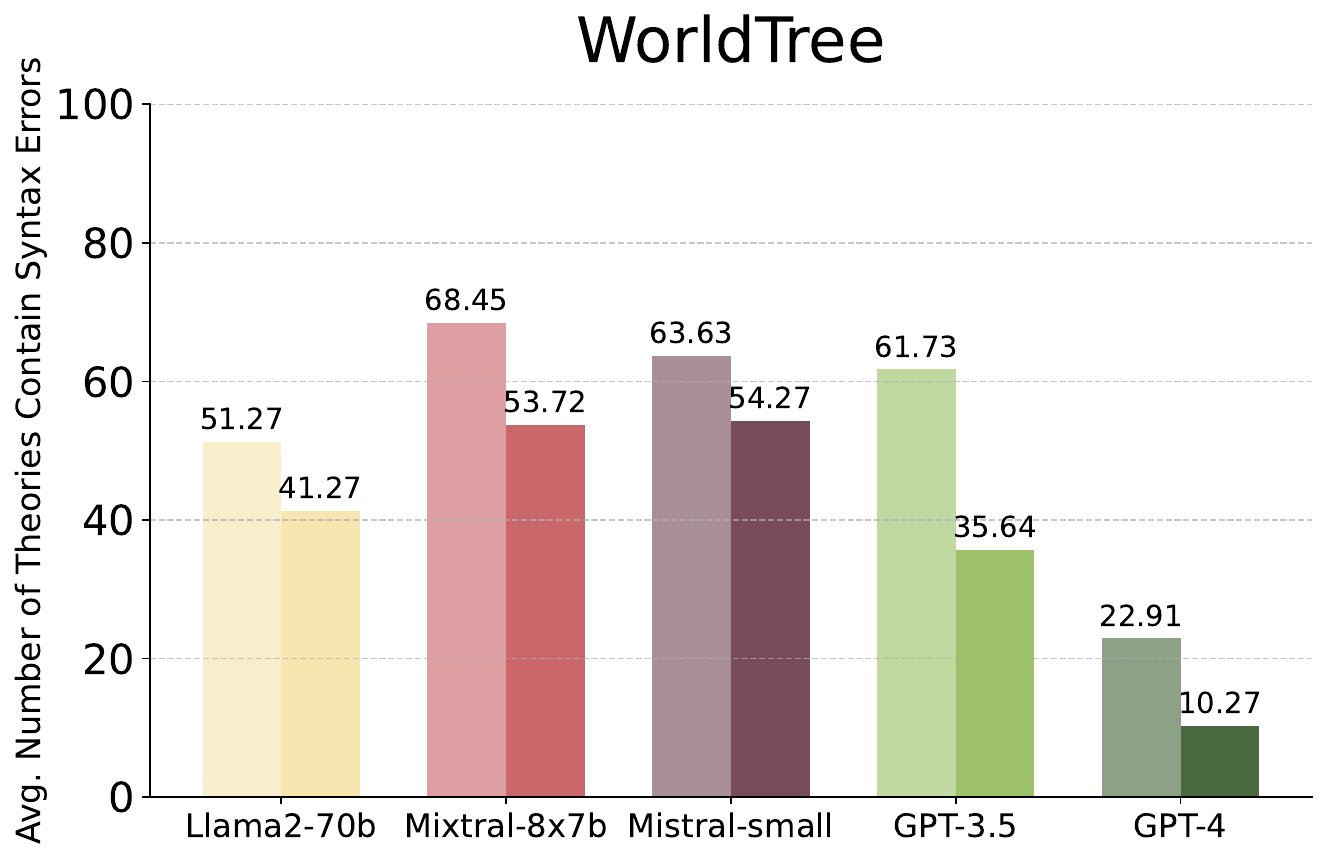}
  \caption{}
\end{subfigure}

\caption{The average number of theories containing syntactic errors before and after the syntax refinement process}
\label{fig:syntax_error_graph}
\end{figure*}

\begin{figure*}[t]
\centering
\begin{subfigure}{.3\textwidth} 
  \centering
  \includegraphics[width=\linewidth]{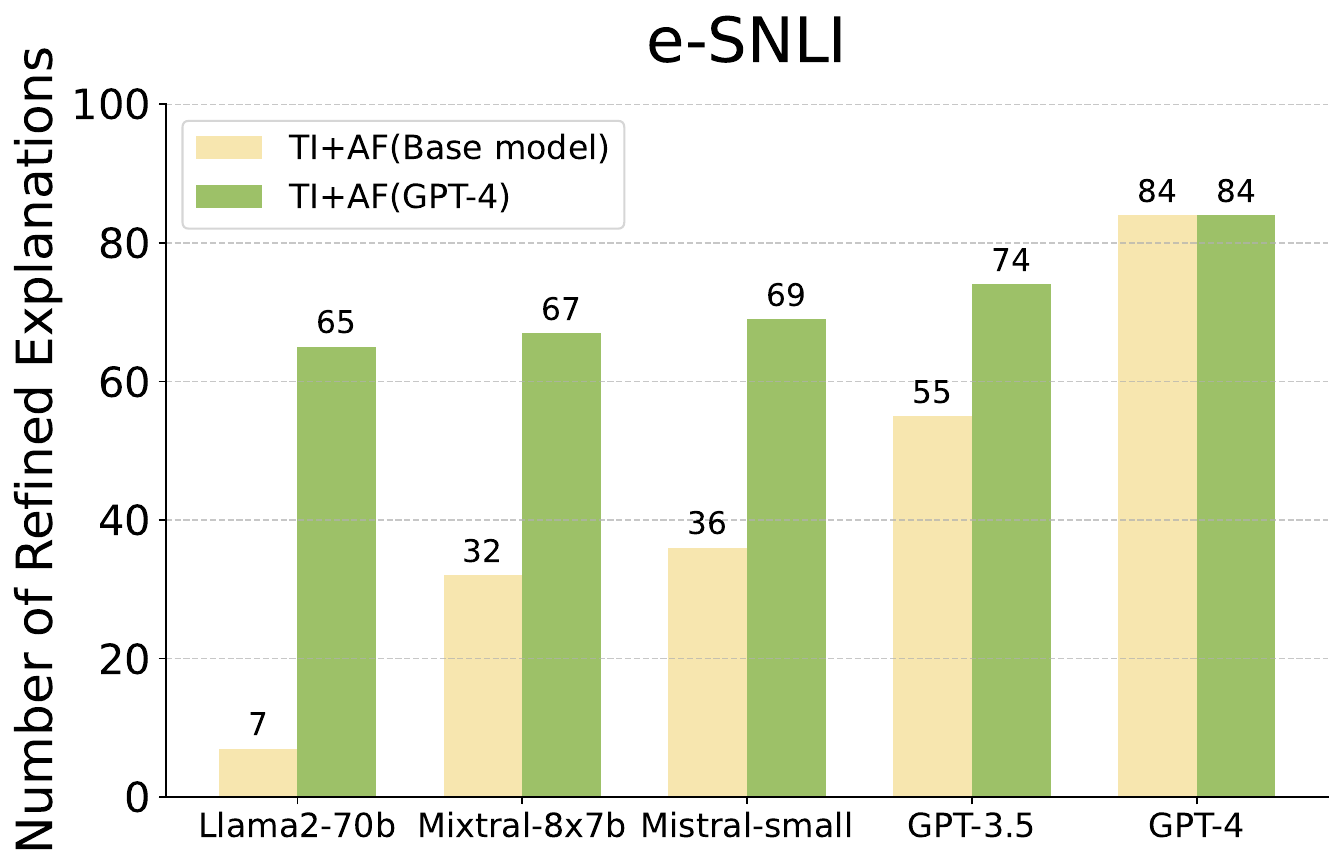}
  \caption{}
  \label{fig:sub1}
\end{subfigure}
\hfill
\begin{subfigure}{.3\textwidth} 
  \centering
  \includegraphics[width=\linewidth]{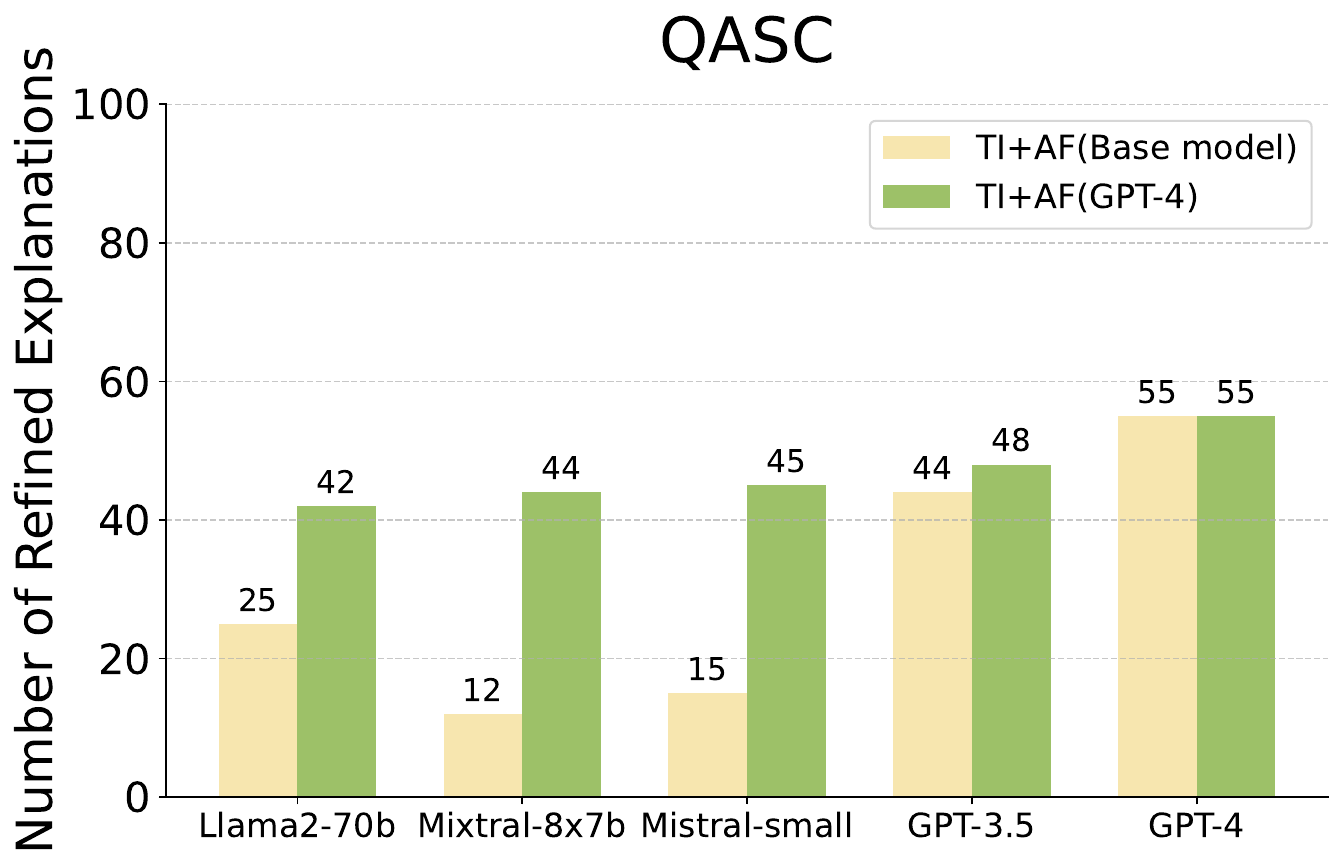}
  \caption{}
  \label{fig:sub2}
\end{subfigure}
\hfill
\begin{subfigure}{.3\textwidth} 
  \centering
  \includegraphics[width=\linewidth]{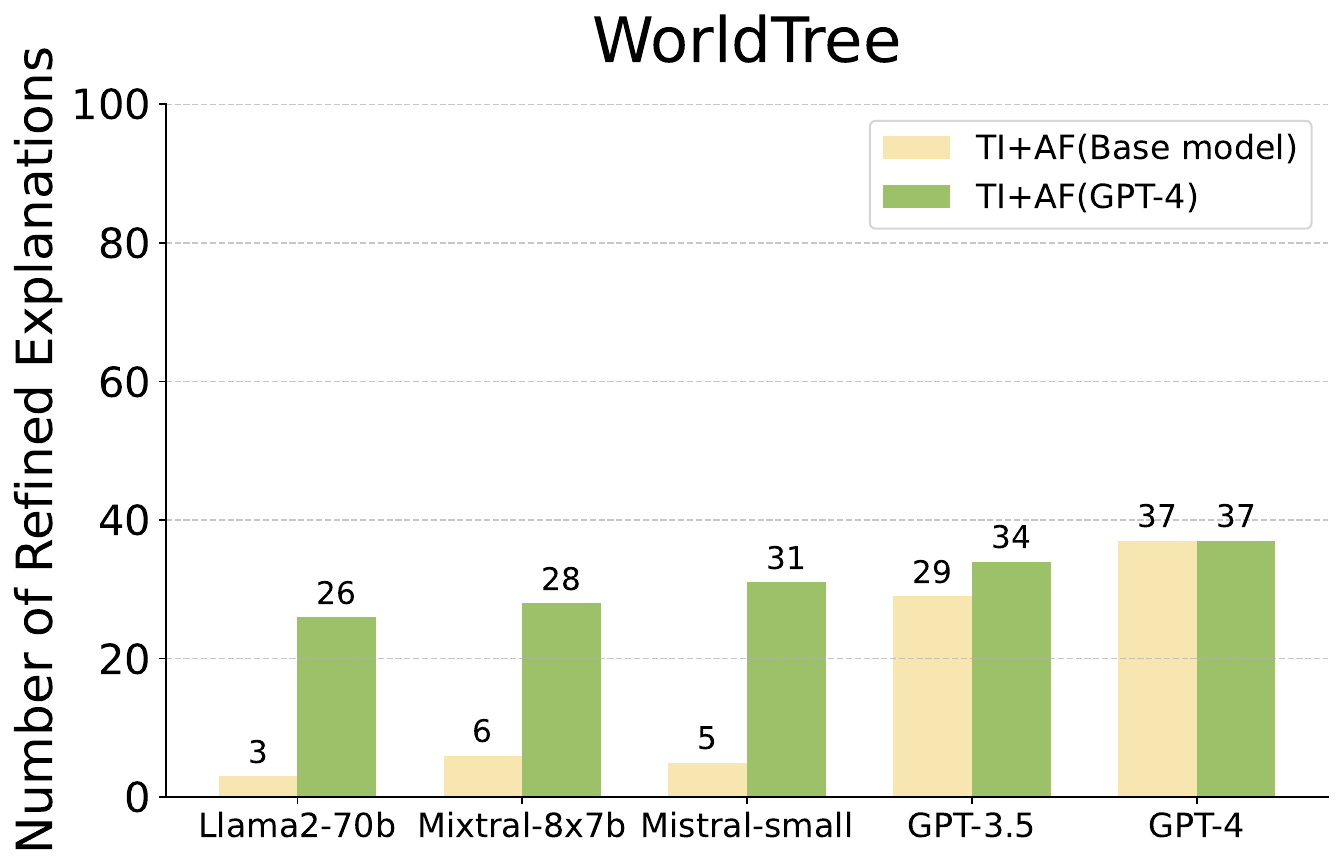}
  \caption{}
\end{subfigure}
\caption{AF represents the autoformalisation components, and TI represents the textual inference components. TI+AF (Base Model) indicates the use of the base model for both the autoformalisation and textual inference components. TI+AF (GPT-4) indicates the use of GPT-4 for the autoformalisation components, while the base model is used for textual inference.}
\label{fig:ablation_bar_chart}
\end{figure*}

\begin{figure*}[t]
\centering
\begin{subfigure}{.32\textwidth}
  \centering
  \includegraphics[width=\linewidth]{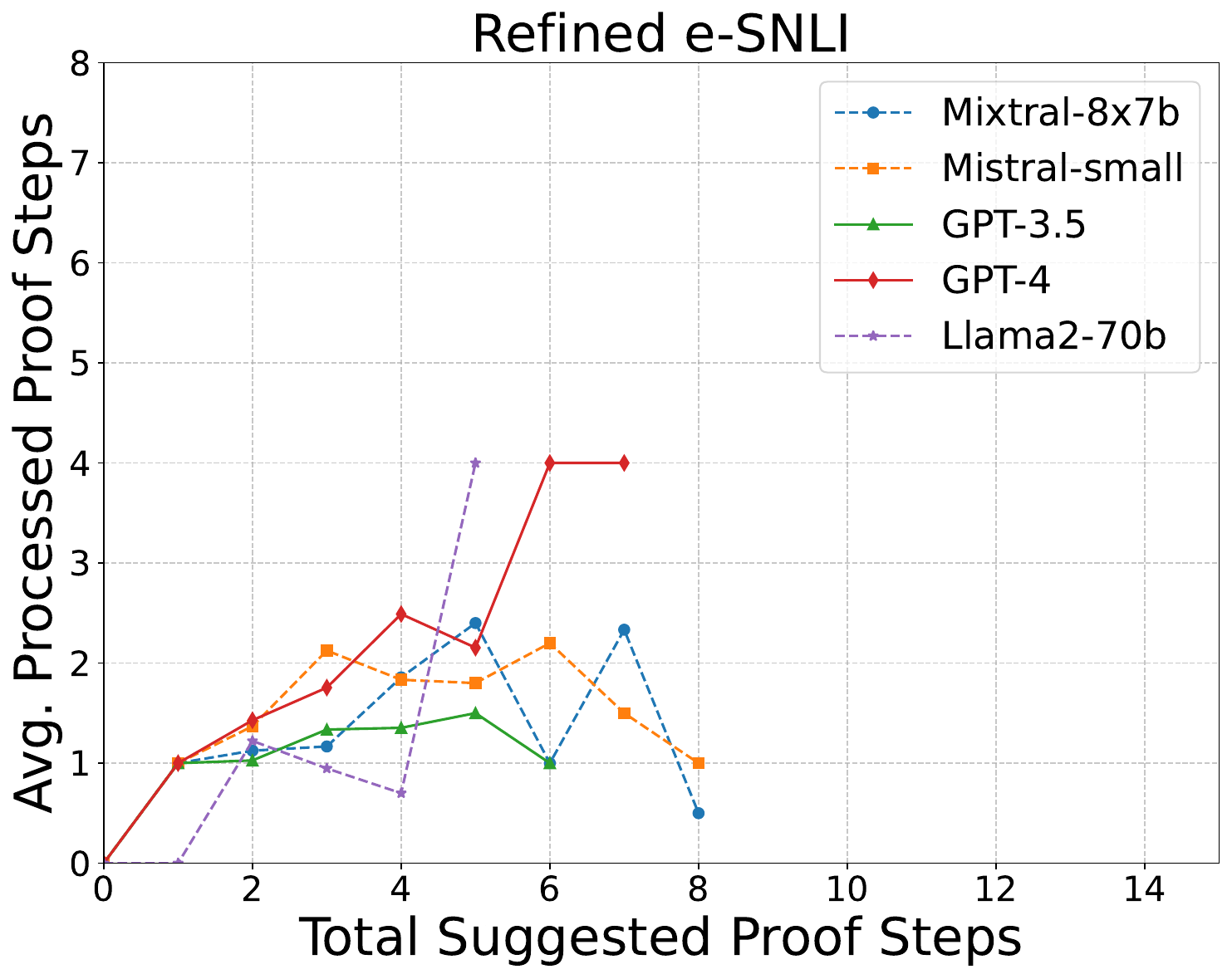}
  \caption{}
  \label{fig:sub1}
\end{subfigure}
\begin{subfigure}{.32\textwidth}
  \centering
  \includegraphics[width=\linewidth]{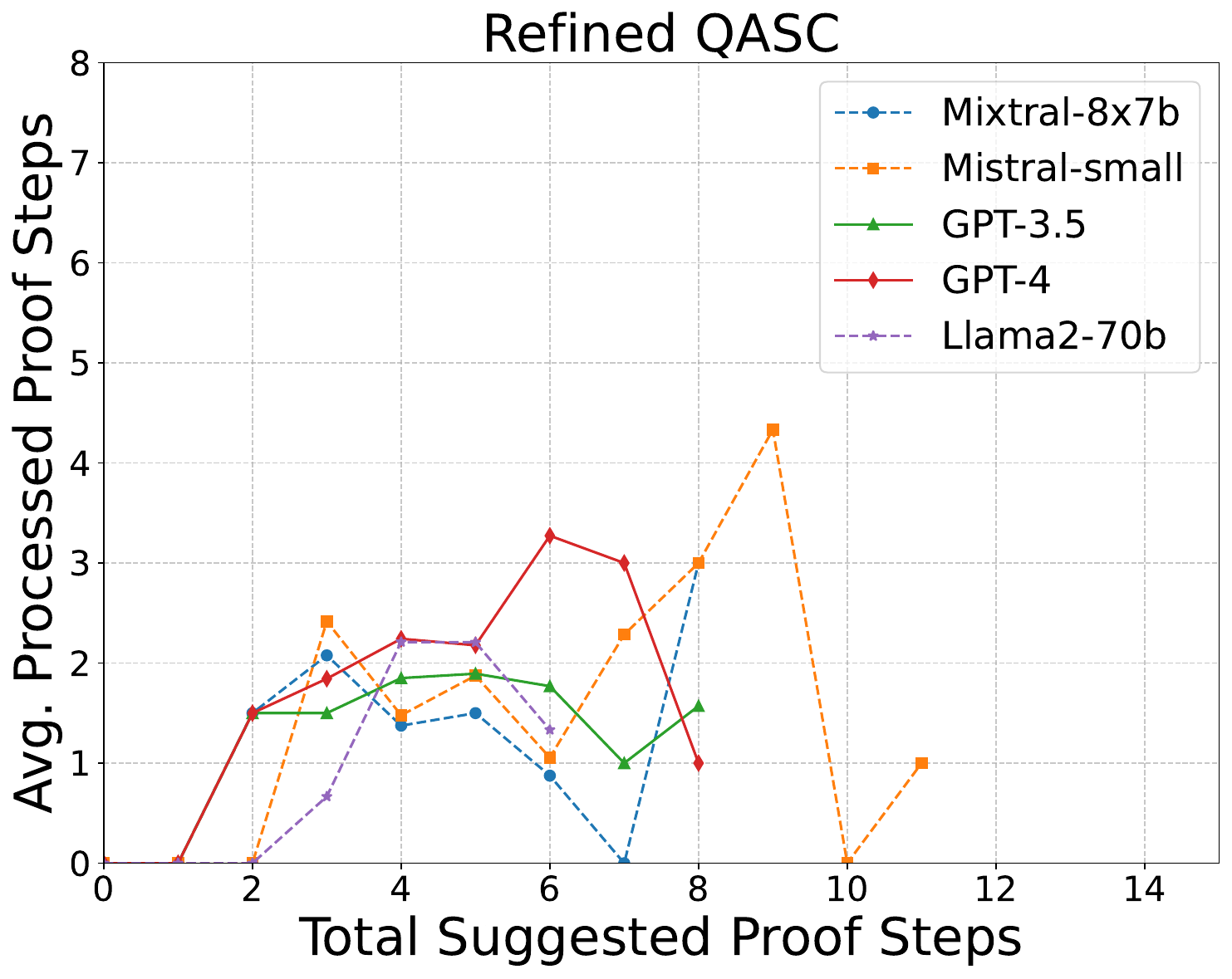}
  \caption{}
  \label{fig:sub2}
\end{subfigure}
\begin{subfigure}{.32\textwidth}
  \centering
  \includegraphics[width=\linewidth]{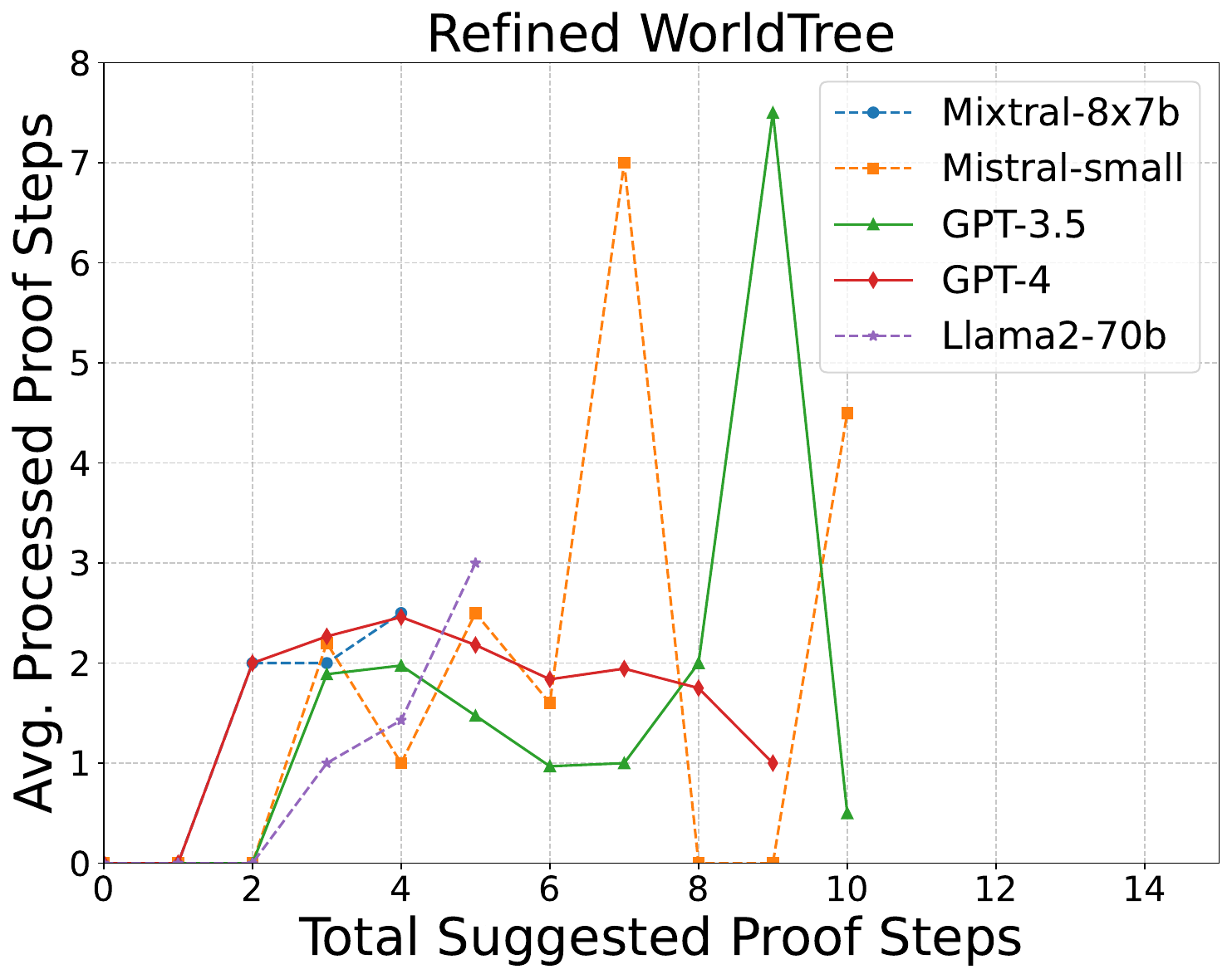}
  \caption{}
  \label{8_c}
\end{subfigure}

\begin{subfigure}{.32\textwidth}
  \centering
  \includegraphics[width=\linewidth]{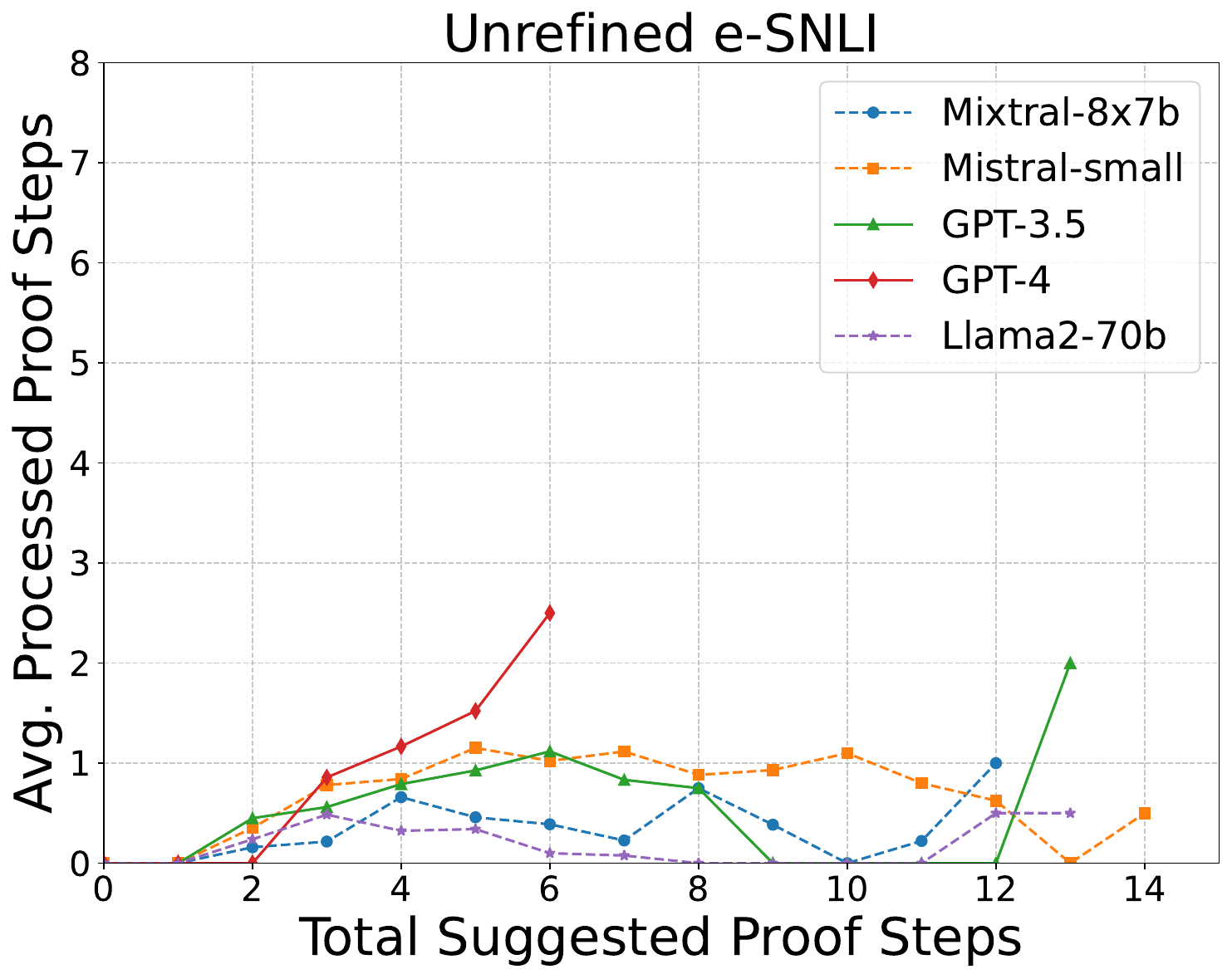}
  \caption{}
   \label{8_d}
\end{subfigure}
\begin{subfigure}{.32\textwidth}
  \centering
  \includegraphics[width=\linewidth]{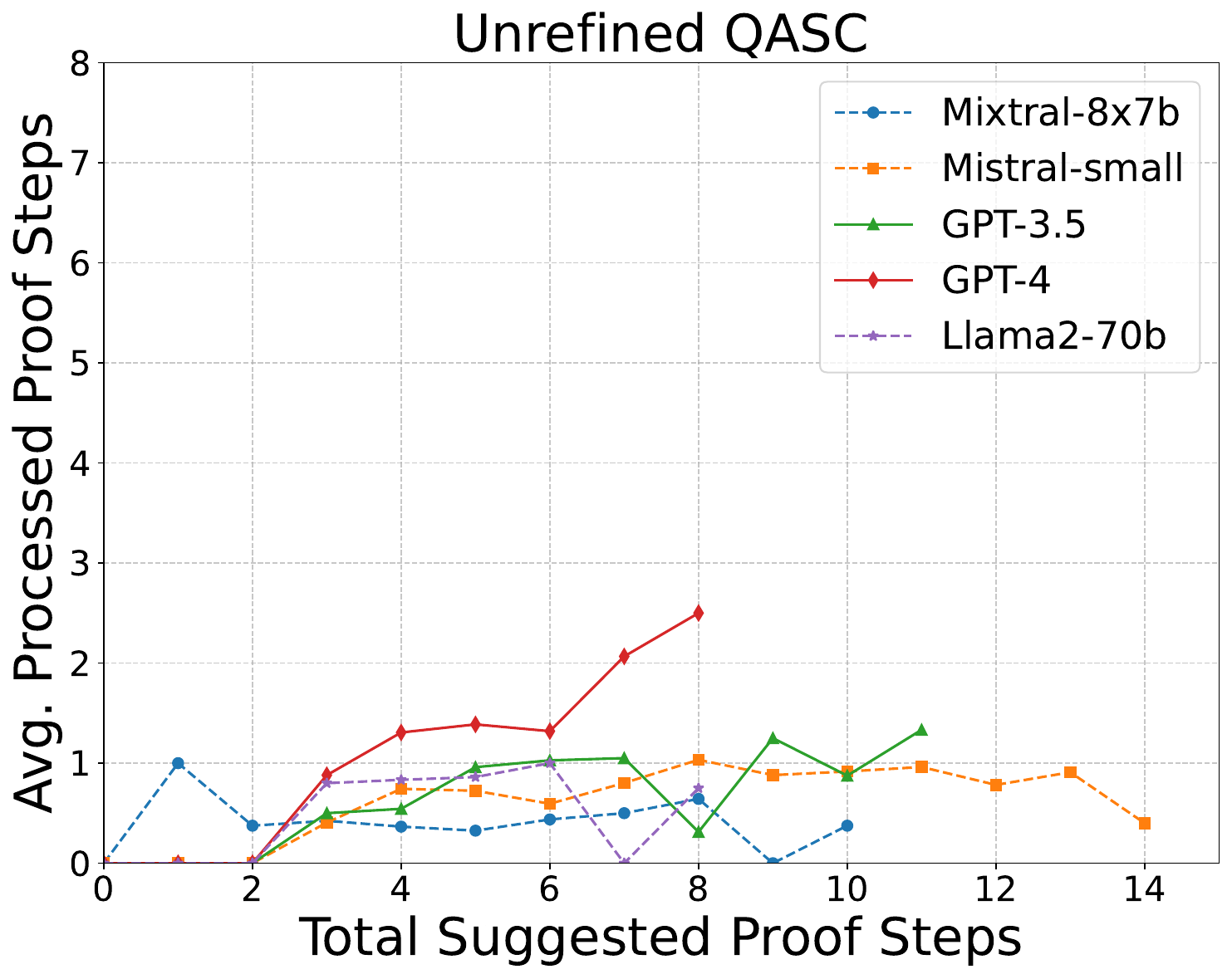}
  \caption{}
   \label{8_e}
\end{subfigure}
\begin{subfigure}{.32\textwidth}
  \centering
  \includegraphics[width=\linewidth]{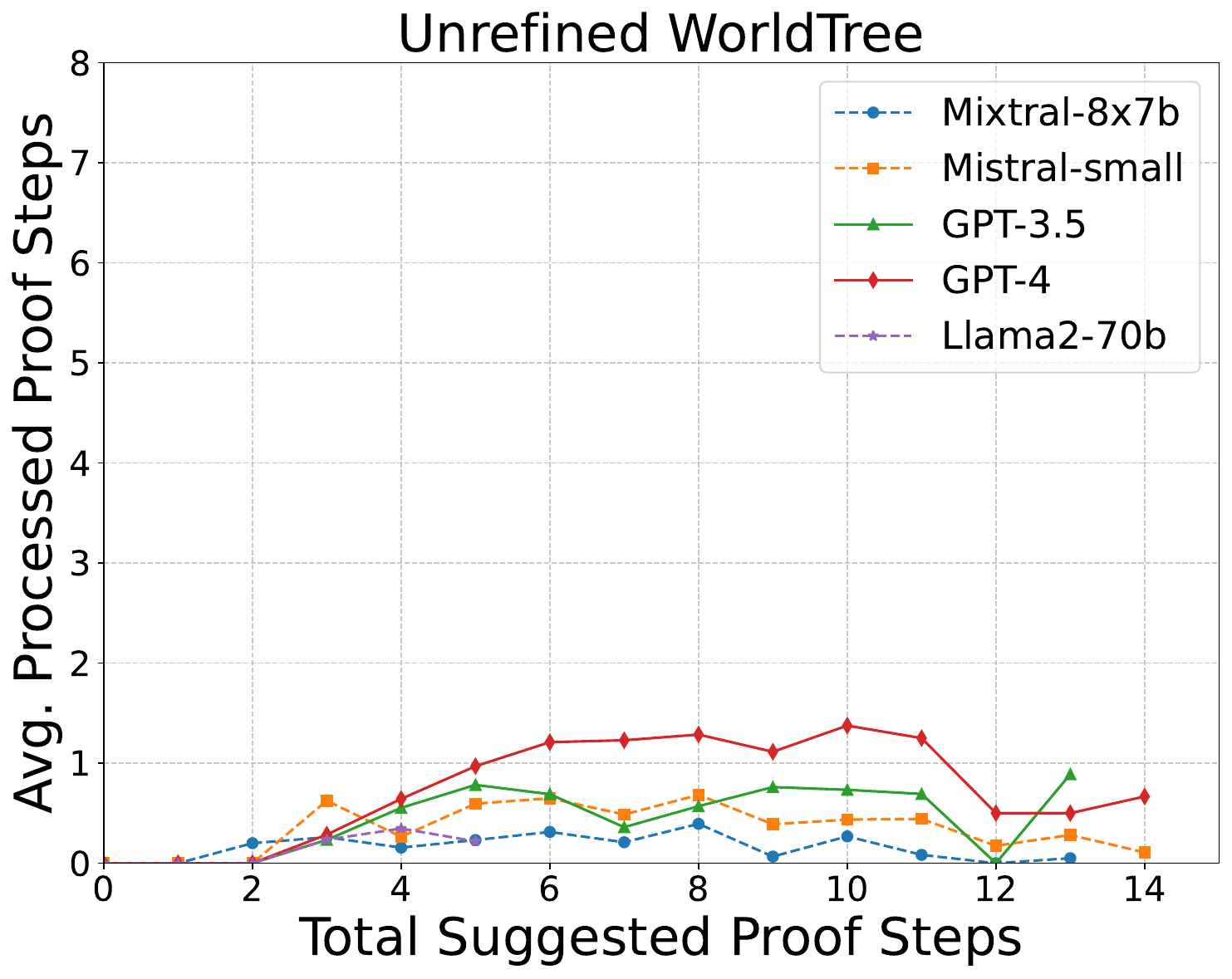}
  \caption{}
  \label{8_f}
\end{subfigure}
\caption{Average of proof steps processed by the proof assistant against the total proof steps suggested by the LLMs in refined and unrefined explanations.}
\label{fig:avg_processed_against_total}
\end{figure*}


\subsection{Datasets}
We adopted three different NLI datasets for evaluation: e-SNLI, QASC, and WorldTree, using a total of 300 samples selected via the sampling strategy defined in \citet{valentino-etal-2021-natural}, which maximises representativeness and mutual exclusivity across syntactic and semantic features expressed in the datasets. For multiple-choice question answering, the task includes a question $q$ accompanied by a set of candidate answers $C = \{c_1, c_2,...,c_n\}$, with $c_i$ identified as the correct answer. To cast this problem into NLI, we simply convert $q$ and the correct answer $c_i$ into a hypothesis $h_i$. On the other hand, the question's context, if present, is used to build the premise $p_i$.  


\subsection{Models}
To integrate Isabelle/HOL as a real-time verification tool with LLMs, we employ a Python client \citep{shminke2022python} which communicates with Isabelle/HOL as a server backend. This enables the communication of the constructed theory files and the extraction of the response messages from Isabelle. We conducted experiments using five LLMs within the proposed framework. The models include two open-sourced models: Llama2-70b \citep{touvron2023llama} and Mixtral-8x7b \citep{jiang2024mixtral}, as well as Mistral-small (mistral-small-latest) \citep{mistral_small_2024}, GPT-3.5 (gpt-3.5-turbo) \citep{NEURIPS2020_1457c0d6}, and GPT-4 (gpt-4-0613) \citep{DBLP:journals/corr/abs-2303-08774}.



\subsection{Results}
\paragraph{Detailed feedback from an external theorem prover effectively guides LLMs in verifying and refining explanations for NLI.}
To assess the effectiveness of employing an external theorem prover to verify and refine explanations in NLI tasks, we conducted a comparative analysis across various LLMs (Figure \ref{fig:overall_figure_bar_chart}). 
The initially valid explanations represent the percentage of explanations that can be verified as logically valid without any further iteration. Although the initial verification results varied among different models, all LLMs demonstrated a consistent improvement in refining the logical validity of the explanations. This process highlights the positive impact of the external feedback but also shows significant differences between models. We found that lower rates of initial valid explanations often resulted from syntactic errors, which impeded the theorem prover's ability to generate proofs. Despite this initial variability, all models demonstrate a consistent improvement in the refinement process across the datasets. Notably, GPT-4 outperformed other models, improving the validity of explanations by 48\%, 43\%, and 35\% across the three datasets, respectively, within a maximum number of ten iterations (Figure \ref{fig:overall_figure_bar_chart}). Figure \ref{fig:overall_figure} shows the number of explanations refined at each iteration across the e-SNLI, QASC, and WorldTree datasets. On average, we found that an increasing number of iterations leads to increasing refinement, with models requiring an average of five iterations across the datasets. 

\paragraph{Explanation length/complexity impacts formalisation and verification.}
The e-SNLI dataset, which includes only a single explanatory sentence per example, shows the best overall performance. In contrast, the multiple-choice question answering datasets, QASC and WorldTree, exhibit comparatively lower performance. QASC typically contains 2 explanatory sentences, while WorldTree ranges from 1 to 16 sentences. As the number of explanatory sentences increases, so does the complexity of the logical reasoning required. Models show lower refinement performance in WorldTree when compared to e-SNLI and QASC, with only 3\%, 5\%, and 5\% of Llama-70b, Mixtral-8x7b, and Mistral-small explanations being refined in WorldTree. Meanwhile, 29\% and 35\% of explanations are refined by GPT-3.5 and GPT-4 in WorldTree, respectively. This process involves synthesising multiple explanatory sentences to fulfill sub-goals, which must then be integrated to meet the overall hypothesis goal.

\paragraph{Iterative and categorical refinement can monotonically reduce syntactic errors in autoformalisation.}
To evaluate the syntax error refinement stage, we quantified the presence of syntax errors in the Isabelle theories both before and after the iterative refinement process. After a maximum of three iterations, all models showed significant reductions, with maximum reductions of 68.67\%, 62.31\%, and 55.17\% from 7.82 to 2.45, 20.27 to 7.64, and 22.91 to 10.27 across the three respective datasets (see Figure \ref{fig:syntax_error_graph}). While models like Llama2-70b and Mixtral-8x7b still exhibit some syntax errors in the refined theories' code, this is primarily due to their inability to perform complex autoformalisation, especially for multiple and more complex explanatory sentences such as those in the WorldTree dataset. This result is consistent with the percentage of explanations that were successfully refined across the models, which suggests that the autoformalisation process plays a critical role in the models' logical reasoning capability.


\begin{figure*}[t]
\centering
\begin{subfigure}{.32\textwidth} 
  \centering
  \includegraphics[width=\linewidth]{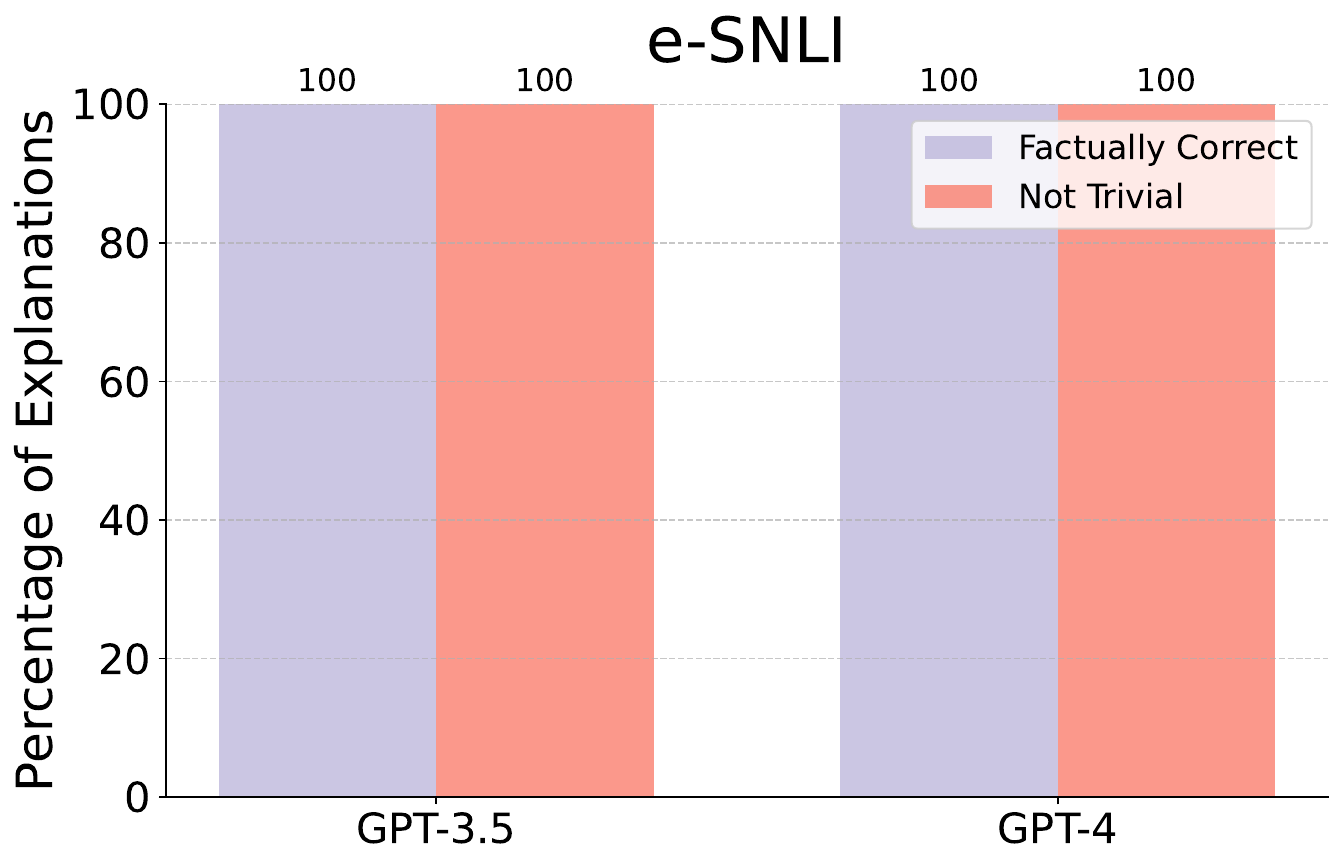}
  \caption{}
  \label{fig:sub1}
\end{subfigure}
\hfill
\begin{subfigure}{.32\textwidth} 
  \centering
  \includegraphics[width=\linewidth]{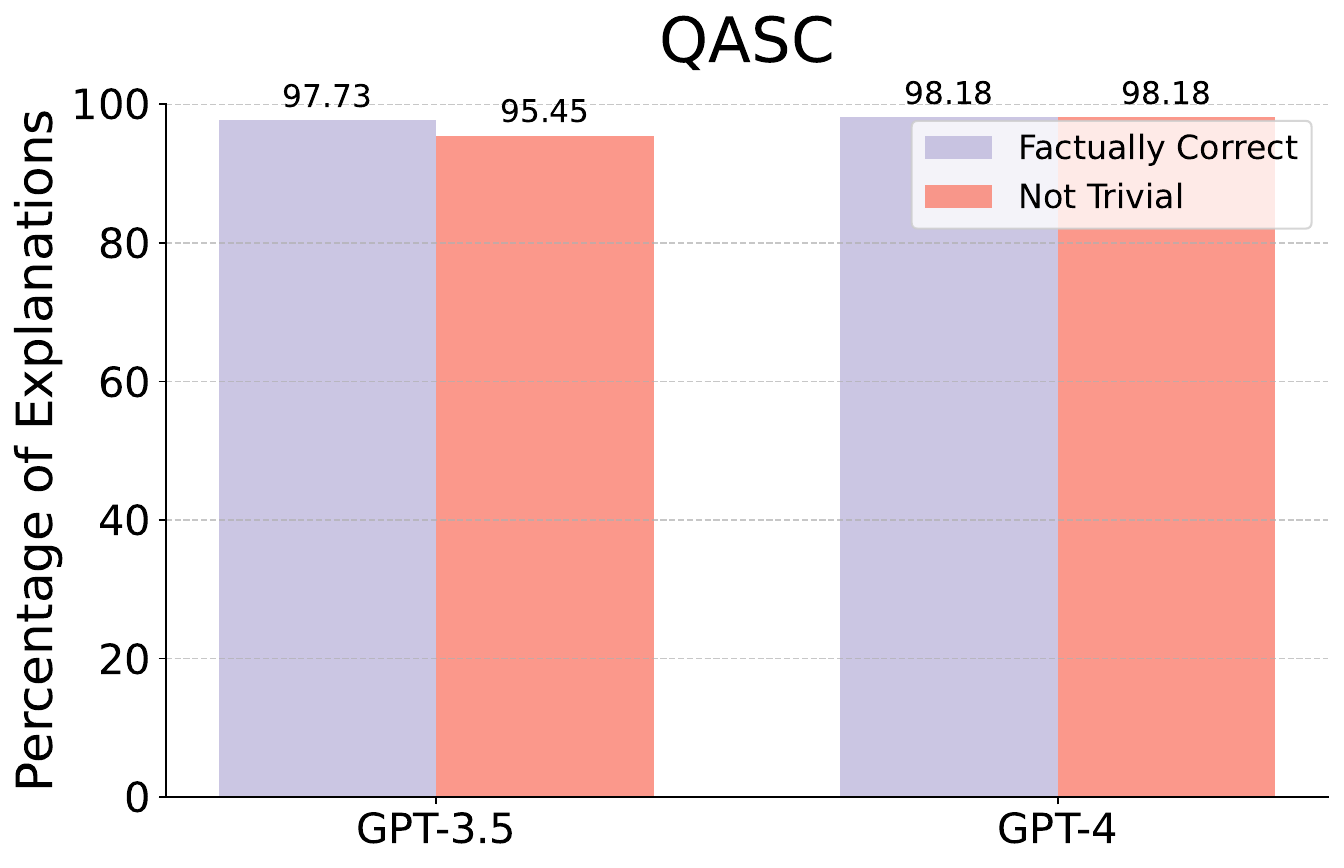}
  \caption{}
  \label{fig:sub2}
\end{subfigure}
\hfill
\begin{subfigure}{.32\textwidth} 
  \centering
  \includegraphics[width=\linewidth]{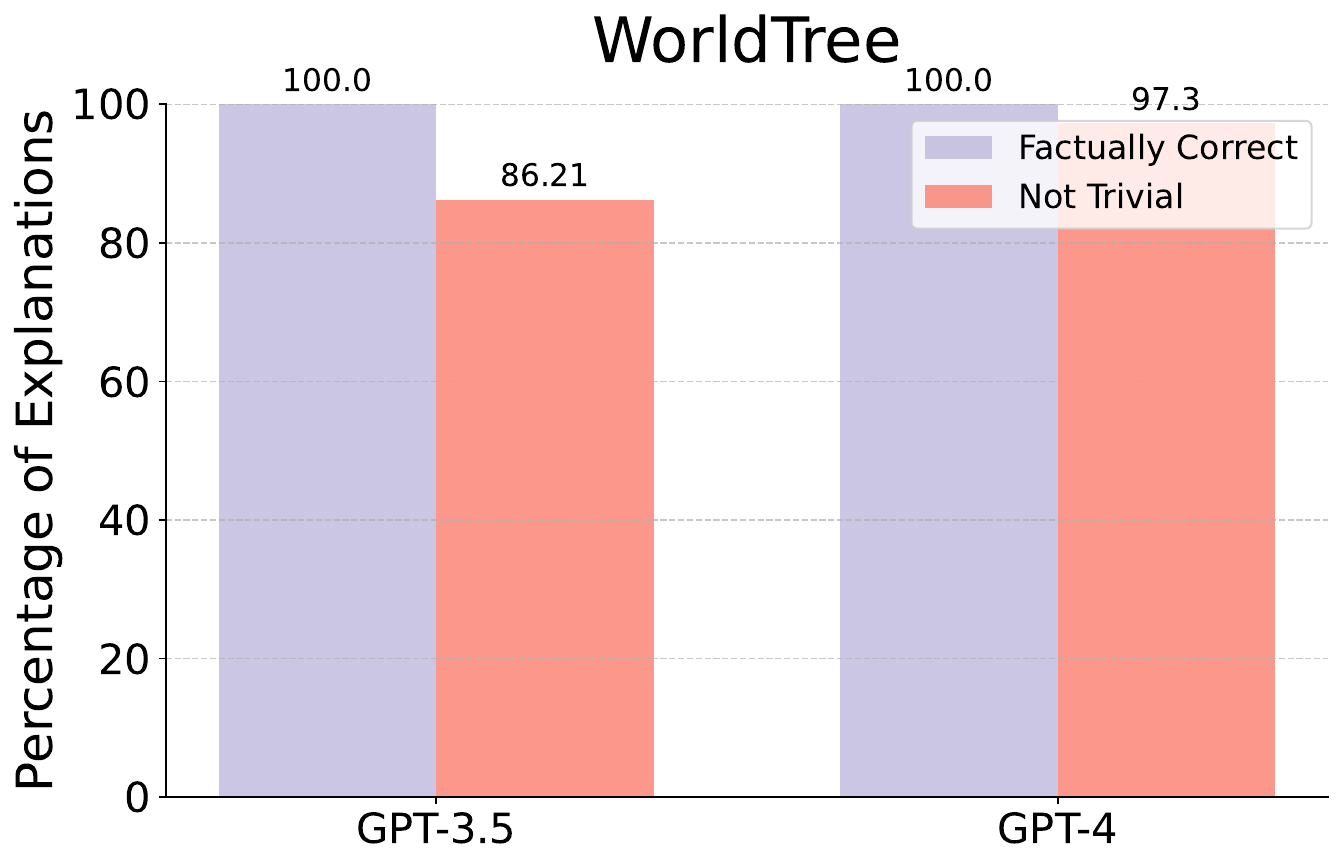}
  \caption{}
\end{subfigure}
\caption{Human evaluation of refined explanations in terms of factuality and triviality.}
\label{fig:factual_analysis}
\end{figure*}

\subsection{Ablation Study}

We conducted an ablation study to further evaluate and disentangle the impact of autoformalisation on performance. To this end, we adopted GPT-4 exclusively for the autoformalisation component, while retaining the original models for explanation refinement and proof strategy generation. As shown in Figure \ref{fig:ablation_bar_chart}, integrating GPT-4 for autoformalisation led to a significant increase in the number of explanations successfully refined across all models. For instance, Llama2-70b with GPT-4 as the formalisation component refined explanations from 7\% to 65\% in the e-SNLI dataset. For the multiple-choice question answering dataset, GPT-3.5 showed a relatively smaller increase from 44\% to 48\% and from 29\% to 34\%. Despite these improvements, a performance gap persists between GPT-4 and the other models, which is attributed to GPT-4's superior symbolic reasoning capabilities required for explanation refinement from the identified logical errors.

\paragraph{Explanations are progressively made more complete and consistent through iterative refinement.}
In order to deliver step-wise logical consistency, explanations need to be made complete and self-contained, leading to the introduction of additional explanatory sentences, which increases the total number of suggested proof steps. Therefore, we further evaluated how the proof steps vary when the total number of suggested proof steps increases, contrasting both refined and unrefined cases. Figure \ref{fig:avg_processed_against_total} illustrates this trend. In general, all models show a positive trend, as the total suggested proof steps increase, the average number of proof steps processed by the proof assistant also increases. Models like Mistral-small and GPT-3.5 tend to suggest more proof steps to accomplish the logical goal, which can result in some redundant steps, such as the significant pulse shown in Figure \ref{8_c}. For unrefined explanations, as shown in Figure \ref{8_d}, \ref{8_e} and \ref{8_f}, the progression is steadier but retains a positive trend, where the models generally suggest more proof steps in response to the additional explanatory sentences introduced to correct a logical error identified from the erroneous step. We analysed the correlation between average successful explanatory sentences and total planned sentences in proofs, detailed in Appendix \ref{appendix_pro_against_plan}. Examples of refined and unrefined explanations are in Appendix \ref{examples_refine_unrefine}.



\subsection{Factual Errors and Trivial Explanations}
In addition to evaluating the logical validity of explanations, we also conducted a human evaluation of the refined explanations, considering factual correctness and explanation triviality for the two best-performing models (GPT-3.5 and GPT-4). This evaluation focused on two questions: \textit{``Are the refined explanatory sentences factually correct?''} and \textit{``Is the explanation trivial, merely repeating or paraphrasing the content of the premise and hypothesis to achieve logical validity?''}. As illustrated in Figure \ref{fig:factual_analysis}, our findings indicate that all refined explanations in the e-SNLI and WorldTree datasets are consistent with commonsense knowledge. In the QASC dataset, 2.27\% and 1.82\% of the explanation refined by GPT-3.5 and GPT-4 contain sentences misaligned with true world knowledge. We found that the majority of these errors result from over-generalisation, such as the sentence \textit{All tetrapods are defined to have four limbs}, which inaccurately includes snakes.

Finally, we found a relatively low number of explanations that repeat or paraphrase the content of premise and hypothesis. This phenomenon is absent in e-SNLI and becomes more evident when the explanatory sentences increase in complexity (i.e., WorldTree), leading models sometimes to generate explanations that do not include any additional information for the entailment to hold.

\section{Related Work}
\subsection{LLMs Self-Refinement from External Feedback}
Self-refinement of LLMs has demonstrated promising effectiveness in generating faithful and trustworthy responses \citep{pan2023automatically}. The use of external feedback to guide LLMs has been extensively studied \citep{yu2023improving, akyurek-etal-2023-rl4f, olausson2024repair}. Previous work such as \citet{peng2023check} have employed facts retrieved from external knowledge bases as sources of feedback, while \citet{paul-etal-2024-refiner} developed a critic model to provide feedback for reasoning refinement. Additionally, \citet{nathani-etal-2023-maf} have explored the use of feedback models for automated feedback generation. Various works have also investigated tasks related to code generation \citep{chen2023teaching, olausson2024selfrepair} and the creation of either synthetic or expert-written logical natural language expressions \citep{olausson-etal-2023-linc}. \citet{quan-etal-2024-enhancing} use a differentiable logic reasoner for verifying and refining explanations via abductive reasoning, improving logical consistency in ethical NLI tasks. This paper focuses on the automated verification and refinement of natural language explanations created by human annotators in NLI tasks. Our method leverages feedback from external solvers to iteratively refine explanations, which require specific modelling interventions such as extracting the exact erroneous steps from the theorem prover to effectively refine logical errors in the explanatory sentences.

\subsection{Explanation Generation}
Existing work has explored robust and effective approaches for multi-hop reasoning tasks in explanation generation \citep{thayaparan-etal-2021-explainable, valentino-etal-2022-case, neves-ribeiro-etal-2022-entailment}. In prior research, metrics such as Mean Average Precision (MAP) \citep{valentino2022hybrid} have been employed to assess the ranking of facts in explanation generation tasks against gold-standard explanations. Although these metrics effectively measure precision relative to these standards, they inadequately capture the logical consistency and completeness of the explanations generated. Such shortcomings are particularly critical in tasks that require not only factual accuracy but also coherence and inferential soundness, as in natural language inference and explanation generation. Our proposed metrics address this gap by incorporating assessments of logical validity. Although some metrics have been proposed to manually evaluate the logical validity of explanations \citep{valentino-etal-2021-natural, yuan-etal-2024-logical}, such as non-redundancy or logical errors, these require significant effort from domain experts in formal languages. In this work, we use human-annotated explanations as a foundational dataset to detect and correct logical discrepancies, offering a framework adaptable for automatically enhancing both the precision and logical integrity of outputs across multi-step inference tasks.

\subsection{Autoformalisation}
Autoformalisation refers to the process of translating natural language descriptions into symbolic representations. Research in this area has included the formalisation of mathematical proofs \citep{cunningham-etal-2022-towards, NEURIPS2022_d0c6bc64, first2023baldur, jiang2023draft}, and efforts to transform natural language sentences into logical forms using LLMs \citep{pan-etal-2023-logic, olausson-etal-2023-linc, jiang2024leanreasoner, dalal2024inference}. However, contextual information is frequently lost when sentences are translated in these logical frameworks. To mitigate semantic loss during the transformation process, we leverage Neo-Davidsonian event semantics, which aims to maximise the preservation sentence-level content. This representation paradigm can facilitate a more systematic content-preserving translation to logical forms, which is more independent from particular choices of representation schema.

\section{Conclusion}
In this work, we present a novel neuro-symbolic framework, Explanation-Refiner, which integrates LLMs and theorem provers for automatic verification and refinement of natural language explanations through iterative cycles. Extensive experiments on textual entailment and multiple-choice QA tasks showed improved logical validity of human-annotated explanations. We investigated the model's performance from simple to complex explanatory/sentence structures and introduced a method to prevent the loss of semantic information in autoformalisation tasks with error correction. In future work, we aspire to enhance the framework's robustness towards complex and unstructured explanations with fewer iterations required to improve the model's efficiency.

\section*{Limitations}
While this work have demonstrated significant improvements in terms of enhancing the logical consistency of explanations, the connection between logical consistency and AI safety still needs further investigation. While the idea of using formal solvers in conjunction with LLMs delivers a promise avenue to improve the consistency of reasoning within LLMs, these methodologies need to be further developed and critically assessed as a mechanism which can provide guarantees of correctness, consistency and completeness within critical application domains. 

\section*{Acknowledgments}
This work was partially funded by the Swiss National Science Foundation (SNSF) project NeuMath (\href{https://data.snf.ch/grants/grant/204617}{200021\_204617}), by the EPSRC grant EP/T026995/1, “EnnCore: End-to-End Conceptual Guarding of Neural Architectures” under Security for all in an AI enabled society, by the CRUK National Biomarker Centre, and supported by the Manchester Experimental Cancer Medicine Centre and the NIHR Manchester Biomedical Research Centre.

\bibliography{anthology,custom}
\bibliographystyle{acl_natbib}

\appendix

\section{Appendix}
\label{sec:appendix}
\subsection{Algorithm}
Algorithm \ref{algorithm_1} shows the overall framework of Explanation-Refiner.

\subsection{Scalability}
\begin{figure*}[t]
\centering
\begin{subfigure}{.5\textwidth}
  \centering
  \includegraphics[width=\linewidth]{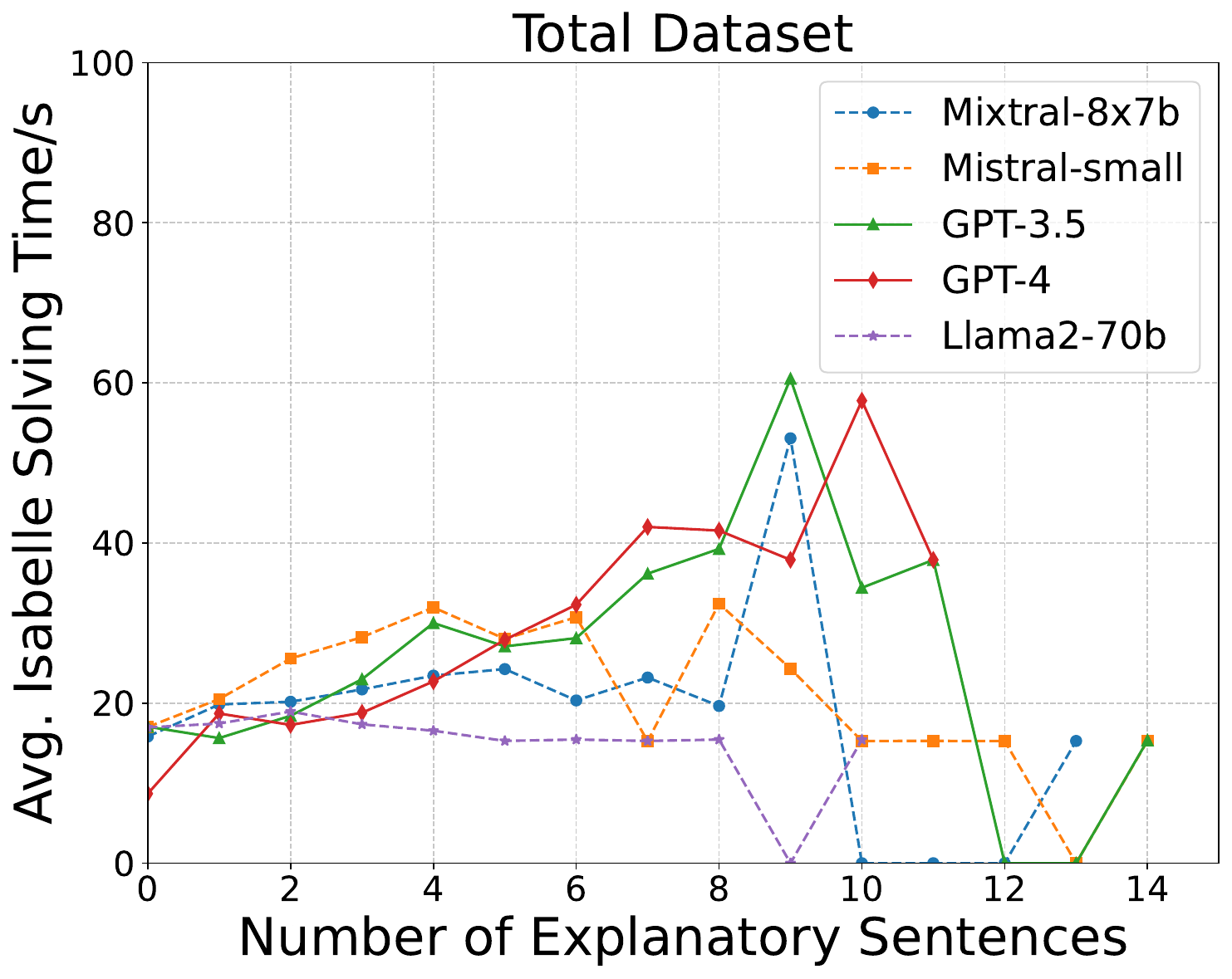}
  \caption{}
  \label{fig:sub1}
\end{subfigure}
\begin{subfigure}{.5\textwidth}
  \centering
  \includegraphics[width=\linewidth]{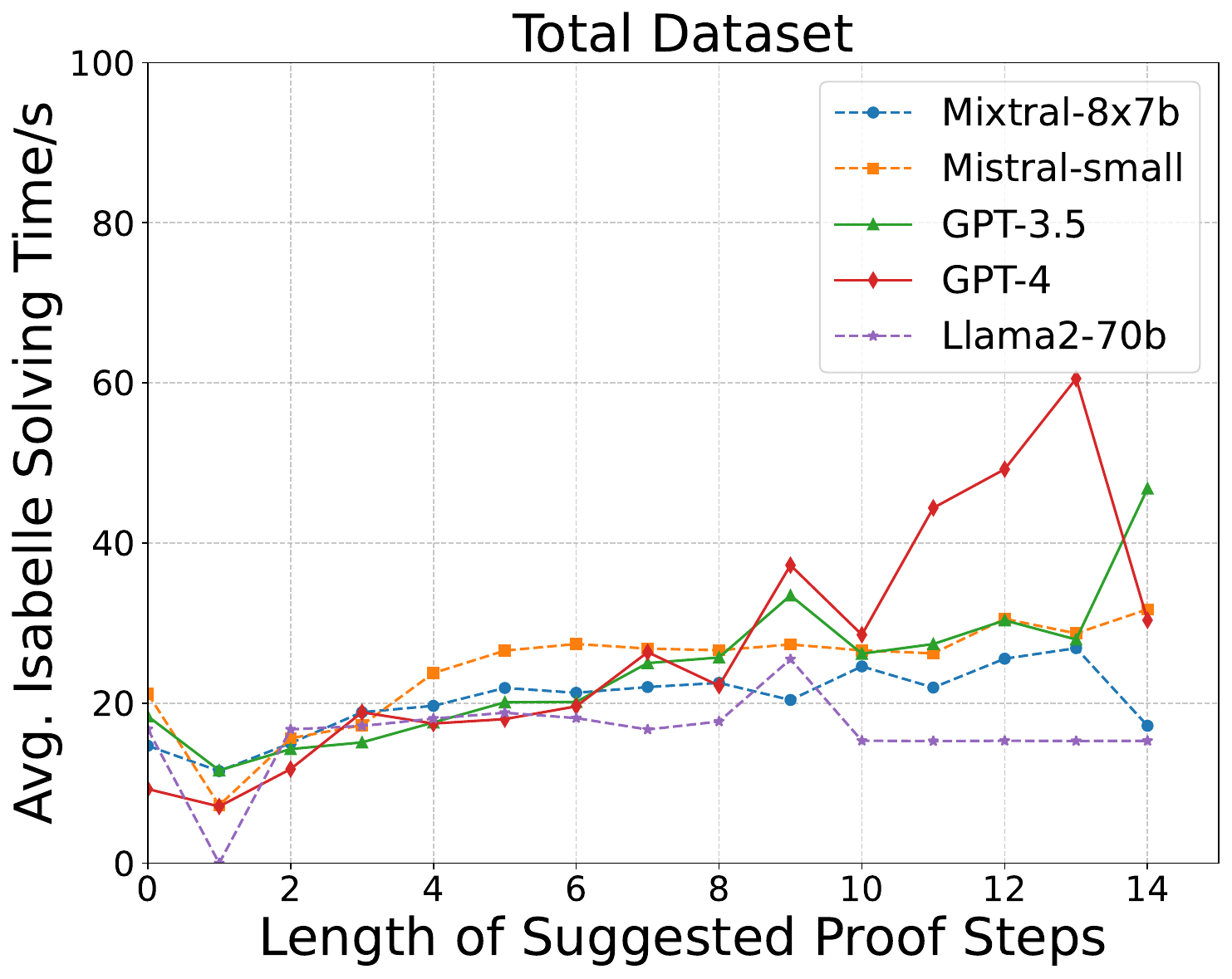}
  \caption{}
  \label{fig:sub2}
\end{subfigure}
\caption{(a) Average Isabelle/HOL solving time against number of explanatory sentences planned in a proof. (b) Average Isabelle/HOL solving time against number of suggested proof steps in a proof. }
\label{fig:scalability}
\end{figure*}

Figure \ref{fig:scalability} shows the average Isabelle/HOL solving time against the number of planned explanatory sentences in a proof and the length of suggested proof steps, including theories that have syntax errors, respectively. In some cases, the theorem prover may get stuck on a proof step, and we have set a termination time if the solving time exceeds 65 seconds.

\subsection{Average Processed vs. Planned Explanatory Sentences per Proof}
\label{appendix_pro_against_plan}
\begin{figure*}[htbp]
\centering
\begin{subfigure}{.32\textwidth}
  \centering
  \includegraphics[width=\linewidth]{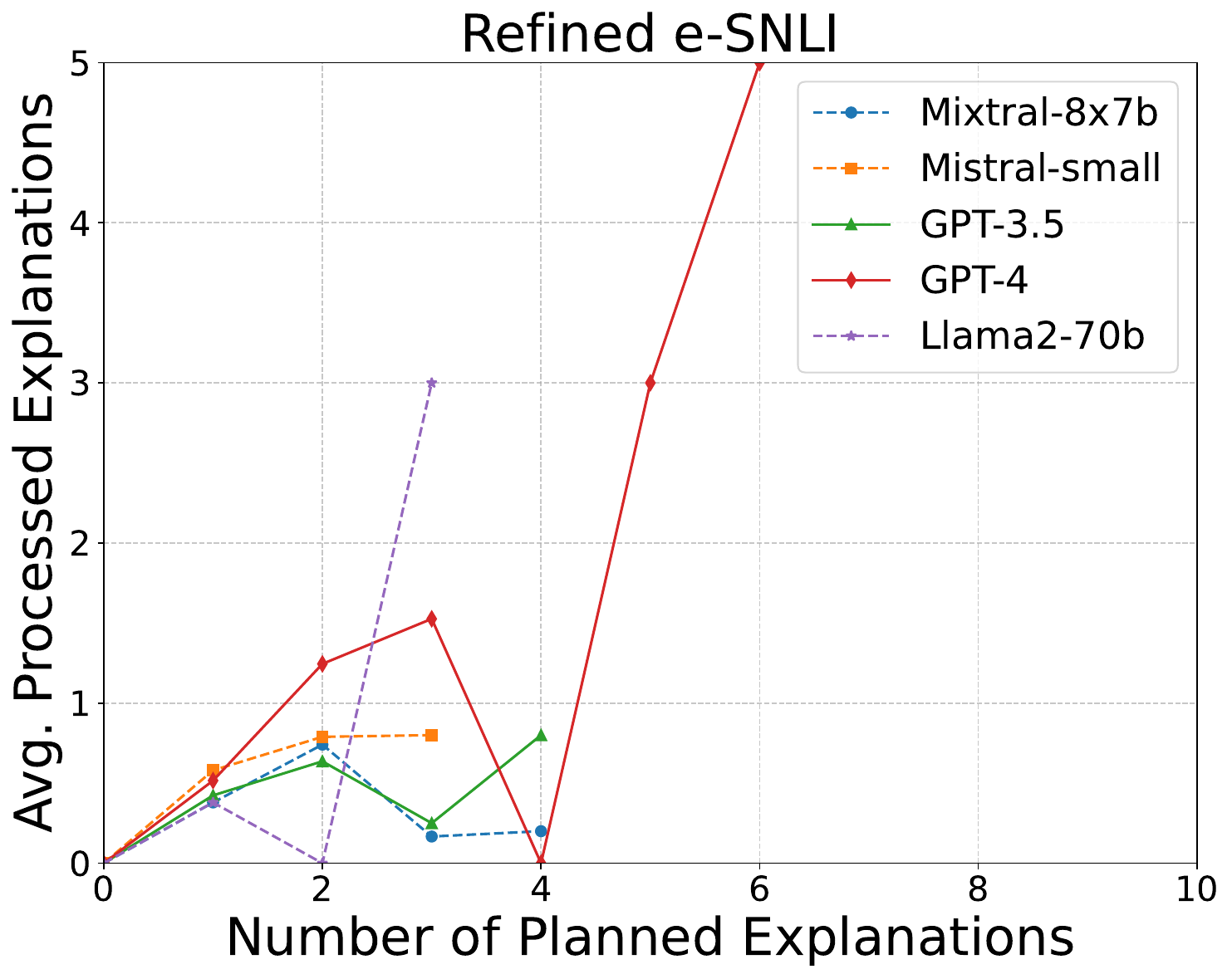}
  \caption{}
  \label{fig:sub1}
\end{subfigure}
\begin{subfigure}{.32\textwidth}
  \centering
  \includegraphics[width=\linewidth]{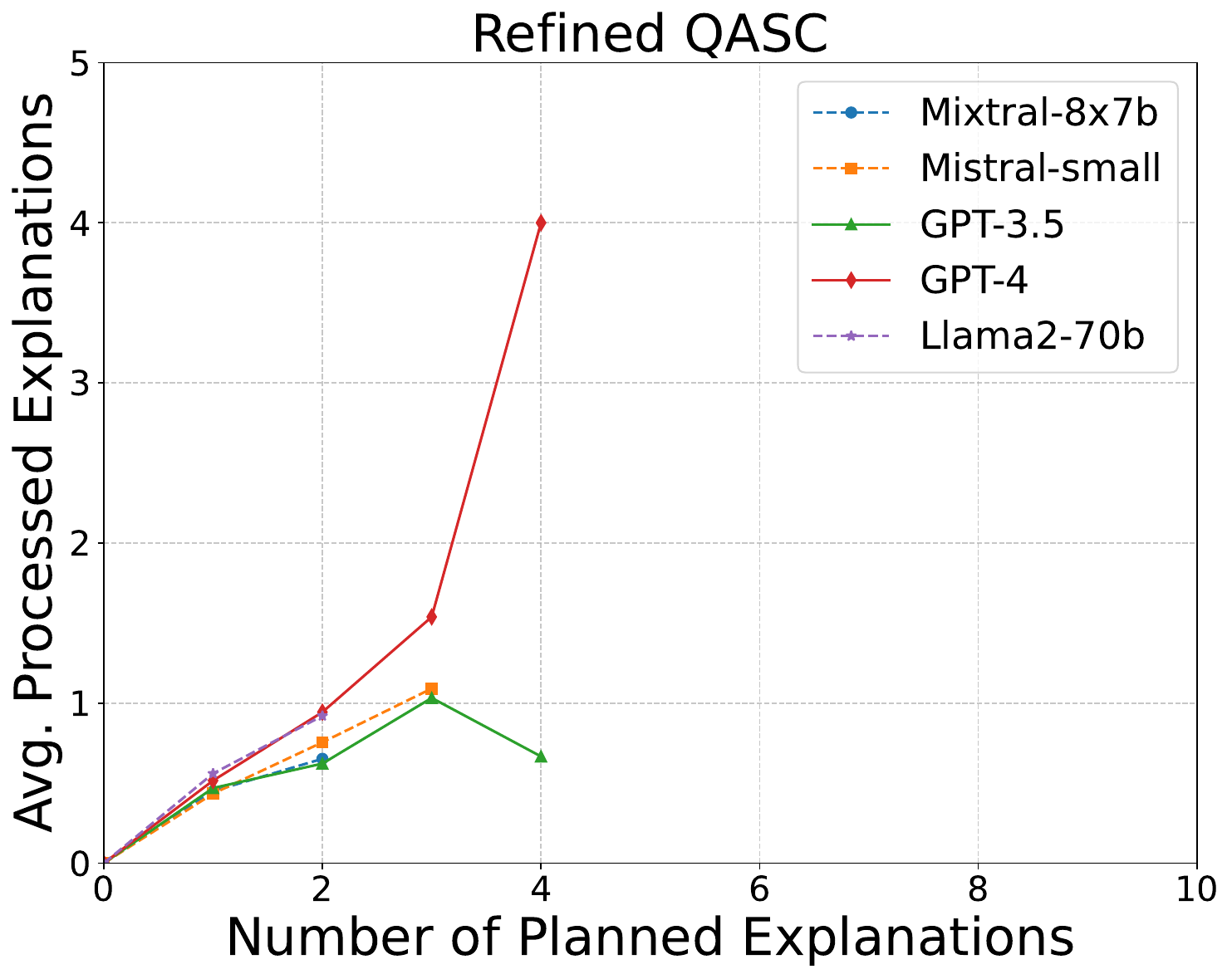}
  \caption{}
  \label{fig:sub2}
\end{subfigure}
\begin{subfigure}{.32\textwidth}
  \centering
  \includegraphics[width=\linewidth]{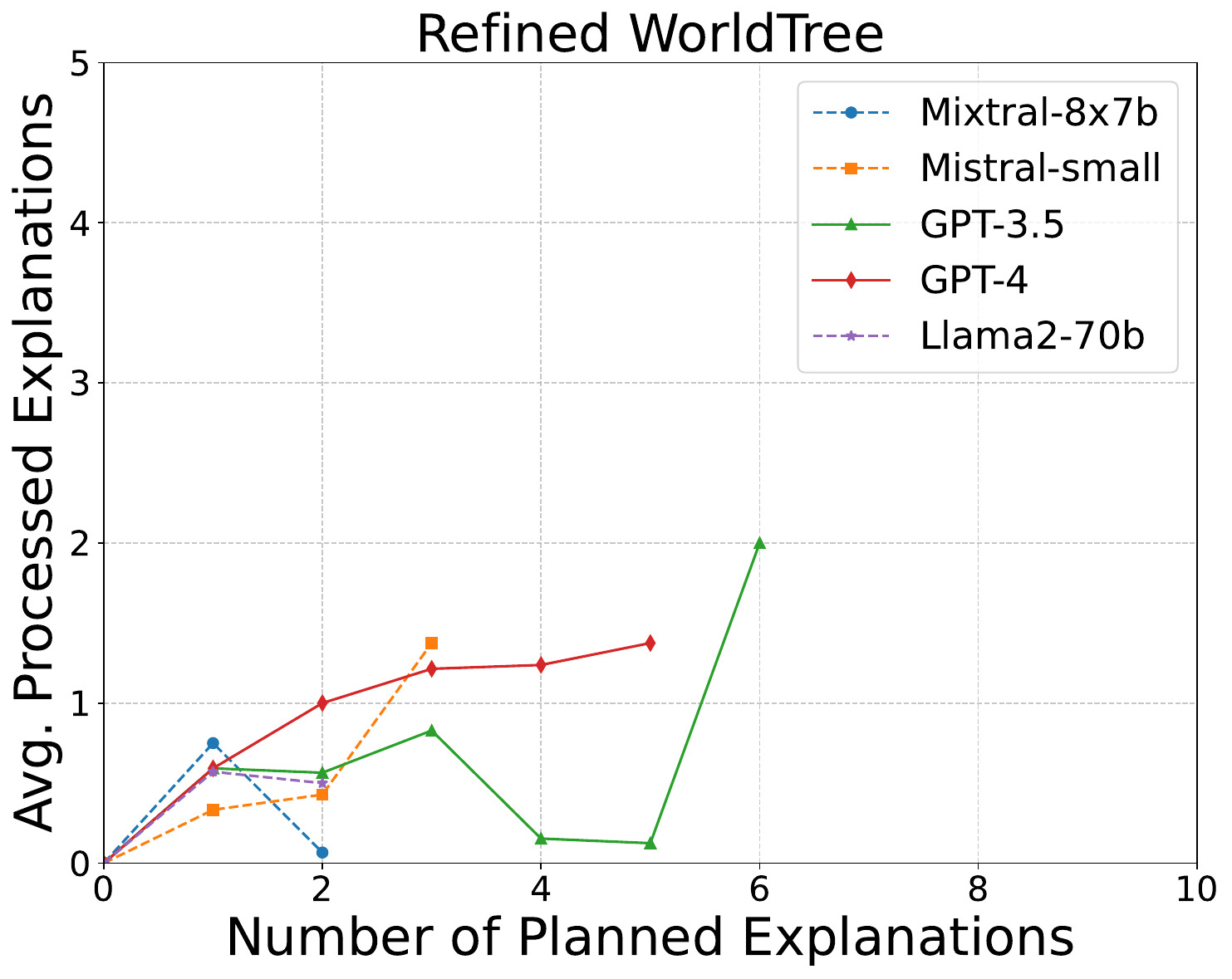}
  \caption{}
\end{subfigure}

\begin{subfigure}{.32\textwidth}
  \centering
  \includegraphics[width=\linewidth]{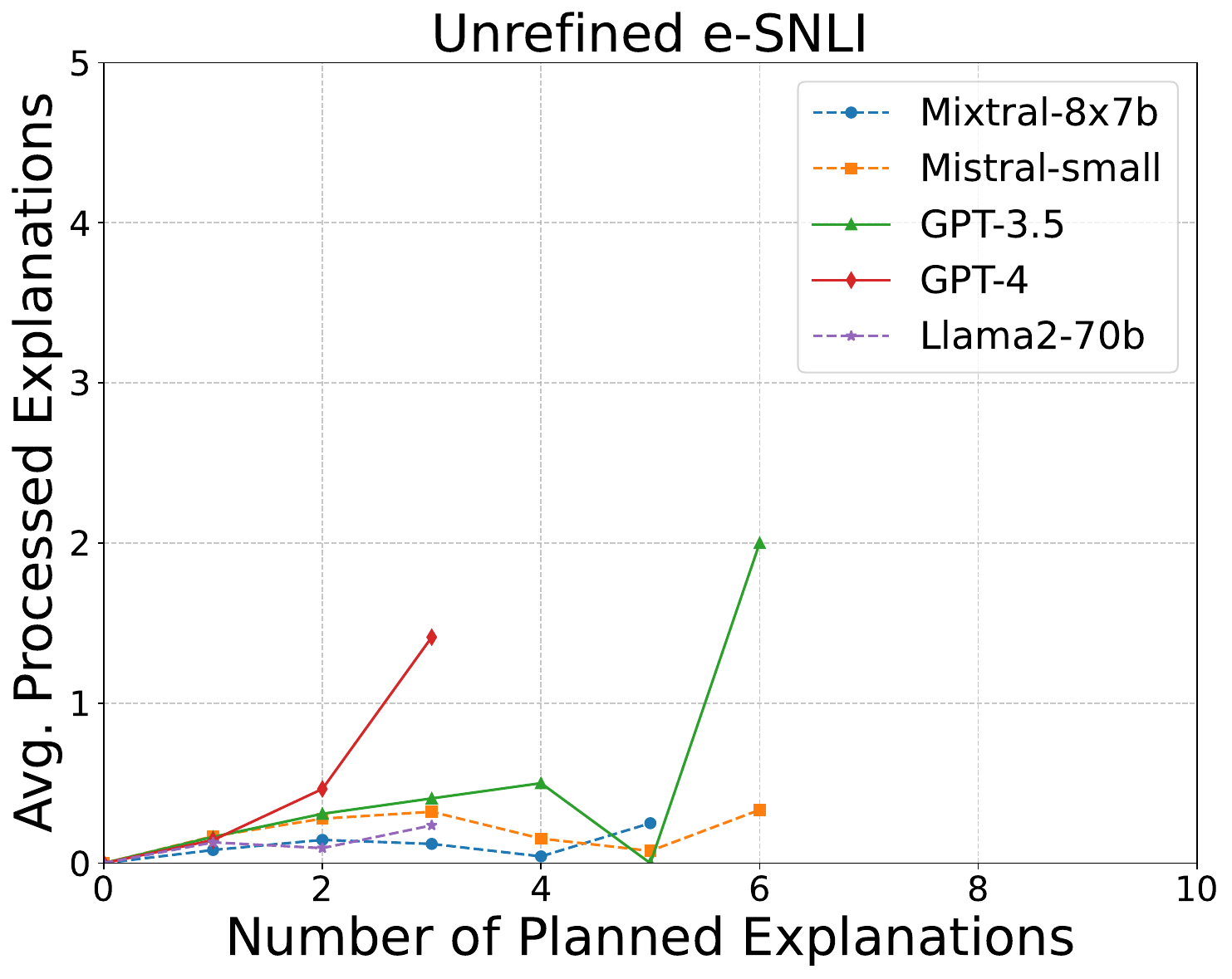}
  \caption{}
\end{subfigure}
\begin{subfigure}{.32\textwidth}
  \centering
  \includegraphics[width=\linewidth]{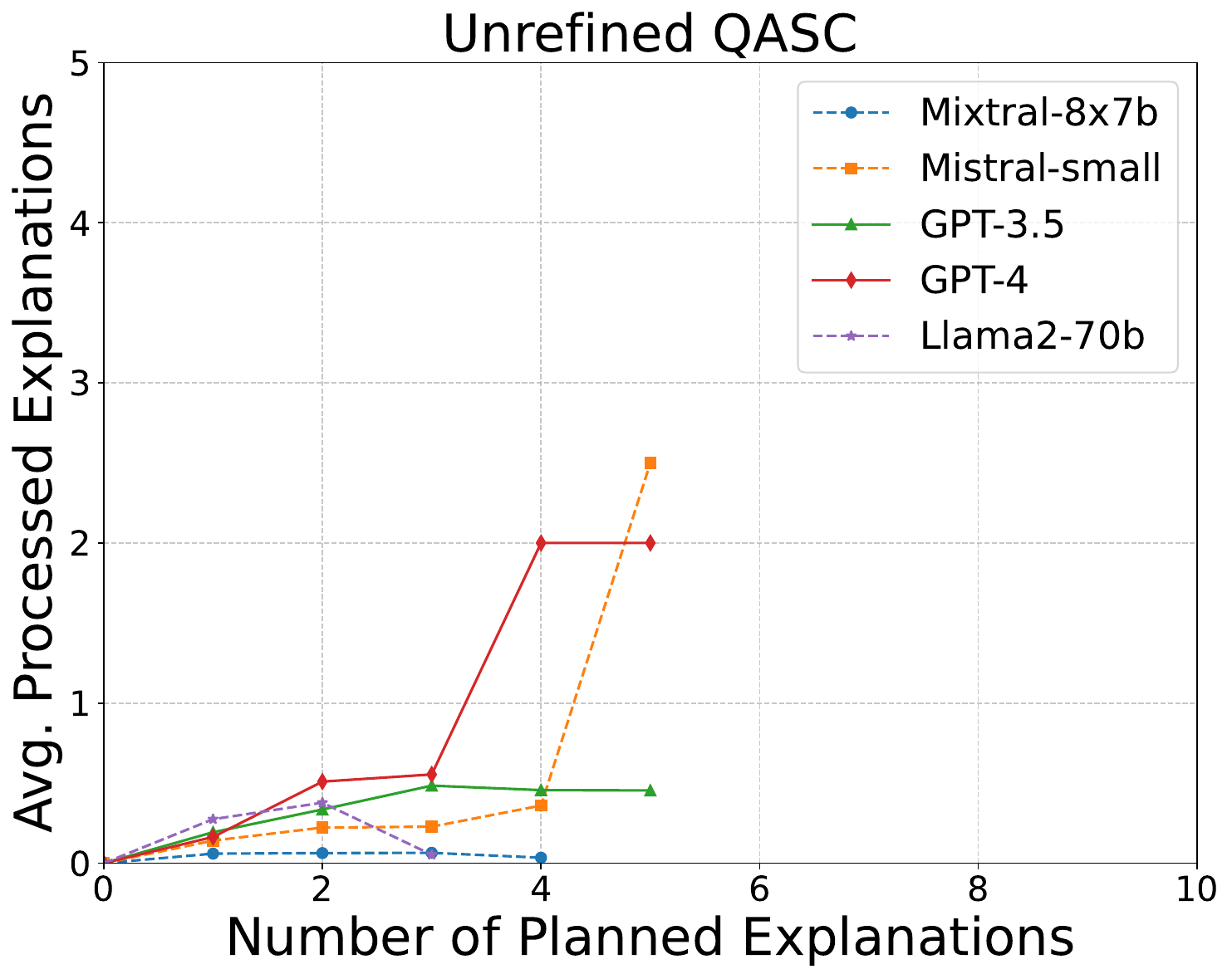}
  \caption{}
\end{subfigure}
\begin{subfigure}{.32\textwidth}
  \centering
  \includegraphics[width=\linewidth]{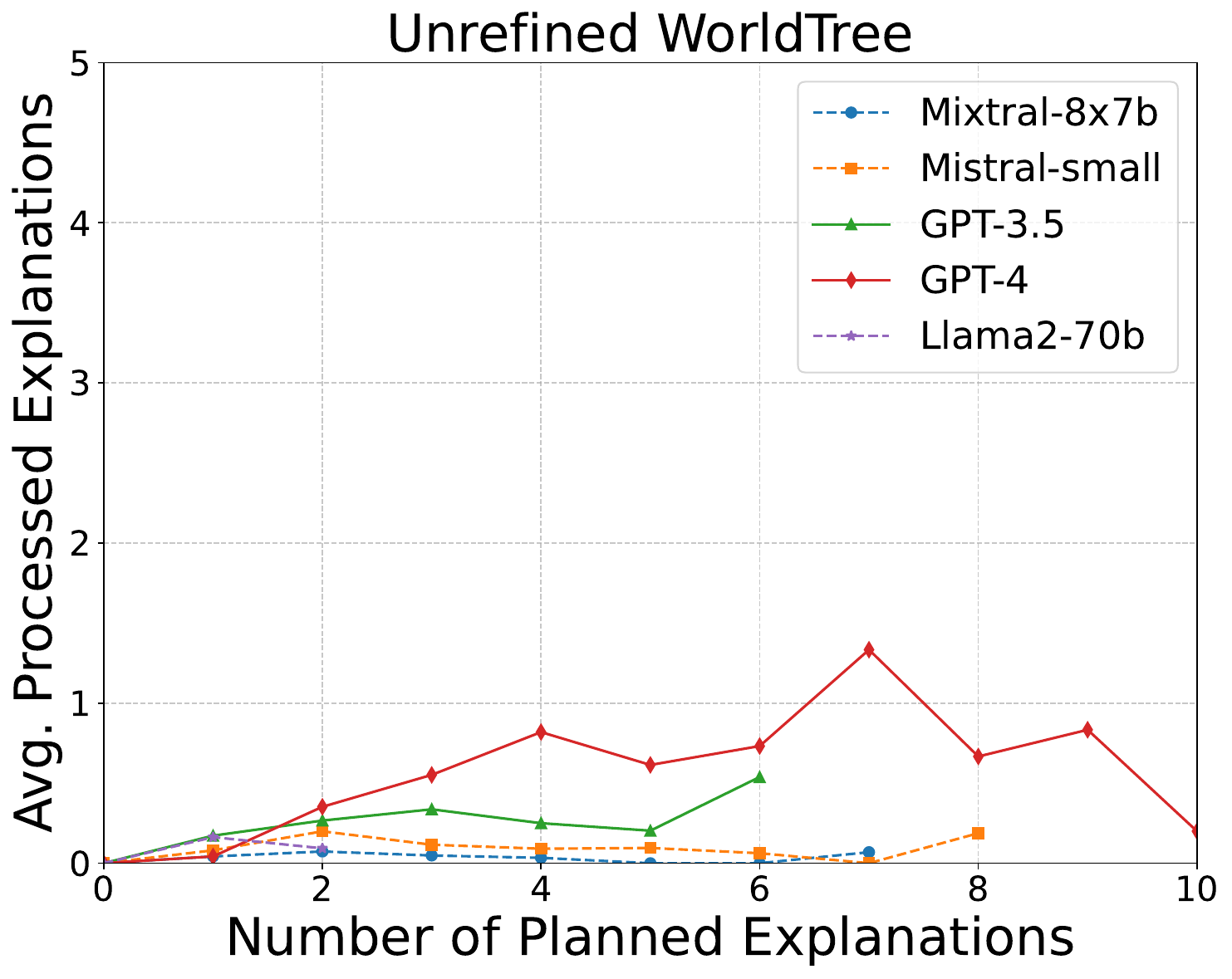}
  \caption{}
\end{subfigure}
\caption{Average Progressed Explanations against Number of Planned Explanations in Refined and Unrefined e-SNLI, QASC and WorldTree Dataset}
\label{fig:avg_pro_against_plan}
\end{figure*}

\begin{figure*}[htbp]
\centering
\begin{subfigure}{.32\textwidth}
  \centering
  \includegraphics[width=\linewidth]{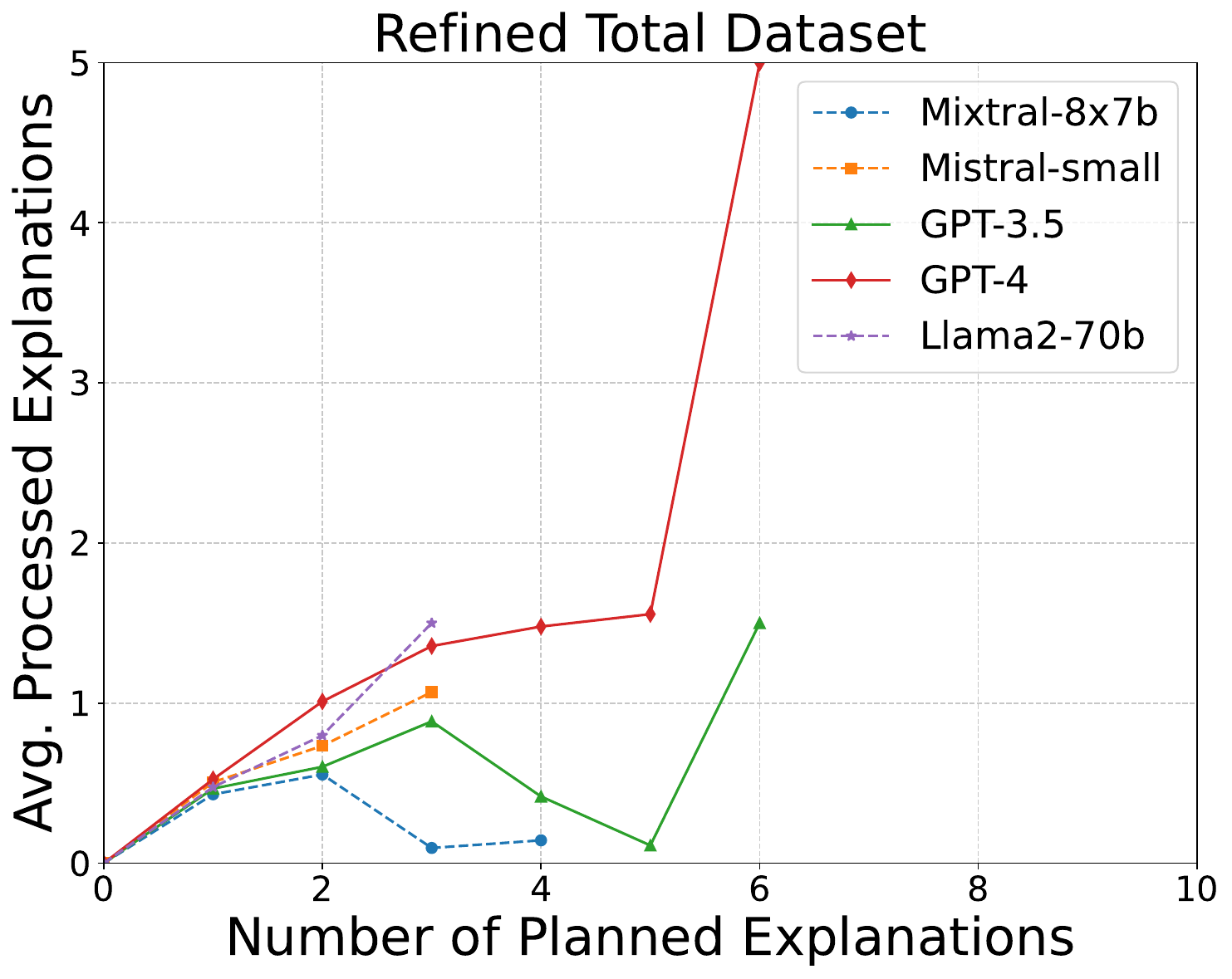}
  \caption{}
  \label{fig:sub1}
\end{subfigure}
\begin{subfigure}{.32\textwidth}
  \centering
  \includegraphics[width=\linewidth]{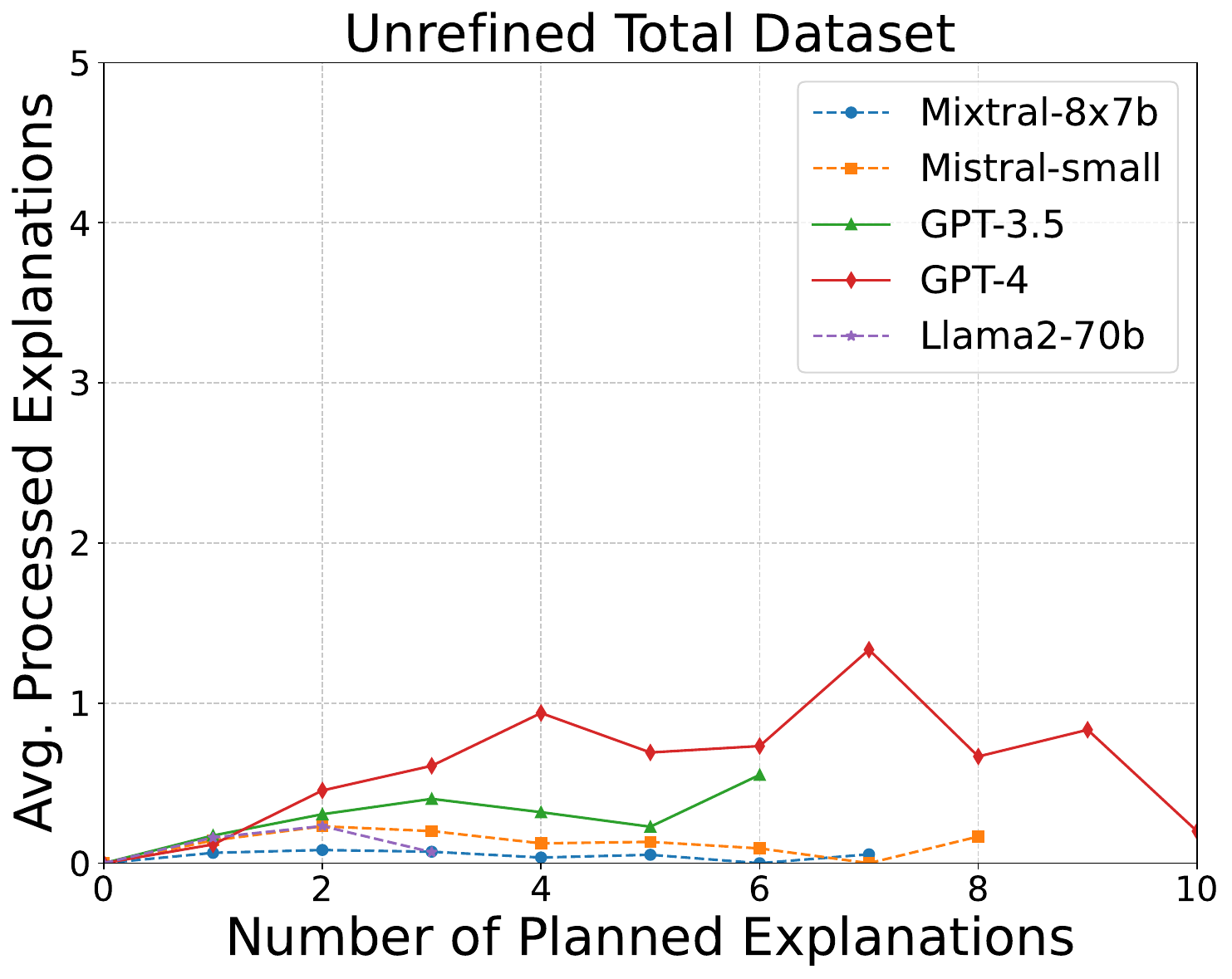}
  \caption{}
  \label{fig:sub2}
\end{subfigure}
\begin{subfigure}{.32\textwidth}
  \centering
  \includegraphics[width=\linewidth]{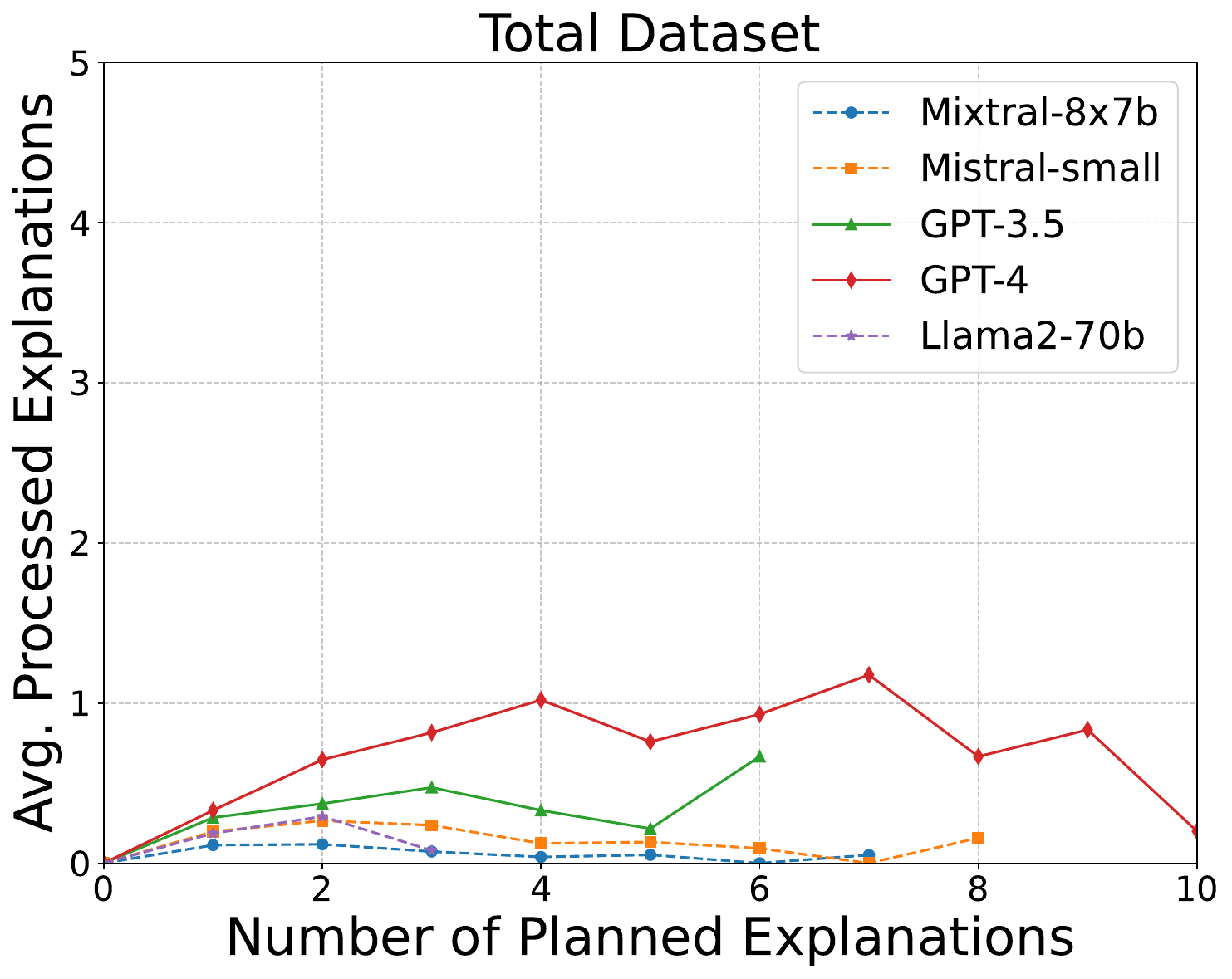}
  \caption{}
\end{subfigure}
\caption{Average Progressed Explanations against Number of Planned Explanations for Refined, Unrefined, and Combined Across All Datasets}
\label{fig:avg_pro_against_total}
\end{figure*}

\begin{figure*}[htbp]
\centering
\begin{subfigure}{.32\textwidth}
  \centering
  \includegraphics[width=\linewidth]{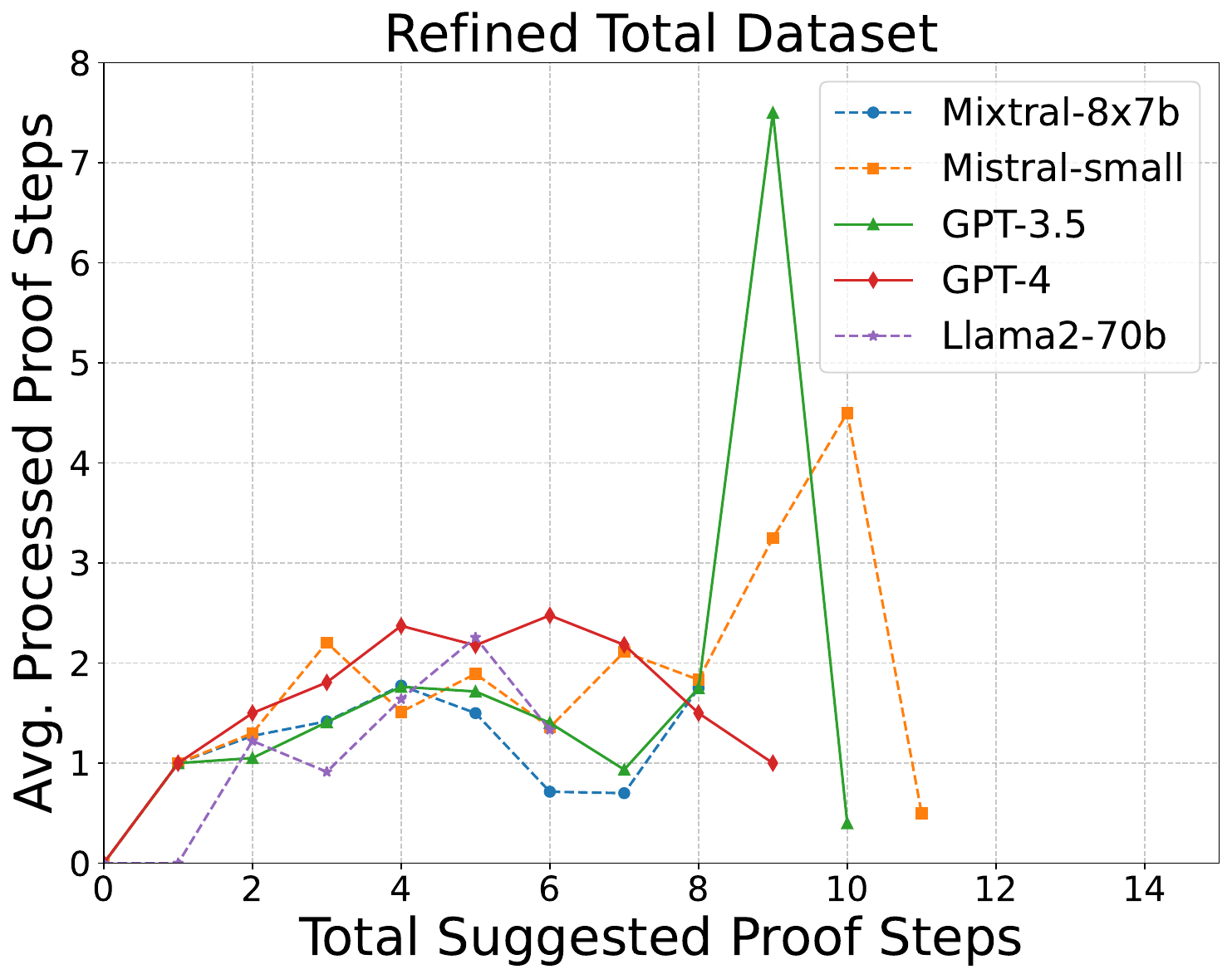}
  \caption{}
  \label{fig:sub1}
\end{subfigure}
\begin{subfigure}{.32\textwidth}
  \centering
  \includegraphics[width=\linewidth]{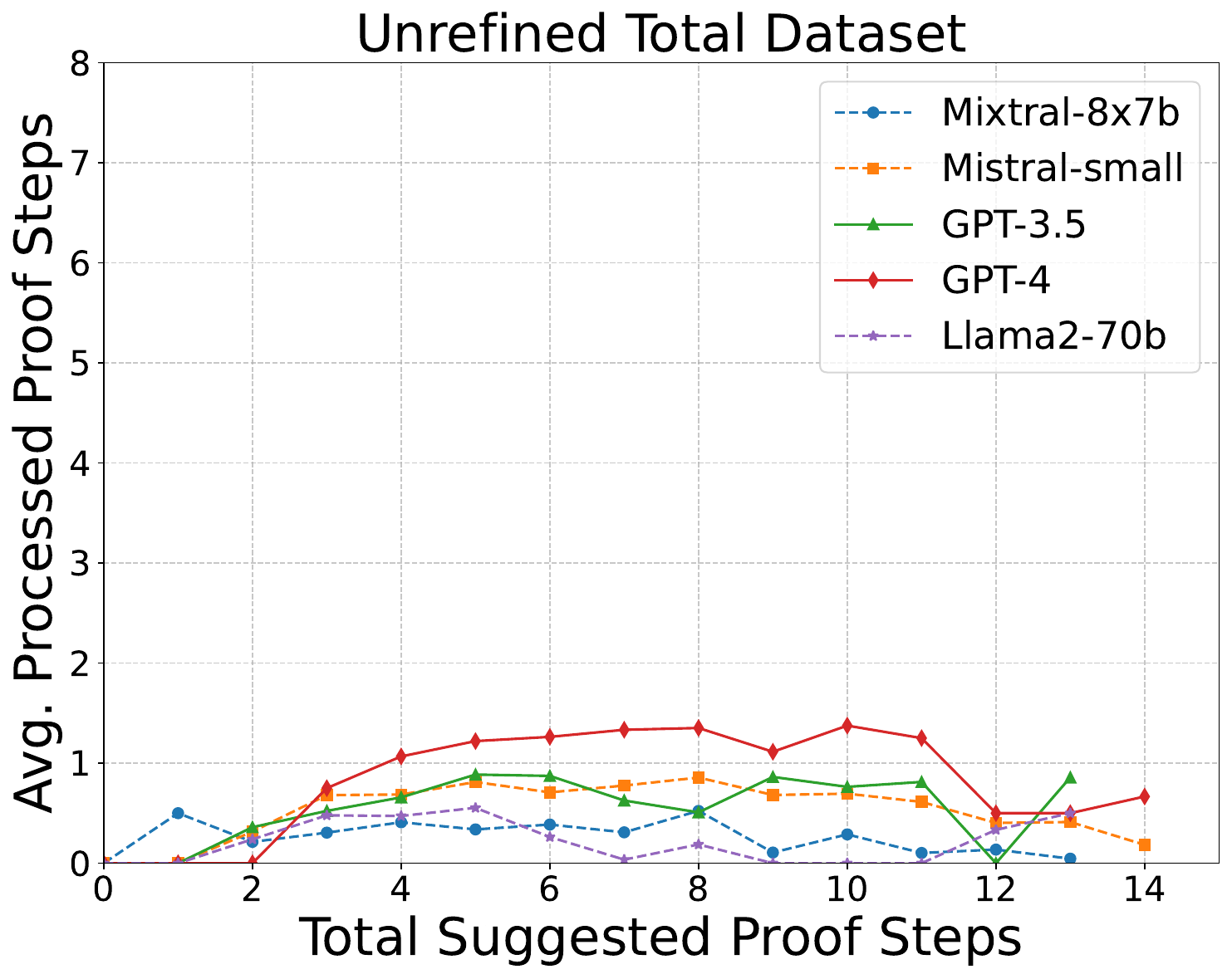}
  \caption{}
  \label{fig:sub2}
\end{subfigure}
\begin{subfigure}{.32\textwidth}
  \centering
  \includegraphics[width=\linewidth]{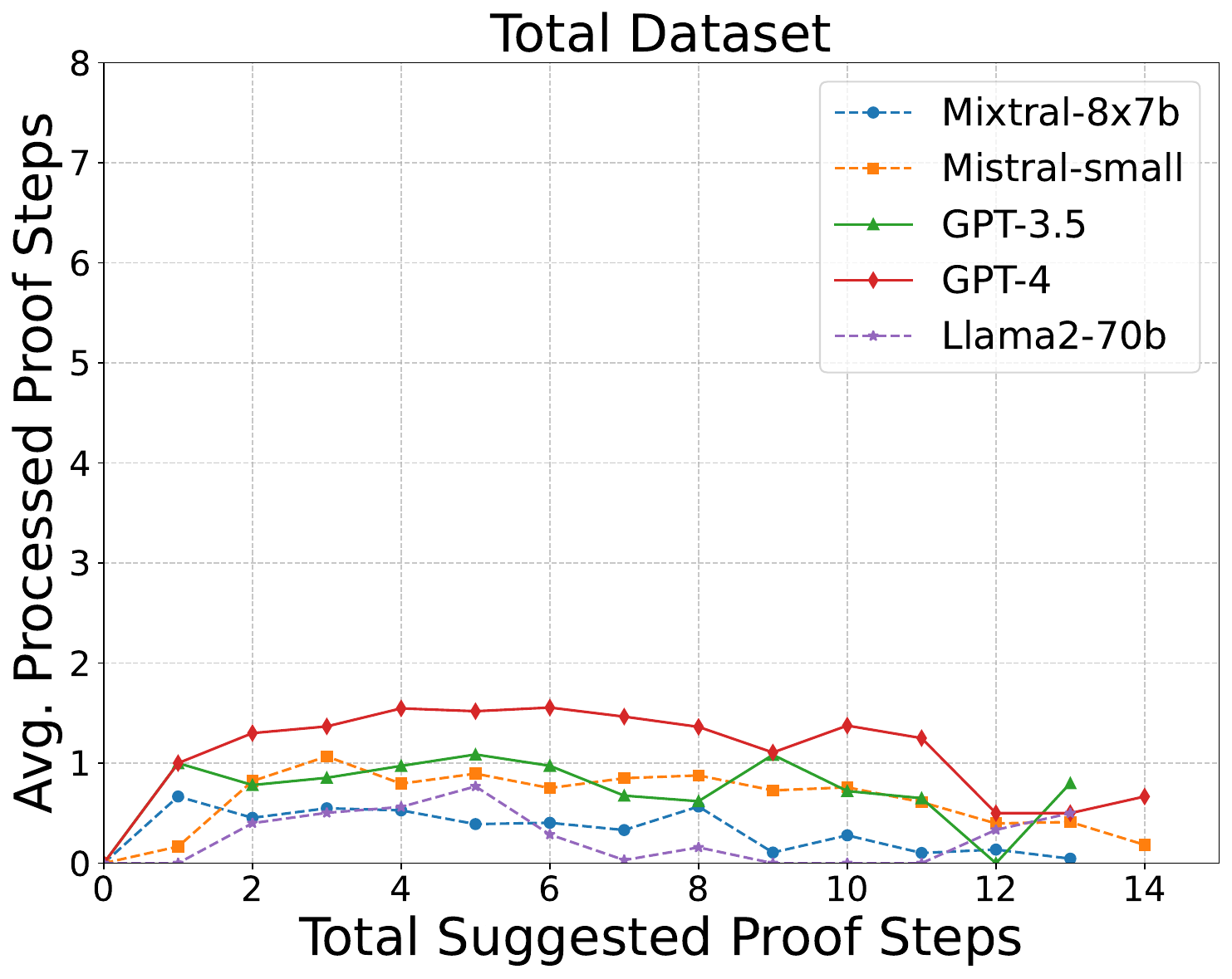}
  \caption{}
\end{subfigure}
\caption{Average Processed Proof Steps against Total Suggested Proof Steps for Refined, Unrefined, and Combined Across All Datasets}
\label{fig:avg_processed_against_total_total_dataset}
\end{figure*}

Figure \ref{fig:avg_pro_against_plan} and Figure \ref{fig:avg_pro_against_total} shows experiments on average number of successfully processed explanatory sentences in one proof against total planned explanatory sentences in a suggest proof. Figure \ref{fig:avg_processed_against_total_total_dataset} also shows the comparison of average processed proof steps against total suggested proof steps in all dataset.

\subsection{Prompts}
Temperature settings were adjusted to 0 for GPT-3.5 and GPT-4, and to 0.01 for Llama2-70b, Mixtral-8x7b, and Mistral-small, aiming to achieve both determinism in the output and effective code generation for theorem prover. 
\subsubsection{Autoformalisation}
\label{appendix_autoformalisation_prompts}
\begin{figure}[htp]
    \centering
    \includegraphics[width=0.5\textwidth]{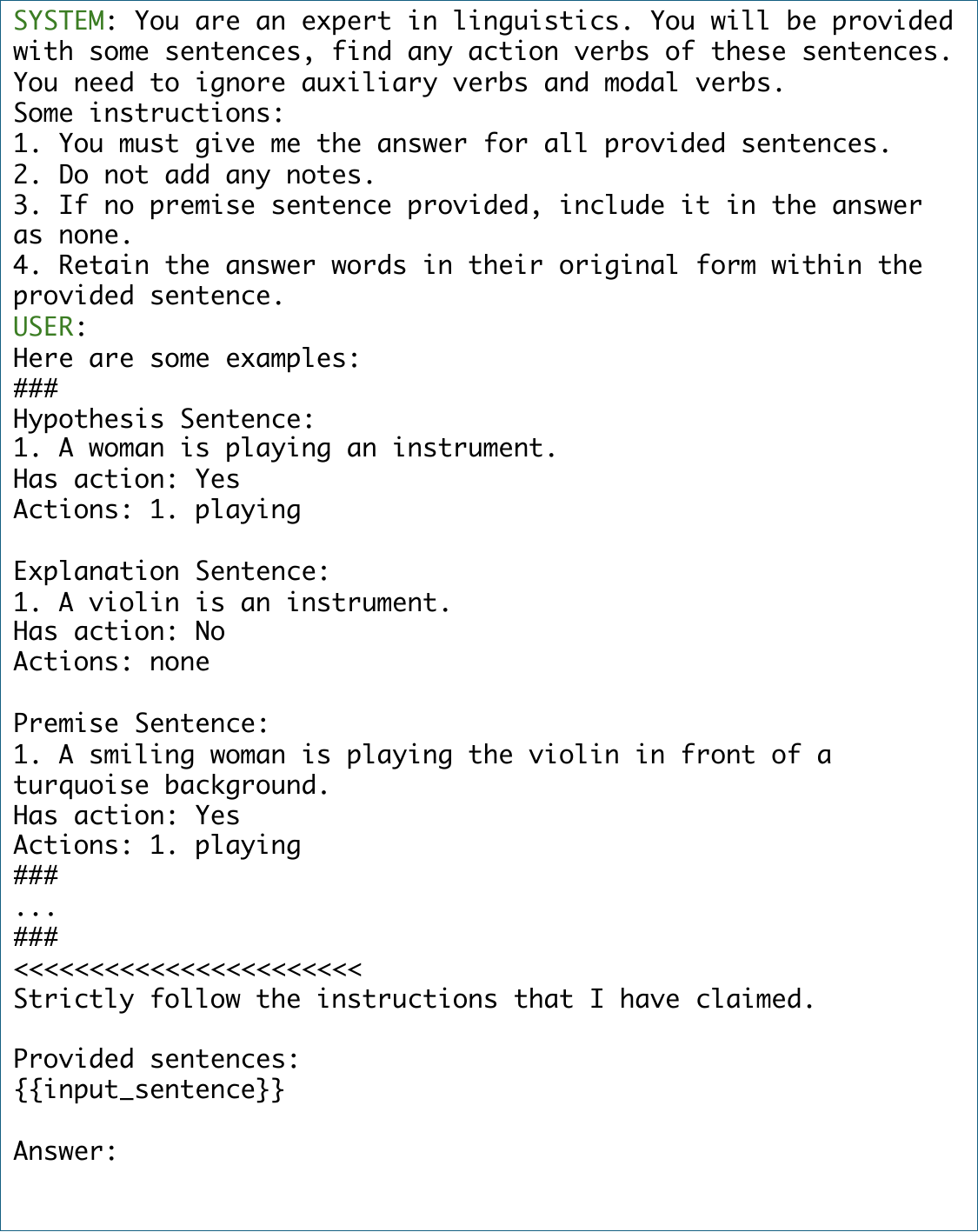}
    \caption{Prompts for detecting event-related words in the given sentences}
\label{fig:prompts_get_event}
\end{figure}

\begin{figure}[htp]
    \centering
    \includegraphics[width=0.5\textwidth]{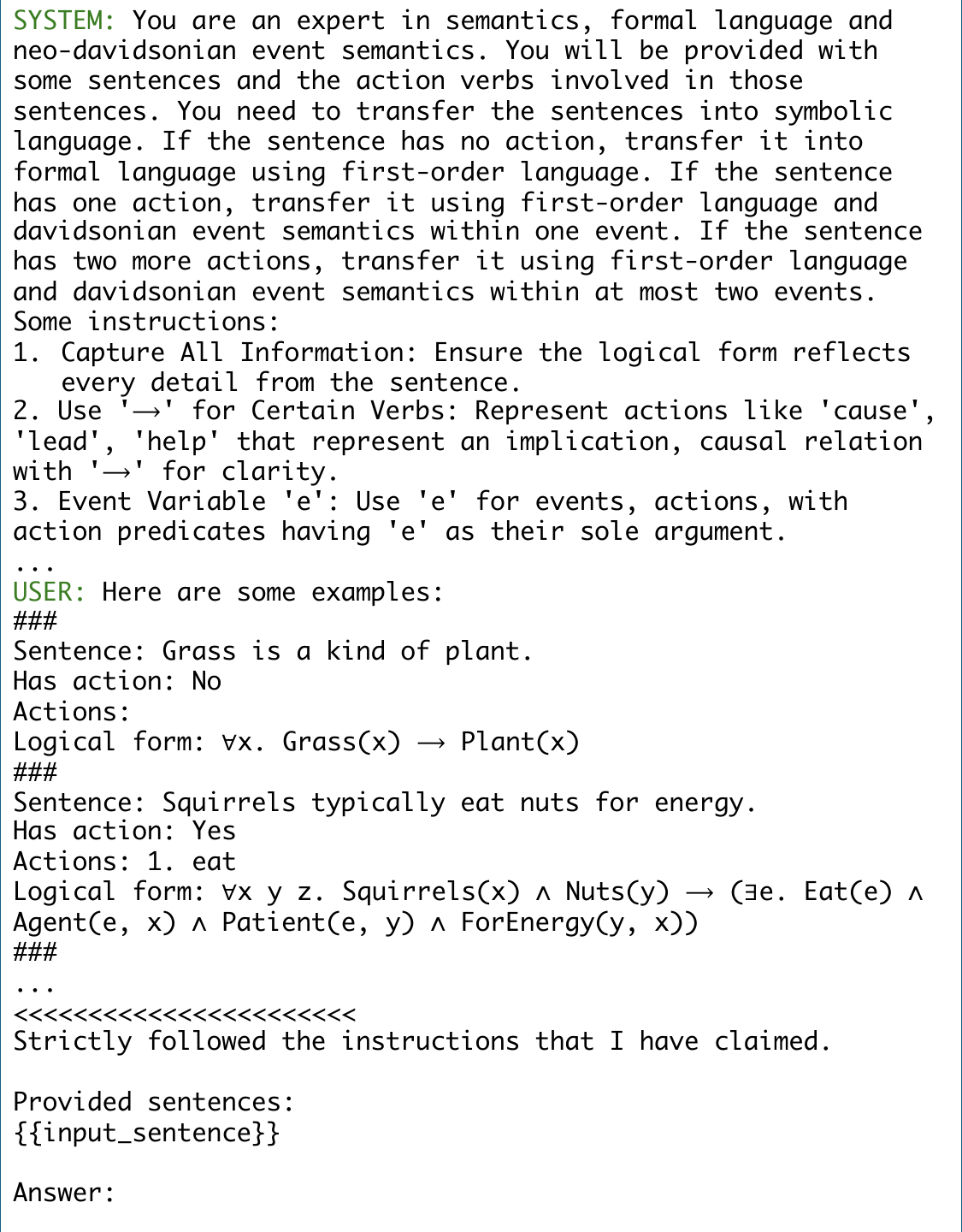}
    \caption{Prompts for converting natural language sentences into logical form representations}
\label{fig:prompts_transfer_logical}
\end{figure}

\begin{figure}[htp]
    \centering
    \includegraphics[width=0.5\textwidth]{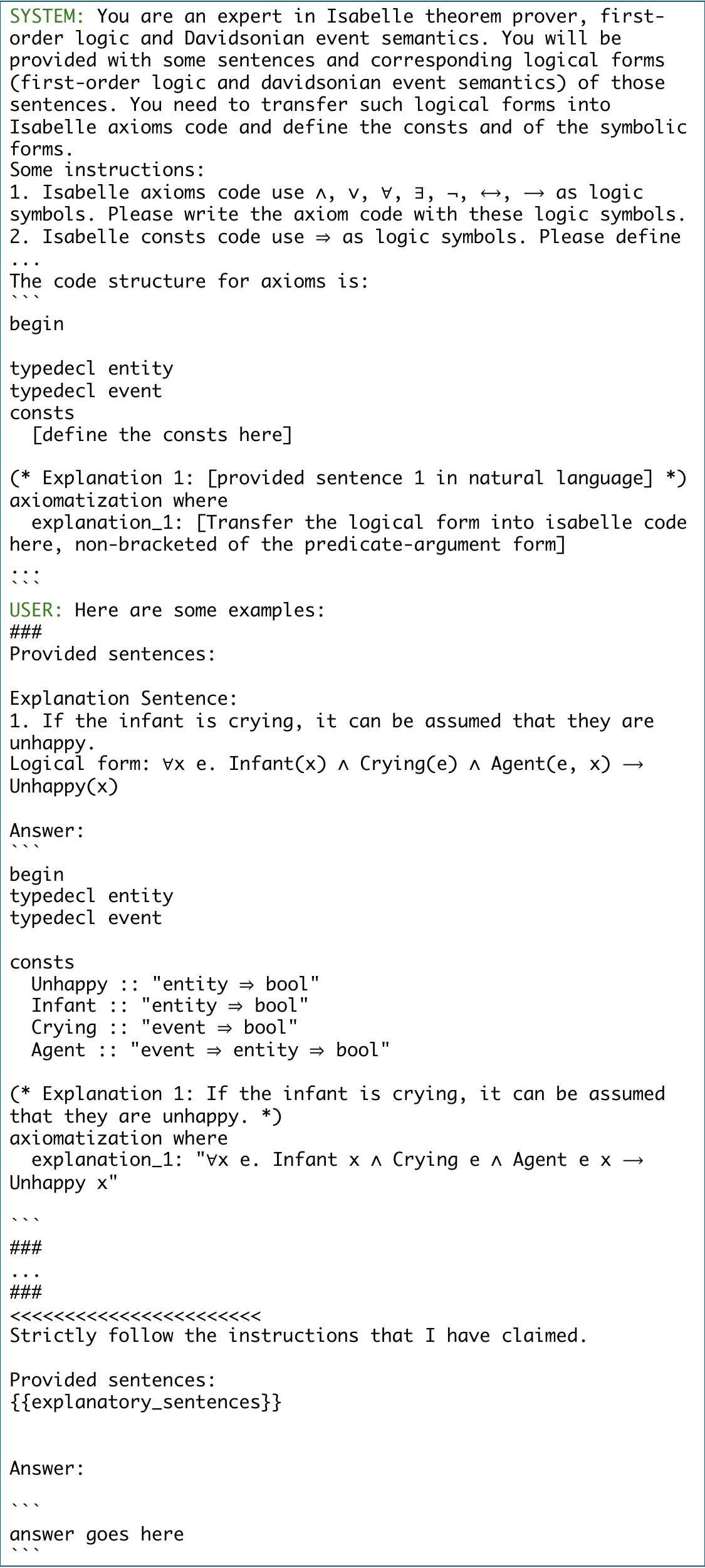}
    \caption{Prompts for converting logical form into Isabelle/HOL code format for building the axioms and type declaration}
\label{fig:prompts_build_axiom}
\end{figure}

\begin{figure}[htp]
    \centering
    \includegraphics[width=0.5\textwidth]{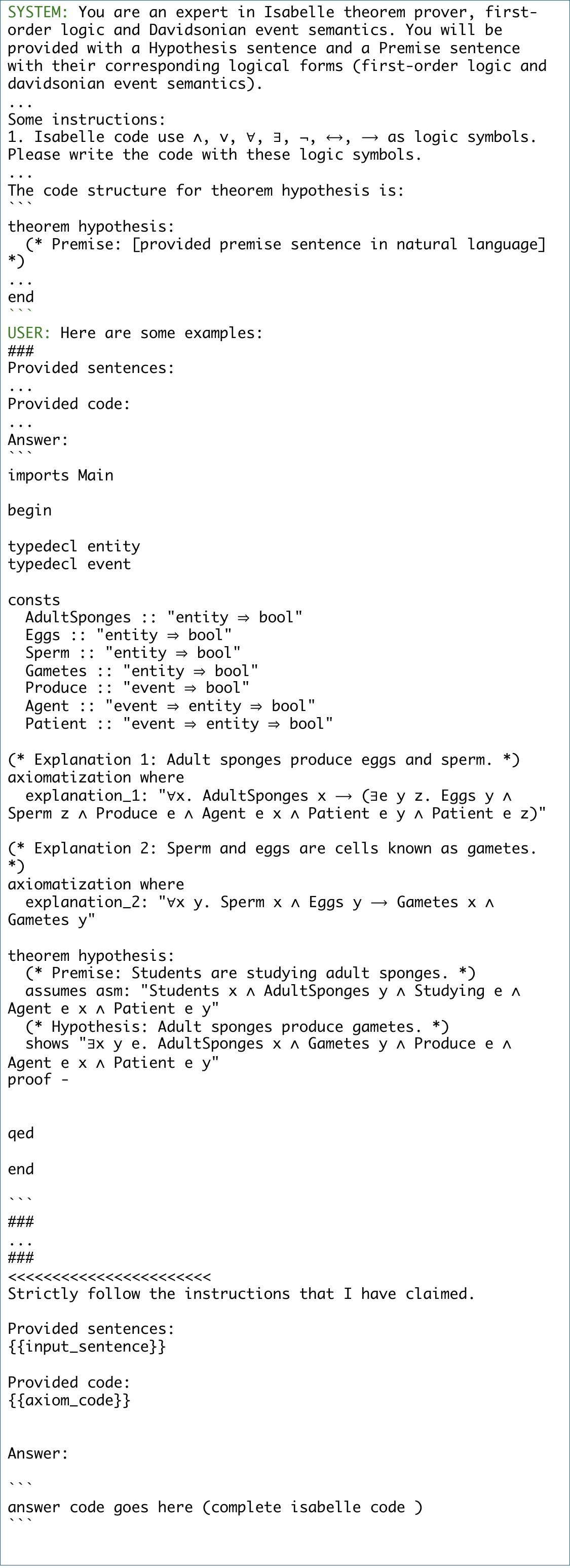}
    \caption{Prompts for building the theorem code part of the Isabelle/HOL theory}
\label{fig:prompts_build_theorem}
\end{figure}

\begin{figure}[htp]
    \centering
    \includegraphics[width=0.5\textwidth]{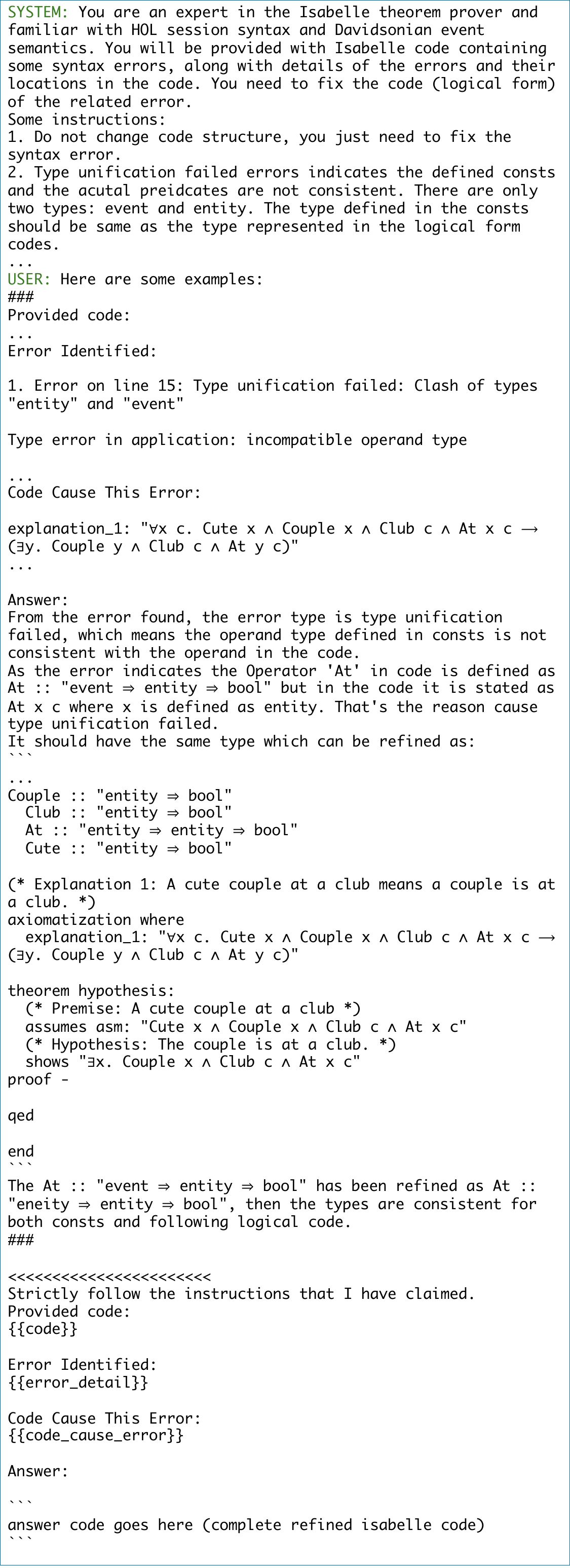}
    \caption{Prompts for how to refine the identified syntax errors in the constructed code}
\label{fig:prompts_refine_syntax}
\end{figure}

Figure \ref{fig:prompts_get_event} displays the prompts used to identify action verbs (events) within the premise, explanation, and hypothesis sentences, representing events in Davidsonian-event semantics. Figure \ref{fig:prompts_transfer_logical} displays the prompts used to transfer natural language to logical forms based on the identified events verbs. Figure \ref{fig:prompts_build_axiom} shows how to convert logical forms into Isabelle/HOL code (axioms and type declaration). Figure \ref{fig:prompts_build_theorem} shows how to convert the premise and hypothesis sentences into the Isabelle/HOL theorem code, based on the previously constructed axioms code. Figure \ref{fig:prompts_refine_syntax} shows how to refine the syntax errors based on the types of errors, the provided code, the error messages, and the locations of the errors within the code.

\subsubsection{Proof Construction}
\label{appendix_proof_construction_prompts}
\begin{figure}[htp]
    \centering
    \includegraphics[width=0.5\textwidth]{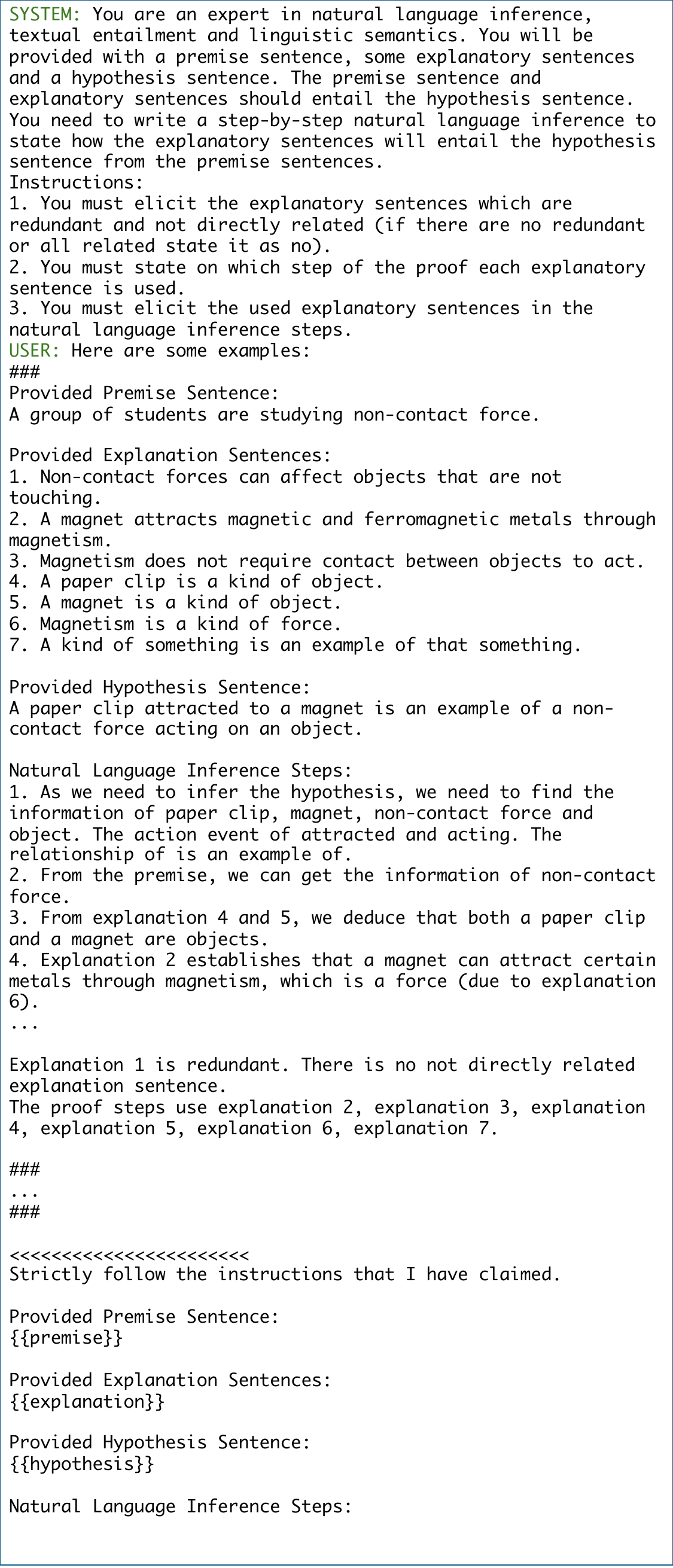}
    \caption{Prompts for how to make a step-by-step preliminary inference strategy}
\label{fig:prompts_rough_inference}
\end{figure}

\begin{figure}[htp]
    \centering
    \includegraphics[width=0.5\textwidth]{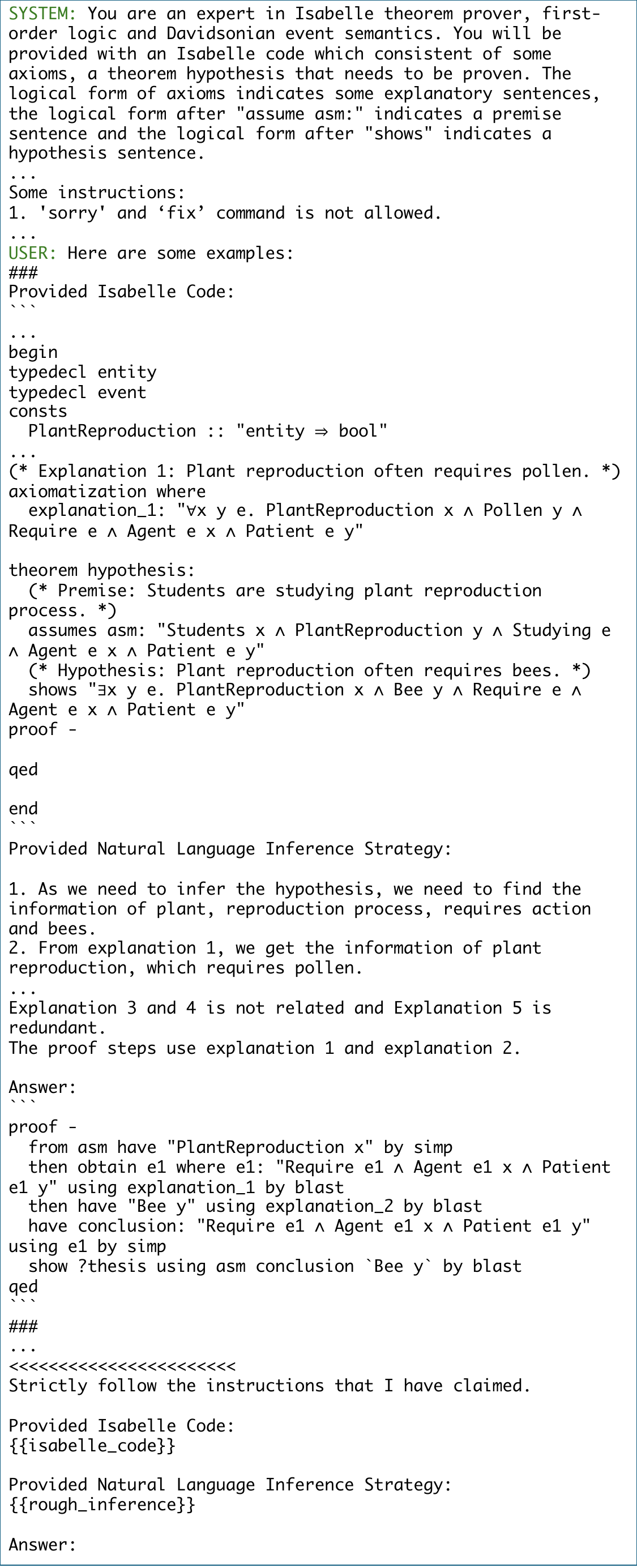}
    \caption{Prompts for how to build a proof for Isabelle/HOL proof assistant}
\label{fig:prompts_build_proofs}
\end{figure}

Figure \ref{fig:prompts_rough_inference} shows the prompts for making a preliminary inference strategy, which also identifies redundant and related explanatory sentences that will be used for proof generation. Figure \ref{fig:prompts_build_proofs} shows the prompts for building the proof steps used for Isabelle/HOL Proof assistant based on the provided inference strategy. 

\subsubsection{Explanation Refinement}
\label{appendix_refine_explanation_prompts}
\begin{figure}[htp]
    \centering
    \includegraphics[width=0.5\textwidth]{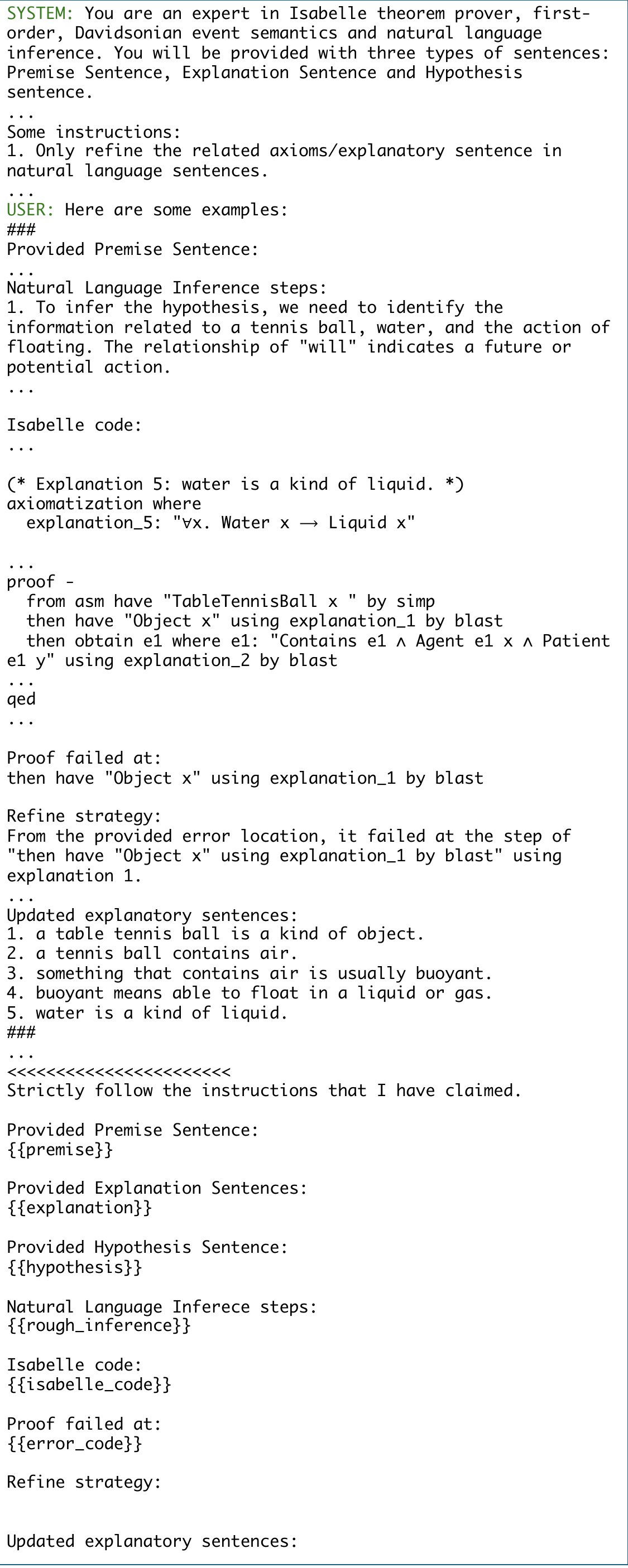}
    \caption{Prompts for how to refine the explanatory sentences}
\label{fig:prompts_refine_explanation}
\end{figure}

Figure \ref{fig:prompts_refine_explanation} shows how to refine the explanatory sentences based on the provided information.

\subsection{Examples of Explanation Refinement}
\label{examples_refine_unrefine}

\begin{figure*}[htbp]
    \centering
    \includegraphics[width=1\textwidth]{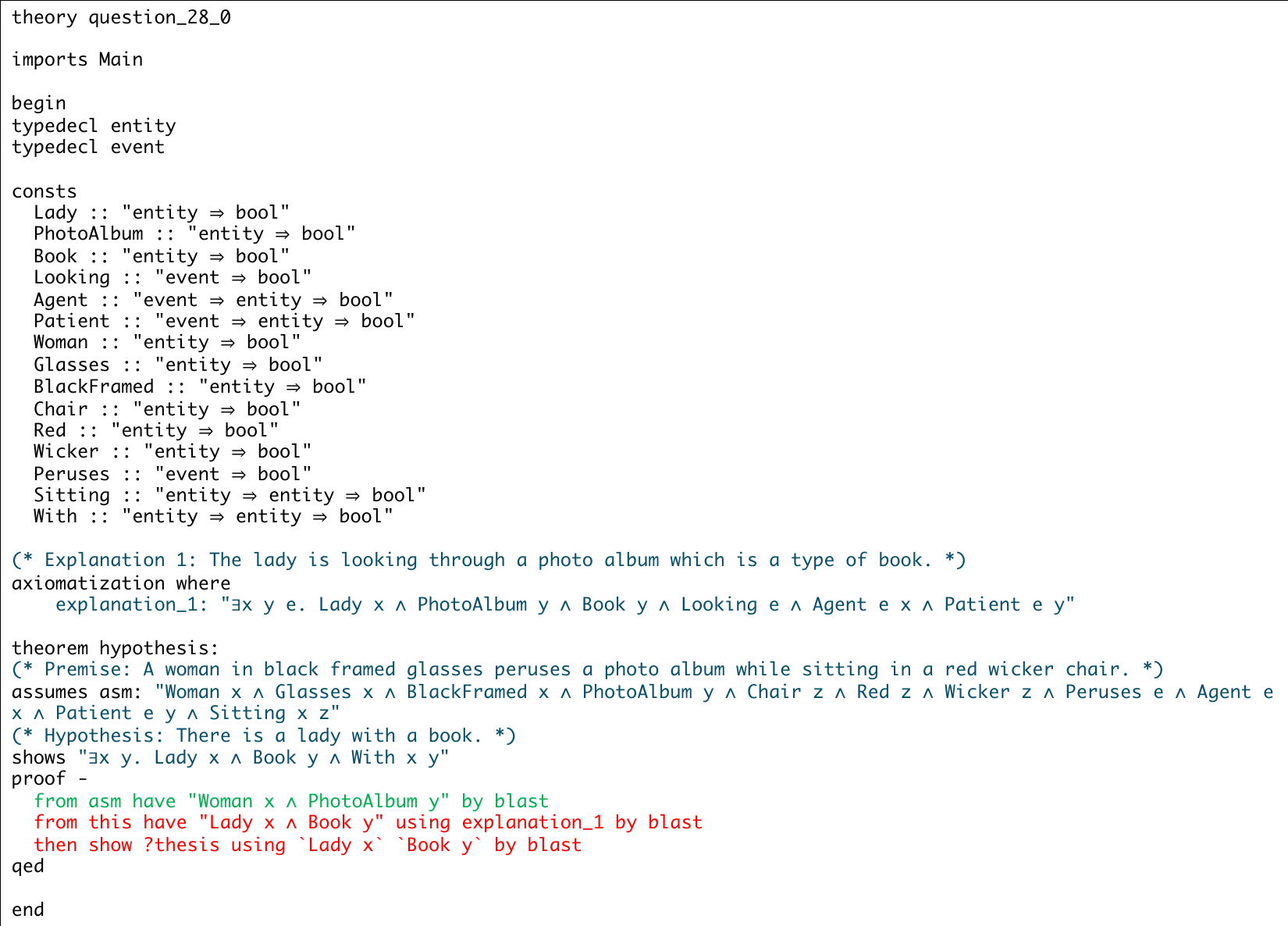}
    \caption{The Isabelle theory code for table \ref{e-snli_example_1_table} iteration 0}
\label{fig:code_esnli_lady_book_it0}
\end{figure*}

\begin{figure*}[htbp]
    \centering
    \includegraphics[width=1\textwidth]{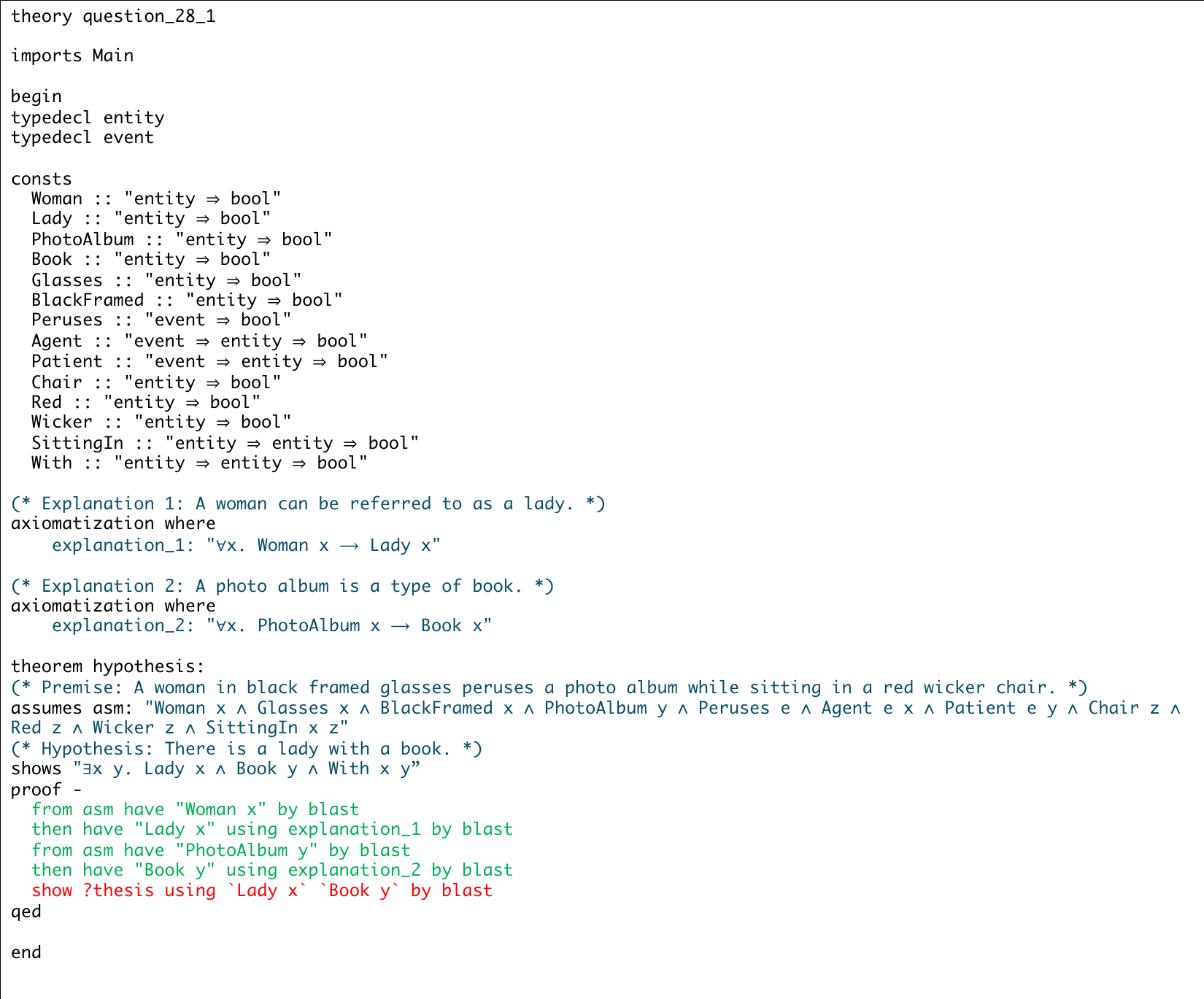}
    \caption{The Isabelle theory code for table \ref{e-snli_example_1_table} iteration 1}
\label{fig:code_esnli_lady_book_it1}
\end{figure*}

\begin{figure*}[htbp]
    \centering
    \includegraphics[width=1\textwidth]{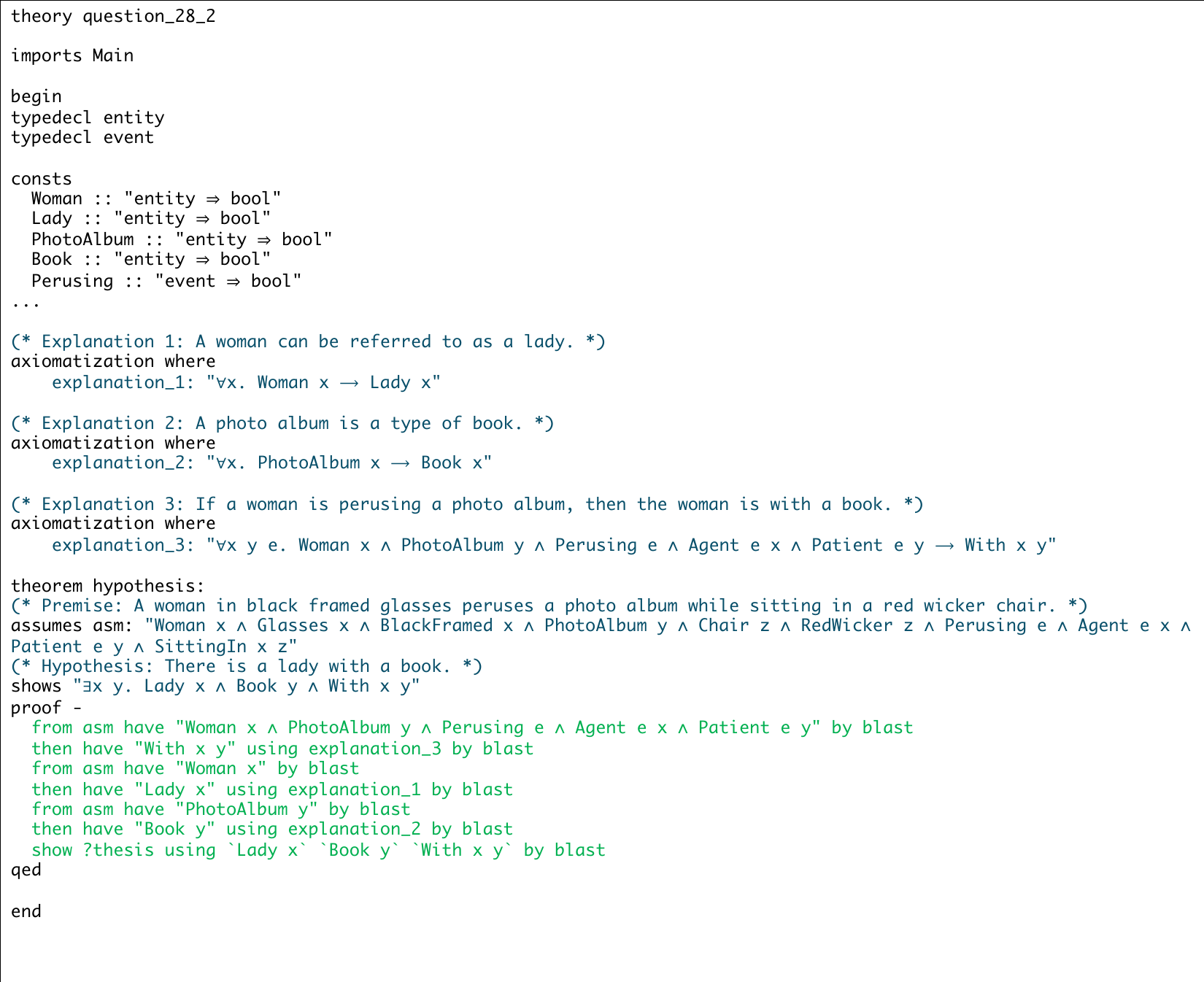}
    \caption{The Isabelle theory code for table \ref{e-snli_example_1_table} iteration 2}
\label{fig:code_esnli_lady_book_it2}
\end{figure*}

\begin{figure*}[htbp]
    \centering
    \includegraphics[width=1\textwidth]{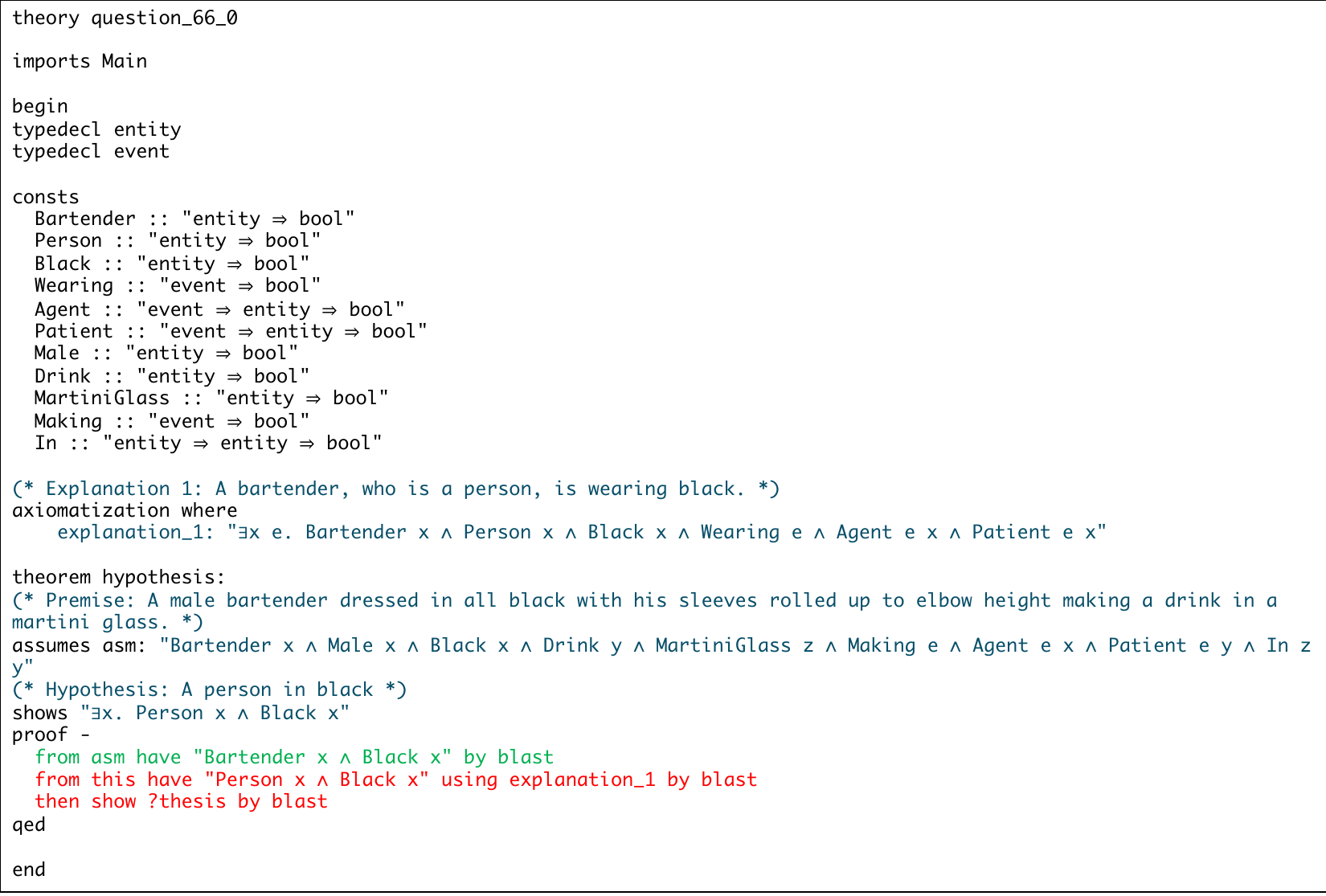}
    \caption{The Isabelle theory code for table \ref{e-snli_example_3_table} iteration 0}
\label{fig:code_esnli_bartender_it0}
\end{figure*}

\begin{figure*}[htbp]
    \centering
    \includegraphics[width=1\textwidth]{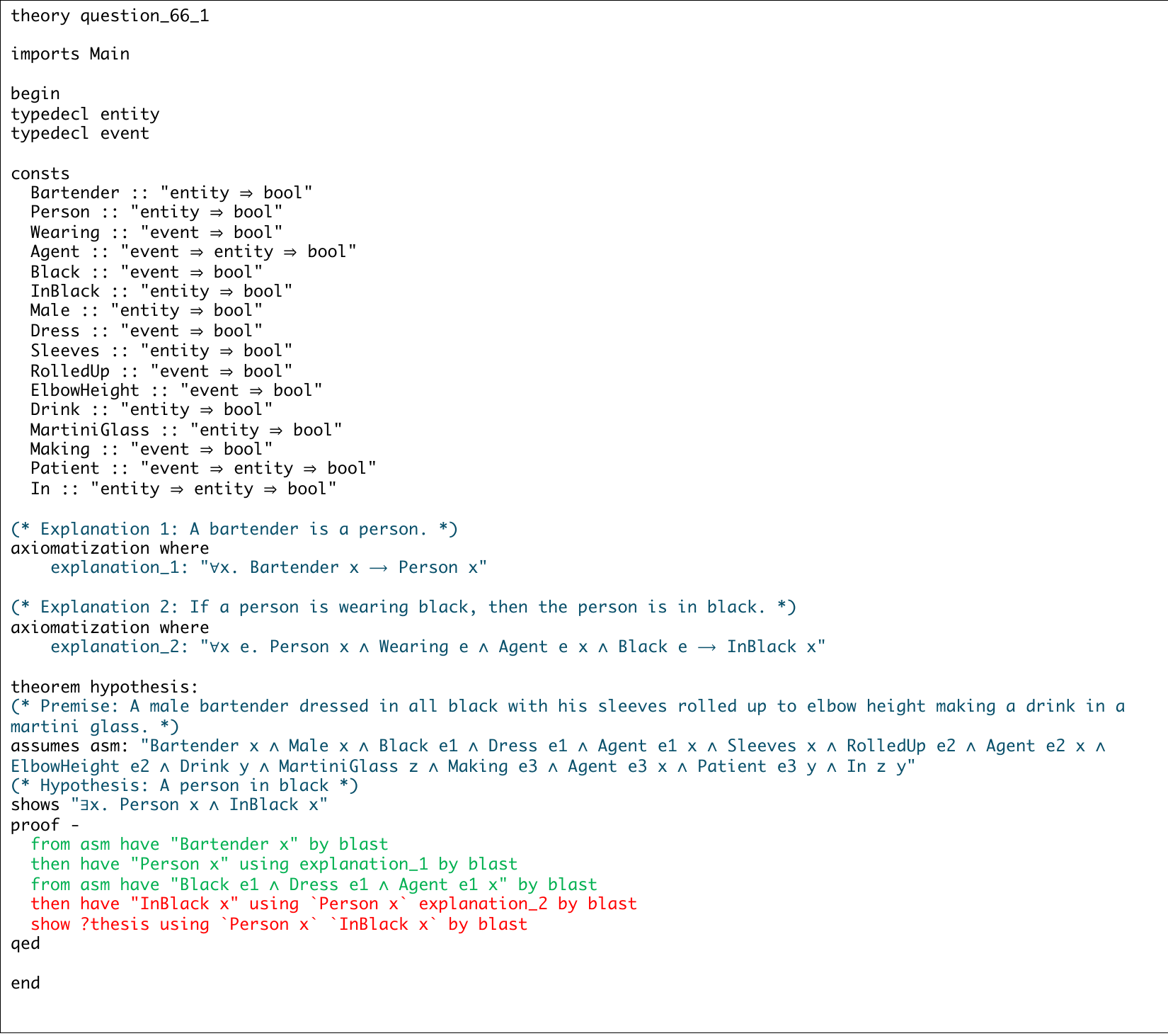}
    \caption{The Isabelle theory code for table \ref{e-snli_example_3_table} iteration 1}
\label{fig:code_esnli_bartender_it1}
\end{figure*}

\begin{figure*}[htbp]
    \centering
    \includegraphics[width=1\textwidth]{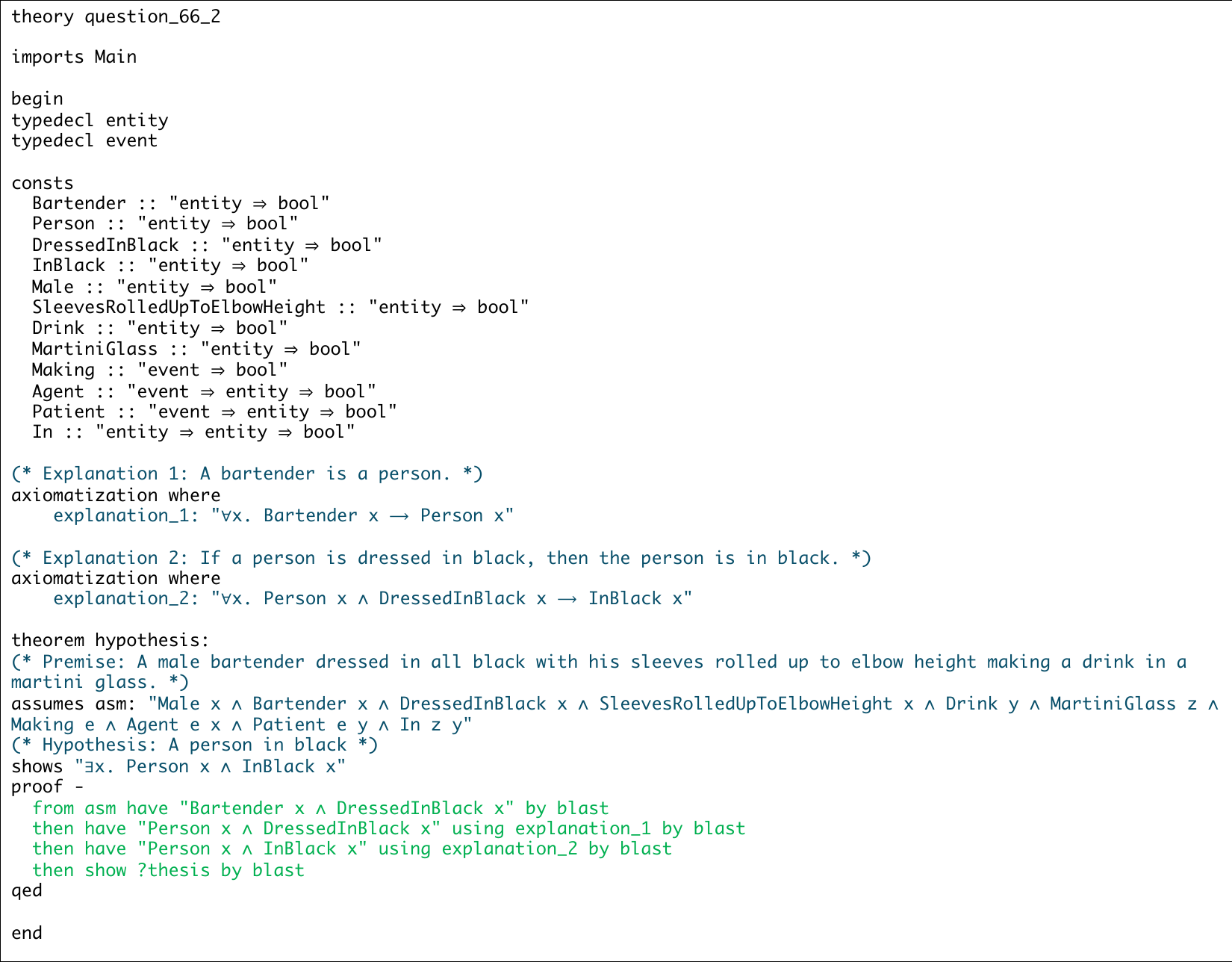}
    \caption{The Isabelle theory code for table \ref{e-snli_example_3_table} iteration 2}
\label{fig:code_esnli_bartender_it2}
\end{figure*}

Table \ref{e-snli_example_1_table} shows an example from the e-SNLI dataset of how the explanation changes after each iteration. Figures \ref{fig:code_esnli_lady_book_it0}, \ref{fig:code_esnli_lady_book_it1}, and \ref{fig:code_esnli_lady_book_it2} illustrate the Isabelle/HOL theory code changes during the refinement process. Table \ref{e-snli_example_3_table} with Figures \ref{fig:code_esnli_bartender_it0}, \ref{fig:code_esnli_bartender_it1}, and \ref{fig:code_esnli_bartender_it2} also show another example of how the explanation is refined after each iteration.

Green code indicates the proof steps that have successfully progressed, while red code shows where the proof failed at that step. More examples can be found at \href{https://github.com/neuro-symbolic-ai/explanation_refinement}{https://github.com/neuro-symbolic-ai/explanation\_refinement}.

\subsection{Datasets and Theorem Prover}
The datasets used in our experiments, including samples from e-SNLI \citep{NEURIPS2018_4c7a167b}, QASC \citep{Khot2019QASC}, and WorldTree \citep{jansen-etal-2018-worldtree}, are all sourced from open academic works. We employed Isabelle as the theorem prover, which is distributed under the revised BSD license. Additionally, the TCP client used for the Isabelle server \citep{shminke2022python} is licensed under Apache-2.0.

\clearpage

\clearpage

\begin{algorithm*}
    
    \SetKwFunction{isOddNumber}{isOddNumber}
    
    \SetKwInOut{KwIn}{Input}
    \SetKwInOut{KwOut}{Output}

    \KwIn{Premise $p$, Explanation $E$, Hypothesis $h$, Isabelle//HOL server $isabelle$, Autoformalisation model $m_a$, Isabelle syntax refinement model $m_{sr}$, Rough inference model $m_{ri}$, Proof step build model $m_{pr}$, Facts filter model $m_f$, Explanation refinement model $m_e$}
    \KwOut{Updated Explanation $E$}
    $valid$ $\leftarrow false$ \\
    isabelle\_theory $\leftarrow [\ ]$ \\
    $iterations \leftarrow 0$  \\
    $max\_iterations \leftarrow 11$ \\
    has\_syntax\_error $\leftarrow$ $false$ \\
    \While{not valid {\bf and} $iterations < max\_iterations$}{
        session\_id $\leftarrow$ session\_build($HOL,isabelle$) \\
        $isabelle$.start(session\_id) \\
        isabelle\_theory $\leftarrow$ transfer\_to\_symbolic($p, E, h$, $m_a$) \\
        messages, error\_content, error\_code $\leftarrow$ $isabelle$.check(isabelle\_theory)\\
        \If{syntax\_errors in messages}{
             has\_syntax\_error $\leftarrow true$ \\
             $it \leftarrow 0$\\
             \While{has\_syntax\_error {\bf and} $it < 3$}{
                  isabelle\_theory = refine\_syntax(messages, error\_content, error\_code, isabelle\_theory, $m_{sr}$)\\
                  messages, error\_content, error\_code $\leftarrow$ $isabelle$.check(isabelle\_theory)\\
                  \eIf{syntax\_errors in messages}{
                      has\_syntax\_error $\leftarrow true$ \\
                       $it \leftarrow it+1$\\
                  }{
                     \textbf{break} 
                  }
                  
             }
        }
        rough\_inference $\leftarrow$ make\_rough\_inference($p, E, h, m_{ri}$)\\
        proof\_steps $\leftarrow$ build\_proof(rough\_inference, $m_{pr}$)\\
        isabelle\_theory $\leftarrow$ isabelle\_theory + proof\_steps\\
        messages, error\_content, error\_code $\leftarrow$ $isabelle$.check(isabelle\_theory)\\
        \eIf{messages is not empty}{
        message $\leftarrow$ messages[0]\\
        $E$ $\leftarrow$ filter($E$, rough\_inference, proof\_steps, $m_f$)\\
        $E$ $\leftarrow$ refine\_explanation(message, error\_content, error\_code, rough\_inference, proof\_steps, $p, E, H, m_e$) \\
        
        }{
          $valid \leftarrow true$\\
          \textbf{break} 
        }
        $iterations \leftarrow iterations+1$\\
        $isabelle$.shutdown()\\
        
    }
    \KwRet{$E$}
    \caption{Explanation-Refiner}
    \label{algorithm_1}
\end{algorithm*}

\clearpage

\clearpage

\begin{table*}[t]
\centering
\small
\begin{tabular}{@{}p{1cm}p{4cm}p{4cm}p{1cm}p{1cm}@{}}  
\toprule
\textbf{Dataset} & \textbf{Sentences} & \textbf{Explanation} & \textbf{Iteration} & \textbf{Validity} \\
\midrule
e-SNLI & \textbf{Premise}: A woman in black framed glasses peruses a photo album while sitting in a red wicker chair. \newline \textbf{Hypothesis}: There is a lady with a book. & The lady is looking through a photo album which is a type of book. &\centering 0\arraybackslash & Invalid \\
\midrule
e-SNLI & \textbf{Premise}: A woman in black framed glasses peruses a photo album while sitting in a red wicker chair. \newline \textbf{Hypothesis}: There is a lady with a book. & A woman can be referred to as a lady. A photo album is a type of book. &\centering 1\arraybackslash & Invalid \\
\midrule
e-SNLI & \textbf{Premise}: A woman in black framed glasses peruses a photo album while sitting in a red wicker chair. \newline \textbf{Hypothesis}: There is a lady with a book. & A woman can be referred to as a lady. A photo album is a type of book. If a woman is perusing a photo album, then the woman is with a book. &\centering 2\arraybackslash & Valid \\
\bottomrule
\end{tabular}
\caption{An example of how the explanation sentences in e-SNLI can be refined with Explanation-Refiner}
\label{e-snli_example_1_table}
\end{table*}

\clearpage

\begin{table*}[htbp]
\centering
\small
\begin{tabular}{@{}p{1cm}p{4cm}p{4cm}p{1cm}p{1cm}@{}}    
\toprule
\textbf{Dataset} & \textbf{Sentences} & \textbf{Explanation} & \textbf{Iteration} & \textbf{Validity} \\
\midrule
e-SNLI & \textbf{Premise}: A male bartender dressed in all black with his sleeves rolled up to elbow height making a drink in a martini glass. \newline \textbf{Hypothesis}: A person in black & A bartender, who is a person, is wearing black. &\centering 0\arraybackslash & Invalid \\
\midrule
e-SNLI & \textbf{Premise}: A male bartender dressed in all black with his sleeves rolled up to elbow height making a drink in a martini glass. \newline \textbf{Hypothesis}: A person in black & A bartender is a person. If a person is wearing black, then the person is in black. &\centering 1\arraybackslash & Invalid \\
\midrule
e-SNLI & \textbf{Premise}: A male bartender dressed in all black with his sleeves rolled up to elbow height making a drink in a martini glass. \newline \textbf{Hypothesis}: A person in black & A bartender is a person. If a person is dressed in black, then the person is in black. &\centering 2\arraybackslash & Valid \\
\bottomrule
\end{tabular}
\caption{An example of how the explanation sentences in e-SNLI can be refined with Explanation-Refiner}
\label{e-snli_example_3_table}
\end{table*}

\end{document}